\documentclass{article}

\usepackage{microtype}
\usepackage{graphicx}
\usepackage{booktabs} 
\usepackage{multirow}
\usepackage[normalem]{ulem}
\usepackage[table,xcdraw]{xcolor}
\usepackage{arydshln}  
\usepackage{pifont}
\newcommand{\xmark}{\ding{55}}

\usepackage{titlesec}
\usepackage{balance}
\usepackage{caption}
\usepackage{xcolor}

\titlespacing*{\section}{0pt}{1.5ex plus .0ex minus .0ex}{1.5ex plus .0ex}
\titlespacing*{\subsection}{0pt}{1.5ex plus .0ex minus .0ex}{1.5ex plus .0ex}

\useunder{\uline}{\ul}{}
\usepackage{hyperref}

\usepackage[accepted]{icml2026}

\usepackage{amsmath}
\usepackage{amssymb}
\usepackage{mathtools}
\usepackage{amsthm}
\usepackage{bbm}
\usepackage{subcaption} 

\usepackage[capitalize,noabbrev]{cleveref}

\theoremstyle{plain}

\theoremstyle{definition}

\theoremstyle{remark}

\usepackage[textsize=tiny]{todonotes}

\icmltitlerunning{EEmo-Logic: A Unified Dataset and Multi-Stage Framework for Comprehensive Image-Evoked Emotion Assessment}

\begin{document}

\twocolumn[
\icmltitle{EEmo-Logic: A Unified Dataset and Multi-Stage Framework for\\Comprehensive Image-Evoked Emotion Assessment}



\icmlsetsymbol{equal}{*}

\begin{icmlauthorlist}
\icmlauthor{Lancheng Gao}{equal,sch1}
\icmlauthor{Ziheng Jia}{equal,sch1}
\icmlauthor{Zixuan Xing}{sch1}
\icmlauthor{Wei Sun}{sch2}
\icmlauthor{Huiyu Duan}{sch1}
\icmlauthor{Guangtao Zhai}{sch1}
\icmlauthor{Xiongkuo Min}{sch1}
\end{icmlauthorlist}

\icmlaffiliation{sch1}{Institute of Image Communication and Network Engineering, Shanghai Key Laboratory of Digital Media Processing and Transmissions, Shanghai Jiao Tong University, Shanghai}
\icmlaffiliation{sch2}{School of Communication \& Electronic Engineering, East China Normal University, Shanghai}

\icmlcorrespondingauthor{Xiongkuo Min}{minxiongkuo@sjtu.edu.cn}

\icmlkeywords{Machine Learning, ICML}

\vskip 0.3in
]



\printAffiliationsAndNotice{\icmlEqualContribution} 

\begin{abstract}
Understanding the multi-dimensional attributes and intensity nuances of  image-evoked emotions is pivotal for advancing machine empathy and empowering diverse human-computer interaction applications. However, existing models are still limited to coarse-grained emotion perception or deficient reasoning capabilities. To bridge this gap, we introduce \textbf{EEmoDB}, the largest image-{\ul e}voked {\ul emo}tion understanding {\ul d}ataset to date. It features $5$ analysis dimensions spanning $5$ distinct task categories, facilitating comprehensive interpretation. Specifically, we compile $1.2M$ question-answering (QA) pairs (EEmoDB-QA) from $125K$ images via automated generation, alongside a $36K$ dataset (EEmoDB-Assess) curated from $25K$ images for fine-grained assessment. Furthermore, we propose \textbf{EEmo-Logic}, an \textbf{all-in-one} multimodal large language model (MLLM) developed via instruction fine-tuning and task-customized group relative preference optimization (GRPO) with novel reward design. Extensive experiments demonstrate that EEmo-Logic achieves robust performance in in-domain and cross-domain datasets, excelling in emotion QA and fine-grained assessment. 
The dataset and code are available at \href{https://github.com/workerred/EEmo-Logic}{https://github.com/workerred/EEmo-Logic}.
\end{abstract}

\begin{figure}[!t]
    \centering
    \vspace{-4pt}
    \includegraphics[width=0.95\linewidth]{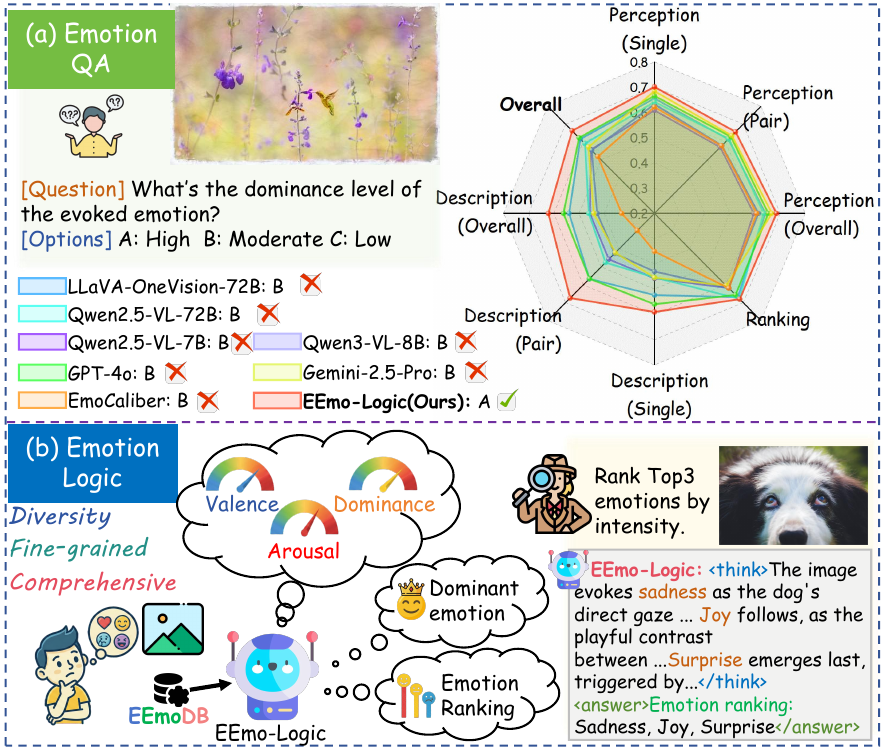}
    \caption{Trained on our constructed \textbf{EEmoDB} via a two-stage paradigm, \textbf{EEmo-Logic} (a) achieves robust question-answering performance on EEmo-Bench, and (b) demonstrates strong capabilities in fine-grained emotion assessment and causal reasoning.}
    \label{fig:spotlight}
\end{figure}

\vspace{-0.6cm}
\section{Introduction} \label{sec:introduction}

Affective image content analysis (AICA) addresses the ``affective gap" linking low-level visual features to high-level semantics \cite{Artphoto}. This challenge is compounded by the psycho-physiological nature of emotion, characterized by intrinsic subjectivity and diversity. Specifically, subjectivity entails inter-personal variance across viewers \cite{emotion_subjective}, while diversity encompasses intra-personal nuances such as intensity fluctuations and mixed emotions \cite{Emotion6}. Mastering these complexities is vital for advancing computational empathy \cite{human-robot-interaction}, thereby facilitating applications ranging from personalized advertising \cite{advertising} and psychological counseling \cite{counseling} to human-computer interaction \cite{human-robot}.

Traditional AICA methods incorporate two psychological paradigms, including categorical emotion states (CES) and dimensional emotion space (DES) \cite{emotion_subjective}. CES models emotions via discrete categories, often focusing on the dominant emotion category (DEC) for robustness \cite{dominant_1,dominant_2}, or employing label distribution learning to address subjectivity \cite{LDL_1}. Conversely, DES analysis typically employs the valence-arousal-dominance (VAD) model to predict scores via regression \cite{VAD, VAD_regression_1}. However, early methods treat these frameworks in isolation and remain limited to label perception, neglecting critical reasoning and problem-solving abilities.

Recent advances in multimodal large language models (MLLMs) demonstrate robust reasoning and task versatility, enabling the development of more practical emotion analysis systems \cite{SFT}.
Although supervised fine-tuning (SFT) improves chain-of-thought (CoT) for DEC classification \cite{EmoVIT}, it neglects emotional diversity. To address this, some studies utilize open-vocabulary (OV) lexicons to list multiple emotions \cite{AffectGPT} or apply confidence validation for DEC \cite{EmoCaliber}, but fail to model intensity nuances or perform DES analysis. Furthermore, diversity-focused efforts stem primarily from the facial expression recognition domain \cite{AffectGPT-R1}, exhibiting limited generalization in AICA contexts (see Sec.~\ref{sec: ood}).
This necessitates a comprehensive AICA framework that resolves fine-grained diversity and intensity issues while incorporating DES analysis.

To this end, we begin by constructing \textbf{EEmoDB}, the largest instruction dataset for image-evoked emotion understanding to date. Synthesized from $7$ AICA datasets, it comprises $128K$ images and $1.23M$ instructions. 
The dataset characterizes $5$ core dimensions, comprising \textbf{valence}, \textbf{arousal}, and \textbf{dominance} at the DES level, alongside \textbf{dominant emotion} and \textbf{emotion ranking} at the CES level. Specifically, emotion ranking captures emotion diversity and intensity \cite{EEmobench}, circumventing the limitations of MLLM distribution prediction \cite{not_distribution}. 
Moreover, we refine the raw annotations via downward emotion mapping as well as automated VAD and reasoning generation to enrich the analysis corpus.
EEmoDB comprises two subsets: \textbf{EEmoDB-QA}, containing $1.2M$ instructions generated from $128K$ single and $355K$ paired images to cover \textbf{perception}, \textbf{ranking}, and \textbf{description} tasks; and \textbf{EEmoDB-Assess}, a curated set of $36K$ instructions from $25K$ images, targeting fine-grained emotional assessment.

Building upon EEmoDB, we propose \textbf{EEmo-Logic}, a MLLM-based framework designed for comprehensive and fine-grained emotion understanding. We employ a two-stage training paradigm: Stage $1$ utilizes low-rank adaptation (LoRA) \cite{lora} SFT on EEmoDB-QA to establish basic emotional comprehension and question-answering (QA) capabilities. Subsequently, Stage $2$ applies group relative preference optimization (GRPO) on the EEmoDB-Assess subset to optimize emotion ranking, VAD scoring, and DEC classification. By tailoring $3$ hierarchical rewards to specific task characteristics, we guide the model toward precise emotion assessment while enhancing causal reasoning.
EEmo-Logic serves as a unified AICA framework that synergizes CES and DES perception, yielding robust zero-shot performance on both in-domain and cross-domain datasets. Our core contributions are threefold:

\vspace{-0.5em} 
\begin{itemize}
    \item We construct \textbf{EEmoDB}, the largest AICA instruction dataset to date, enriched via automated label refinement and corpus expansion. It features two specialized subsets tailored for QA and fine-grained assessment.
    \item We propose \textbf{EEmo-Logic}, which leverages a two-stage training strategy and task-customized GRPO to advance the model from basic emotion QA to fine-grained understanding of complex emotion attributes.
    \item Experiments confirm that EEmo-Logic establishes state-of-the-art results on in-domain tasks while maintaining robust cross-domain performance, validating its effectiveness as a comprehensive AICA solution.
\end{itemize}

\begin{figure*}[t] 
    \vspace{-4pt}
    \centering
    \includegraphics[width=\textwidth]{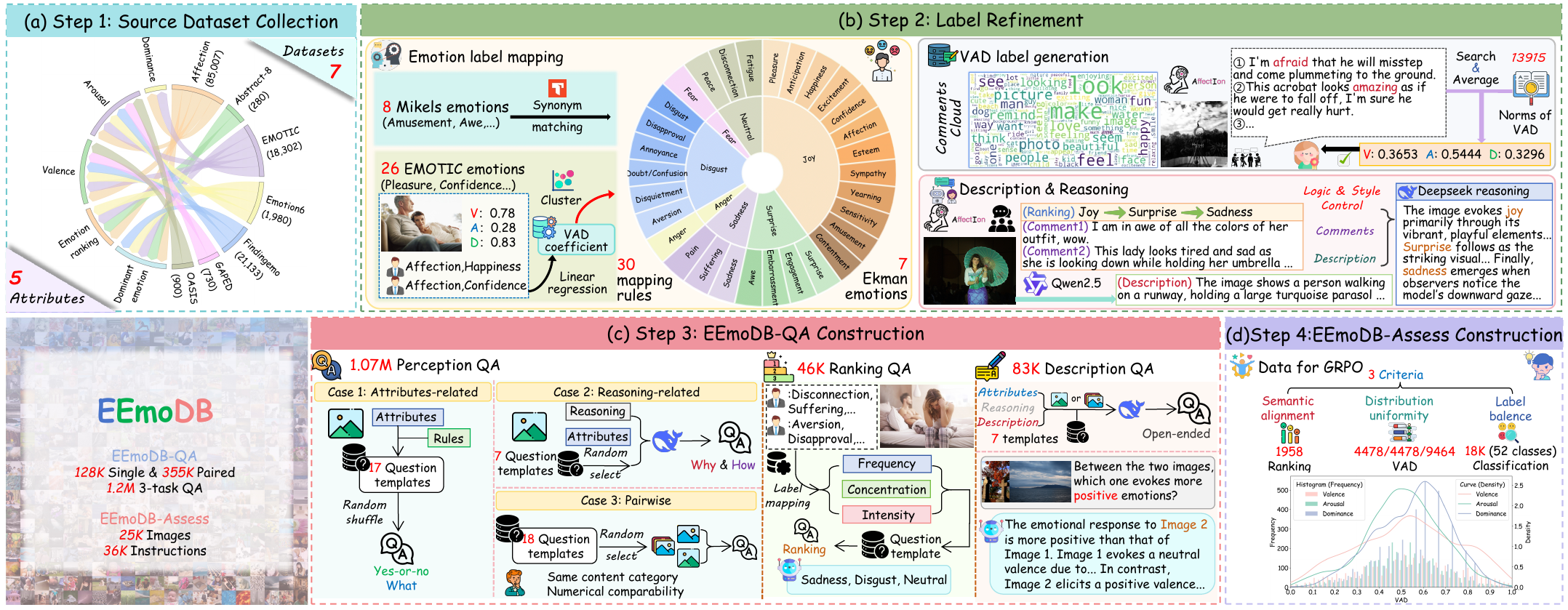}
    \caption{EEmoDB construction pipeline. (a) Selects source datasets based on $5$ dimensions. (b) Refines emotional labels via $3$ distinct processes. (c) Compiles EEmoDB-QA subset for perception, ranking, and description tasks using rule-based and model-assisted methods. (d) Curates EEmoDB-Assess subset to support GRPO training for fine-grained ranking, VAD scoring, and DEC classification.}
    \label{fig: dataset construction}
    \vspace{-0.5cm}
\end{figure*}

\vspace{-0.3cm}
\section{Related Works} \label{sec:related works}

\subsection{Datasets for AICA}

Traditional AICA datasets generally adhere to either DES or CES paradigms. DES datasets like OASIS \cite{OASIS} and GAPED \cite{GAPED} utilize VA \cite{VA} or VAD models to quantify affective attributes, while CES datasets such as Artphoto \cite{Artphoto} and EmoSet \cite{EmoSet} focus on dominant emotion classification. To capture subjectivity, FlickrLDL \cite{LDL_1} employs probability distributions, whereas Affection \cite{Affection} and ArtEmis \cite{ArtEmis} utilize interpretive captions. Some benchmarks, including EMOTIC \cite{EMOTIC} and Emotion6 \cite{Emotion6}, integrate both paradigms. However, the rise of MLLMs demands reasoning-oriented QA data that transcends static labels. While QA datasets like AesBench \cite{AesBench} and UNIAA \cite{UNIAA} focus primarily on aesthetic emotions, and EEmo-Bench \cite{EEmobench} is restricted by its scale to evaluation, a comprehensive training corpus remains absent. Consequently, we introduce \textbf{EEmoDB}, the largest AICA instruction dataset, leveraging large-scale emotion QA and fine-grained assessment to endow MLLMs with computational empathy and deep understanding.

\subsection{Models for AICA}

Traditional AICA methods handle DES via regression \cite{VAD_regression_1} or probability modeling \cite{VAD_regression_2}, and handle CES via discriminative networks for dominant emotion or distribution recognition \cite{dominant_1,PCNN}. However, these approaches lack reasoning capabilities and fail to unify these two paradigms.
Recently, MLLMs have adopted CoT strategies for emotion perception. 
EmoVIT \cite{EmoVIT} uses SFT but neglects emotional diversity, while AffectGPT \cite{AffectGPT}, rooted in facial expression recognition, lists emotions from an OV lexicon but struggles with intensity nuances and AICA domain generalization.
Although EmoCaliber \cite{EmoCaliber} utilizes confidence calibration, it disregards DES analysis.
To bridge this gap, we propose \textbf{EEmo-Logic}, a unified framework that leverages SFT to handle $3$ core emotion QA tasks while adapting the GRPO mechanism to facilitate fine-grained CES and DES assessment \cite{GRPO}. Results confirm that EEmo-Logic establishes robust performance on both in-domain and cross-domain datasets.

\begin{figure*}[] 
    \centering
    \vspace{-4pt}
    \begin{subfigure}[b]{0.24\textwidth}
        \centering
        \includegraphics[width=\linewidth]{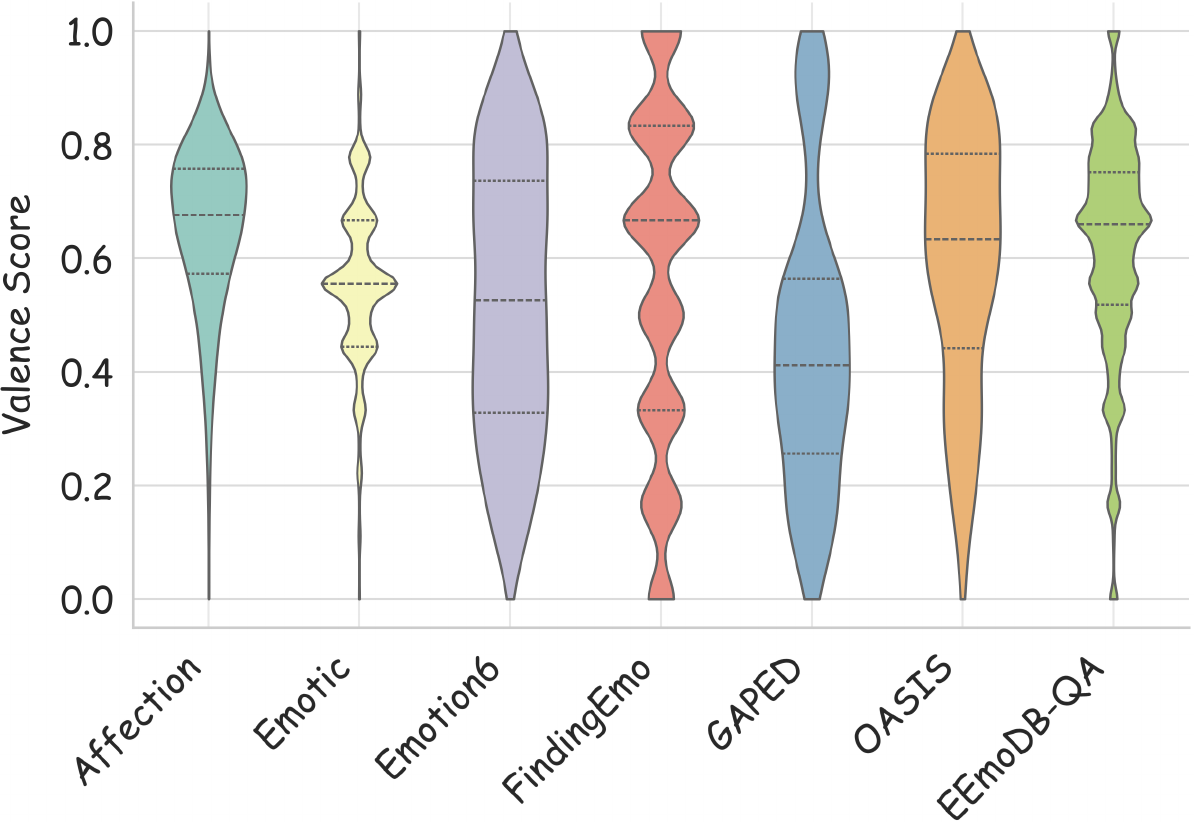}
        \caption{Valence}
        \label{fig:valence}
    \end{subfigure}
    \hfill 
    \begin{subfigure}[b]{0.24\textwidth}
        \centering
        \includegraphics[width=\linewidth]{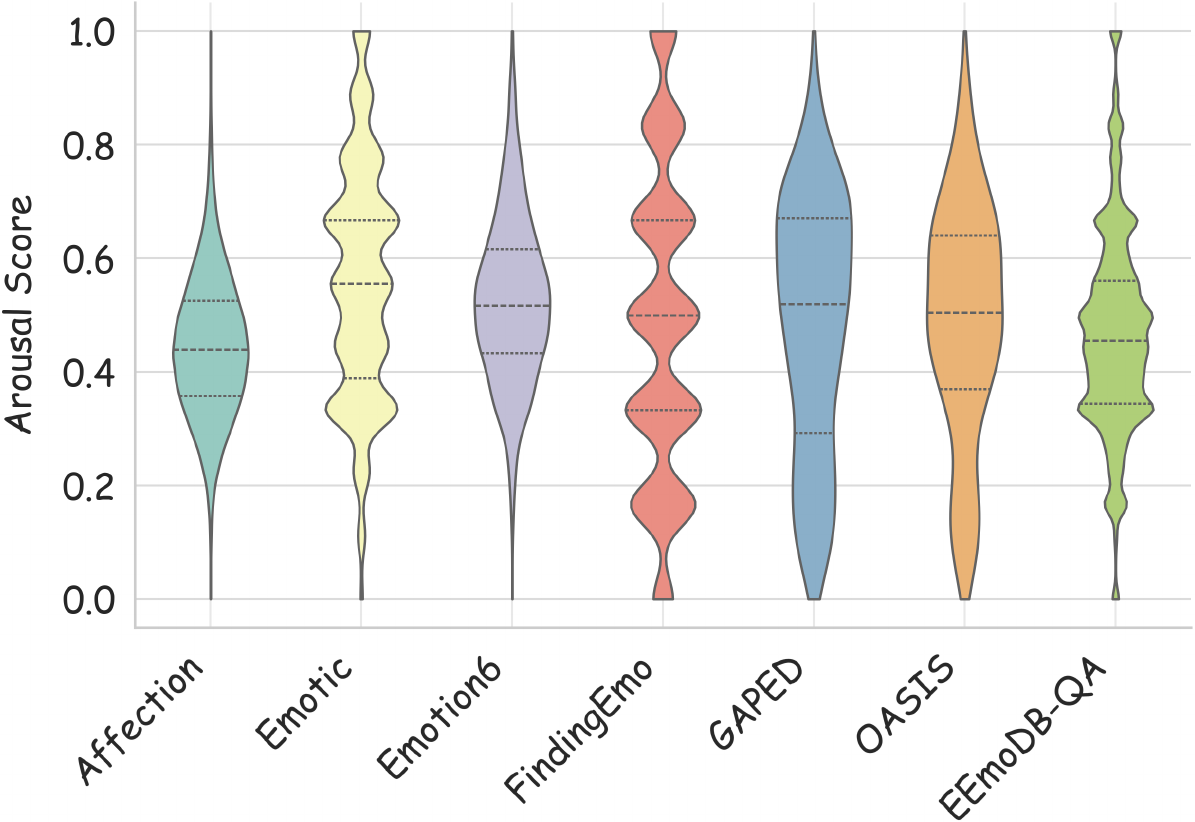}
        \caption{Arousal}
        \label{fig:arousal}
    \end{subfigure}
    \hfill 
    \begin{subfigure}[b]{0.12\textwidth}
        \centering
        \includegraphics[width=\linewidth]{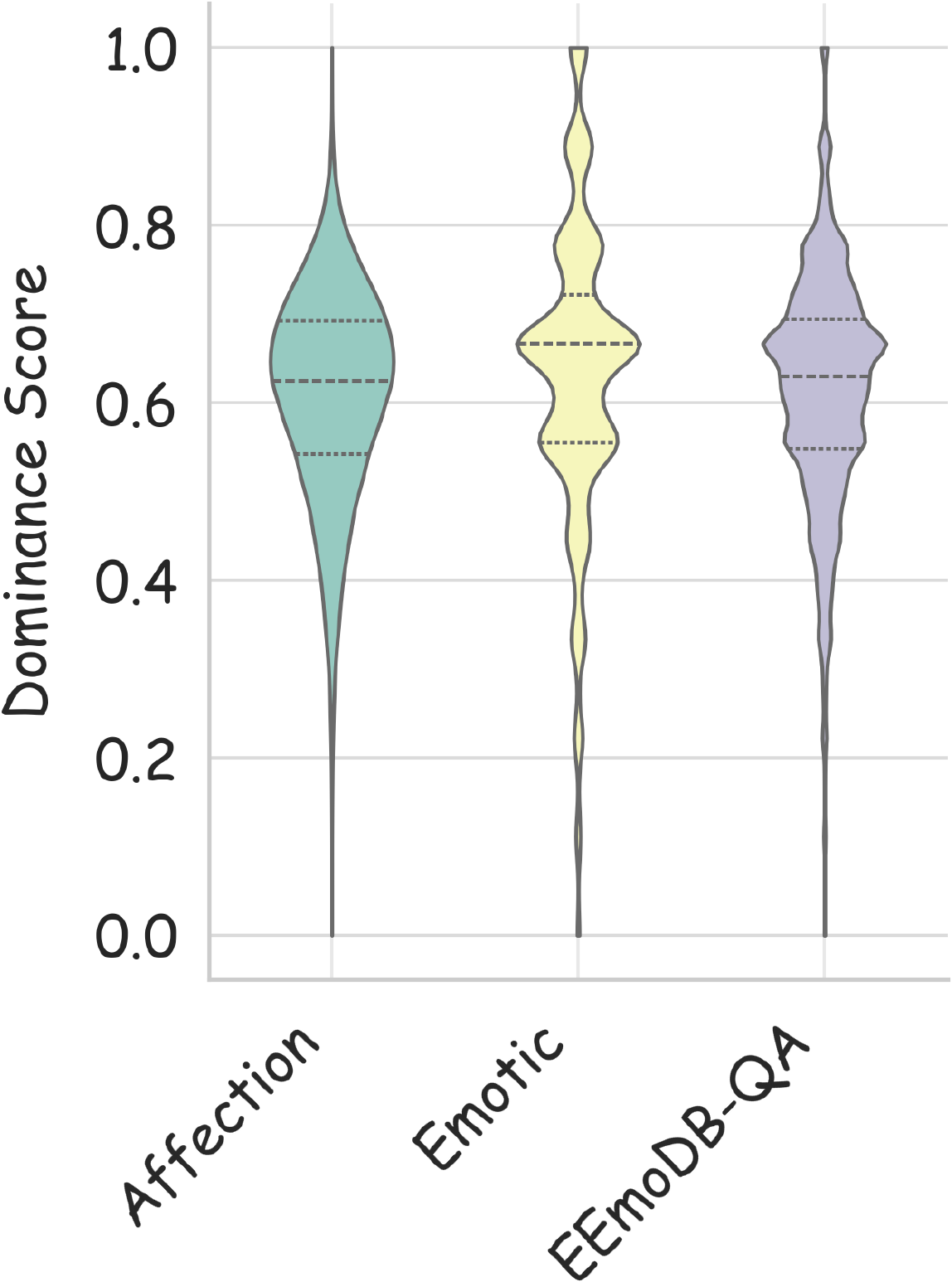}
        \caption{Dominance}
        \label{fig:dominance}
    \end{subfigure}
    \hfill 
    \begin{subfigure}[b]{0.36\textwidth}
        \centering
        \includegraphics[width=\linewidth]{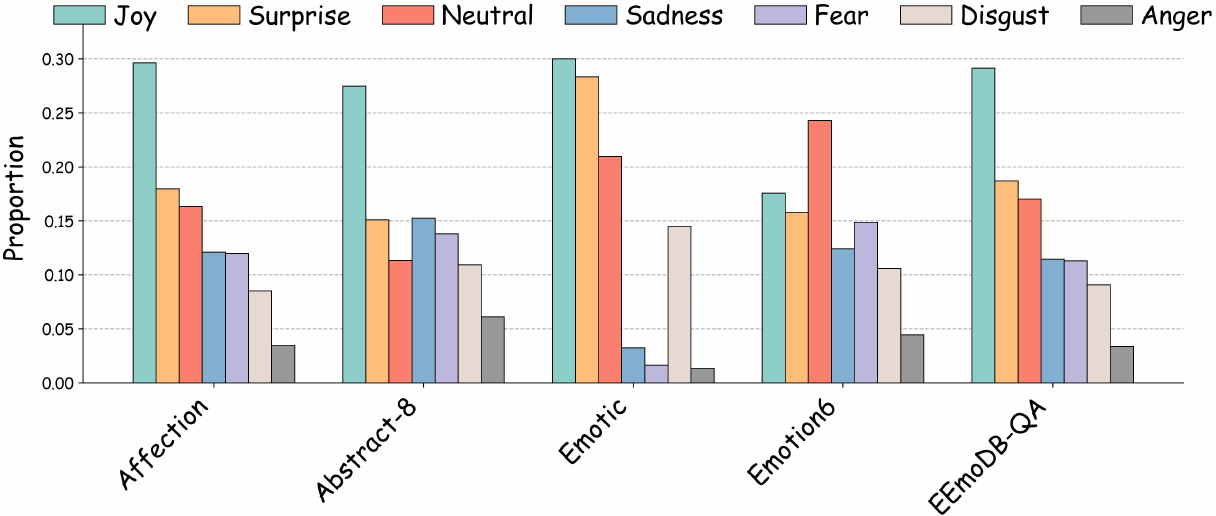}
        \caption{Emotion Freqency} 
        \label{fig:emotion}
    \end{subfigure}
    
    \caption{Comparative distributions across the dataset. (a)-(c) show the score distributions for valence, arousal, and dominance, respectively. (d) illustrates the overall emotional frequency statistics in the EEmoDB-QA subset.}
    \label{fig: distribution}
    \vspace{-0.5cm}
\end{figure*}

\section{Dataset Construction} \label{sec:dataset construction}

The proposed EEmoDB is structured into two components, with EEmoDB-QA designed for emotion QA and EEmoDB-Assess tailored for the fine-grained assessment of VAD, DEC, and ranking. This section details the automated construction pipeline for both subsets, including label mapping, generation, and filtering, as illustrated in Fig.~\ref{fig: dataset construction}.

\subsection{Source Dataset Collection}

To comprehensively characterize emotions, we utilize labels across $5$ dimensions spanning \textbf{dominant emotion} and \textbf{emotion ranking} at the CES level, as well as \textbf{valence}, \textbf{arousal}, and \textbf{dominance} at the DES level. We employ dominant emotion \cite{FI-1} for the primary response and incorporate top-$3$ emotion ranking \cite{EEmobench} to accommodate subjectivity and diversity, mitigating MLLMs' deficiencies in distribution generation \cite{not_distribution}. Additionally, intensity is quantified via the VAD model.

Unlike facial expression recognition, which targets subject-expressed emotions, AICA addresses viewer perception derived from scene composition and atmosphere. Accordingly, we aggregate $7$ datasets covering diverse imagery, including Affection \cite{Affection}, Abstract-8 \cite{Artphoto}, EMOTIC \cite{EMOTIC}, Emotion6 \cite{Emotion6}, FindingEmo \cite{Findingemo}, GAPED \cite{GAPED}, and OASIS \cite{OASIS}. As illustrated in Fig.~\ref{fig: dataset construction}(a), we leverage these datasets to capture diversity and intensity, categorizing tasks based on source annotation characteristics.

\subsection{Label Refinement} \label{sec: refine}

\textbf{Emotion Label Mapping}.
Integrating heterogeneous emotion label spaces introduces semantic ambiguity and overlap, particularly when distinguishing intensities (such as ``rage" versus ``anger"). To mitigate this, we adopt the Emotion6 \cite{Emotion6} schema, comprising Ekman's six basic emotions \cite{Ekman} and ``neutral". This ensures 1) \textbf{distinctness}, via mutually exclusive categories and a ``neutral" option for ambiguous cases; and 2) \textbf{compatibility}, facilitating the downward mapping of fine-grained labels.

We employ two mapping strategies:
1) \textbf{Synonym mapping}. Adopting the method of IESN \cite{synonym}, we link fine-grained labels to basic emotions via semantic synonyms from an online thesaurus\footnote{\href{https://www.thesaurus.com}{https://www.thesaurus.com}}. This allows us to map Mikels' model \cite{Mikels} directly to Ekman's categories.
2) \textbf{VAD-based emotion clustering}. For datasets containing both CES and DES information, we leverage the continuous spatial nature of discrete emotions \cite{emotion_circle,cluster}. Using linear regression with $10$-fold cross-validation, we map $26$ EMOTIC categories to the VAD space. We then utilize Ekman-aligned categories as anchor points to optimize a clustering radius, effectively grouping fine-grained emotions into the $7$ basic classes (see Appendix \ref{supp: cluster}). Finally, we conduct manual validation to ensure alignment with human perception. 

\textbf{VAD Label Generation}.
To expand the VAD label space for our dataset, we also employ the IESN methodology. Leveraging user comments from the Affection dataset \cite{Affection}, we synthesize ground truth by averaging the VAD scores of extracted keywords, derived from norms of 13,915 English lemmas \cite{VAD_norm}. These scores are subsequently normalized to generate the distribution shown in Fig.~\ref{fig: distribution}.
Validation via human expert sampling demonstrates that the generated results maintain high consistency with human perception (see Appendix \ref{supp: vad generation} for details).

\textbf{Description and Reasoning Generation}.
Despite the challenges in multimodal emotion recognition \cite{Emotionqwen}, MLLMs excel at text summarization and logical organization \cite{COT,deepseek}, as well as visual content description \cite{Qwen2.5}. Leveraging the rich expert comments in the Affection dataset, we construct a pipeline that limits model outputs to strict information synthesis. After selecting samples with distinct emotion intensities (Sec.~\ref{sec: QA}), \textit{Qwen2.5-VL} provides comprehensive scene descriptions. Subsequently, \textit{DeepSeek} is tasked with structuring these visual details alongside expert comments under rigid reasoning rules. This approach prevents model-induced inference hallucinations by ensuring all reasoning derives directly from human expert annotations, thereby guaranteeing accuracy and rationality.
(Detailed prompts and the verification process are provided in Appendix~\ref{supp sec: dataset construction}.)

\begin{figure*}[] 
    \centering
    \vspace{-4pt}
    \includegraphics[width=\textwidth]{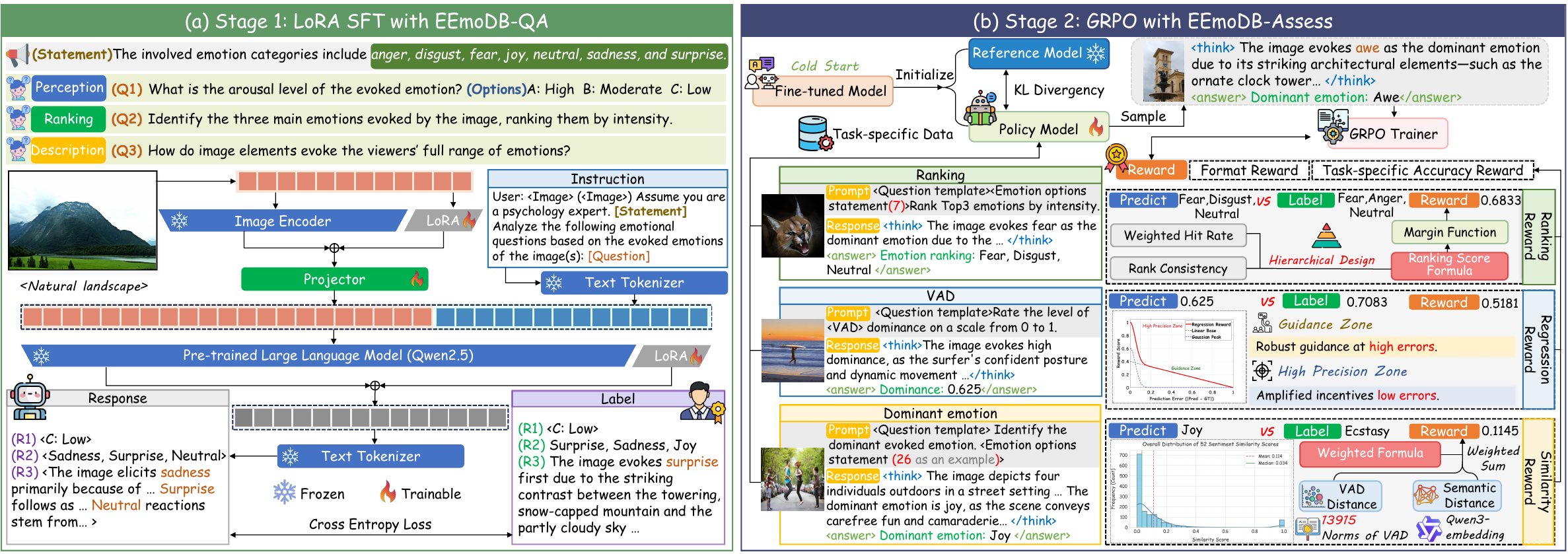}
    \caption{The EEmo-Logic training framework. The pipeline comprises two stages: 1) Supervised LoRA SFT on QA data for foundational QA capability; and 2) GRPO training on curated emotion labels to refine emotion assessment and elicit reasoning.}
    \label{fig: algorithm architecture}
    \vspace{-0.5cm}
\end{figure*}

\subsection{EEmoDB-QA Subset Construction} \label{sec: QA}

Leveraging refined emotion labels, we construct a large-scale QA instruction dataset via an automated pipeline to establish foundational emotional comprehension. As illustrated in Fig.~\ref{fig: dataset construction}(c), this workflow proceeds through $3$ distinct stages.
1) \textbf{Perception}: For coarse-grained understanding, we utilize $17$ templates across $5$ label dimensions, maintaining balance via rule-based selection. 
For reasoning-oriented queries, we instruct \textit{DeepSeek} to generate structured candidate answers strictly grounded in validated reasoning texts, thereby eliminating hallucinations and ensuring expert alignment.
For paired-image QA, we apply content and label constraints to $18$ base templates, guaranteeing semantic compatibility between textual queries and visual inputs.
2) \textbf{Ranking}: To model emotional diversity and hierarchy, we extract ranking labels from source datasets. EEmoDB determines the final hierarchy by sequentially stratifying $3$ metrics: selection frequency, emotional concentration, and intensity order (details in Appendix \ref{supp: ranking method}).
3) \textbf{Description}: This component facilitates open-ended inquiry. By integrating basic labels, visual descriptions, and reasoning texts, we prompt \textit{DeepSeek} to generate QA pairs across $7$ categories, strictly governing reasoning logic and style.
EEmoDB-QA comprises $128K$ single and $355K$ paired samples, totaling $1.2M$ instructions. Fig.~\ref{fig: distribution} illustrates the comparative analysis between EEmoDB-QA and the source datasets.

\subsection{EEmoDB-Assess Subset Construction} \label{sec: SR}

While EEmoDB-QA prioritizes scale, EEmoDB-Assess focuses on precision for fine-grained assessment. Targeting $5$ emotional dimensions across $3$ tasks, this curated corpus enforces strict label accuracy and distributional balance to minimize noise from label refinement.
1) \textbf{Emotion ranking}: Prioritizing semantic alignment, we utilize the Emotion6 \cite{Emotion6} dataset (based on a $7$-emotion taxonomy). We derive unambiguous hierarchical labels by sorting the inherent probability distributions, establishing explicit ranking ground truths.
2) \textbf{VAD scoring}: We augment expert annotations from $5$ datasets with large-scale synthetic VAD samples from the Affection dataset to maintain balanced score distributions for robust learning \cite{uniform_distribution}, thereby achieving the uniform distribution shown in Fig.~\ref{fig: dataset construction}(d).
3) \textbf{Dominant emotion classification}: We retain original DEC labels for classification, while providing specific candidate sets to prevent interference across heterogeneous emotion taxonomies. To mitigate long-tail imbalance, we anchor minority classes and constrain majority class upper bounds, ensuring a balanced label distribution.
Collectively, EEmoDB-Assess comprises $36K$ instruction pairs derived from $25K$ images.

\section{The EEmo-Logic} \label{sec: algorithm}

Building upon EEmoDB, we propose \textbf{EEmo-Logic}, an \textbf{all-in-one} architecture designed for comprehensive evoked emotion understanding. Adopting a two-stage paradigm (Fig.~\ref{fig: algorithm architecture}), Stage $1$ performs LoRA fine-tuning on EEmoDB-QA to instill foundational comprehension. Stage $2$ then applies GRPO on EEmoDB-Assess, refining assessment precision and enabling robust causal emotional reasoning.

\subsection{SFT for Broad Comprehension}

Studies \cite{InstructGPT} show that SFT on large-scale QA corpora significantly improves performance and transferability. Following \cite{EEmobench}, we select \textit{Qwen2.5-VL-7B} \cite{Qwen2.5} as the backbone for its superior affective comprehension. Stage $1$ fine-tunes the model on perception, ranking, and description tasks to build foundational emotional comprehension. Since data emotional bias makes full-parameter tuning prone to overfitting \cite{overfit}, we utilize LoRA \cite{lora} as an efficient implicit regularizer. Specifically, we apply LoRA to both the image encoder and large language model (LLM) layers, optimizing parameters via cross-entropy loss.

\subsection{GRPO for Fine-grained Assessment} \label{sec: grpo}

While recent studies \cite{COT-GRPO, AffectGPT-R1} necessitate a cold-start phase for GRPO to initialize thinking ability, our SFT process inherently establishes essential emotional and causal understanding. We initialize the policy model directly from the Stage $1$ checkpoint and apply GRPO to refine fine-grained perception and autonomous causal reasoning. Building on the standard GRPO framework \cite{GRPO} (details in Appendix \ref{supp: GRPO}), we optimize task-specific behaviors via novel reward functions. Our system integrates a format reward with $3$ hierarchical accuracy rewards tailored to each task. This approach drives simultaneous improvements across all objectives, yielding a unified architecture for the holistic interpretation of image-evoked emotions.
The rewards include: 

\textbf{Format Reward}.
To ensure structural integrity, we define a binary reward $\mathcal{R}_{\text{fmt}}^{(i)}$ that evaluates whether the response contains distinct thinking and answer components alongside the corresponding task identifier. The function returns $1$ for strict compliance and $0$ otherwise.

\textbf{Emotion Ranking Reward}.
To quantify alignment in the emotion ranking task, we leverage the scoring protocol from EEmo-Bench \cite{EEmobench}, prioritizing weighted emotion hit rate and ordinal consistency. To incentivize high precision, we incorporate a margin function that amplifies rewards as predictions converge toward the target. Let $E_{\text{gt}}=[e_1, e_2, e_3]$ and $E_{i}=[\hat{e}_1, \hat{e}_2, \hat{e}_3]$ denote the ground-truth ranking and the $i$-th response, respectively. The ranking reward $\mathcal{R}_{\text{rank}}^{(i)}$ is formulated as:

\vspace{-2em}
\begin{equation}
    \begin{split}
        \mathcal{R}_{\text{rank}}^{(i)}&= \mathcal{M}(\mathcal{N}(\sum_{k=1}^{3}\sum_{j=1}^{3}w_{k}\mathbb{I}(e_{k},\hat{e}_j)) \\
    &+\mathcal{W} (E_{GT},E_{i})\cdot \mathcal{N}(\mathcal{K} (\mathcal{SP} (E_{\text{gt}},E_{i}))),
    \end{split}
    \label{equ:rank}
\end{equation}
\vspace{-1.2em}

where $\mathbb{I}(e_{k},\hat{e}_j)$ denotes the indicator function, which equals $1$ if $\hat{e}_j$ matches the $k$-th ground truth $e_{k}$ and $0$ otherwise. This term is weighted by the positional values $w=\{5,3,2\}$ as defined in EEmo-Bench.
$\mathcal{N}(\cdot)$ normalizes the input to $[0,0.5]$.
$\mathcal{SP} (\cdot)$ extracts the intersection sequence while preserving relative order, which serves as the input for $\mathcal{K} (\cdot)$ (Kendall's $\tau$) \cite{KRCC} to evaluate ranking correlation.
The function $\mathcal{W} (\cdot)$ penalizes length discrepancies, assigning $\{1,\frac{1}{3},0\}$ for correct prediction counts of $\{3,2,1\}$, respectively.
Finally, $\mathcal{M}(\cdot)$ applies a squared margin function to maximize optimization efficacy.

\begin{table*}[]
\scriptsize
\centering
\renewcommand{\arraystretch}{1}   
\setlength{\tabcolsep}{3pt}        
\caption{Performance comparisons on the EEmo-Bench from perception, ranking, and description tasks. The best performance is \textbf{bolded} and the second performance is {\ul underlined.} NA denotes not applicable.}
\resizebox{\textwidth}{!}{%
\begin{tabular}{lcccccccccccc}
\hline
\multicolumn{1}{c|}{\textbf{Task}}                            & \multicolumn{7}{c|}{\textbf{Perception}}                                                                                                                                                                           & \multicolumn{1}{c|}{\textbf{Ranking}}                        & \multicolumn{3}{c|}{\textbf{Description}}                                                                            & \multirow{3}{*}{\textit{Overall↑}} \\ \cline{1-12}
\multicolumn{1}{c|}{\multirow{2}{*}{\textbf{Sub-categories}}} & \multicolumn{3}{c|}{\textit{Single Image}}                                           & \multicolumn{3}{c|}{\textit{Image Pair}}                                             & \multicolumn{1}{c|}{\multirow{2}{*}{\textit{overall↑}}} & \multicolumn{1}{c|}{\multirow{2}{*}{\begin{tabular}[c]{@{}c@{}}\textit{Emotion}\\ \textit{Score↑}\end{tabular}}} & \multirow{2}{*}{\begin{tabular}[c]{@{}c@{}}\textit{Single}\\ \textit{Image↑}\end{tabular}} & \multirow{2}{*}{\begin{tabular}[c]{@{}c@{}}\textit{Image}\\ \textit{Pair↑}\end{tabular}} & \multicolumn{1}{c|}{\multirow{2}{*}{\textit{Overall↑}}} &                          \\ \cline{2-7}
\multicolumn{1}{c|}{}                                & \textit{Yes-or-No↑ }      & \textit{What-How↑}          & \multicolumn{1}{c|}{\textit{Overall↑}}         & \textit{Yes-or-No↑}       & \textit{What-How↑}          & \multicolumn{1}{c|}{\textit{Overall↑}}         & \multicolumn{1}{c|}{}                         & \multicolumn{1}{c|}{}                               &                               &                             & \multicolumn{1}{c|}{}                         &                          \\ \hline
\multicolumn{1}{c|}{Random guess w/o open-ended}     & 50.00\%          & 33.33\%          & \multicolumn{1}{c|}{41.67\%}          & 50\%             & 34.26\%          & \multicolumn{1}{c|}{42.13\%}          & \multicolumn{1}{c|}{41.83\%}                  & \multicolumn{1}{c|}{25.48\%}                        & NA                       & NA                      & \multicolumn{1}{c|}{NA}                   & 33.65\%                  \\ \hline
\multicolumn{13}{l}{\textcolor{gray}{\textit{Medium-scale   open-source MLLMs}}}                                                                                                                                                                                                                                                                                                                                                                                                  \\ \hdashline 
\multicolumn{1}{l|}{mPLUG-Owl3-7B \cite{mplug3}}                   & 59.11\%          & 52.96\%          & \multicolumn{1}{c|}{56.04\%}          & 59.84\%          & 52.84\%          & \multicolumn{1}{c|}{56.34\%}          & \multicolumn{1}{c|}{56.14\%}                  & \multicolumn{1}{c|}{58.53\%}                        & 48.98\%                       & 53.50\%                     & \multicolumn{1}{c|}{50.67\%}                  & 55.11\%                  \\
\multicolumn{1}{l|}{Deepseek-VL-7B-chat \cite{deepseek-vl}}             & 60.18\%          & 50.14\%          & \multicolumn{1}{c|}{55.16\%}          & 52.88\%          & 38.71\%          & \multicolumn{1}{c|}{45.80\%}          & \multicolumn{1}{c|}{51.98\%}                  & \multicolumn{1}{c|}{57.12\%}                        & 43.95\%                       & 45.10\%                     & \multicolumn{1}{c|}{44.38\%}                  & 51.16\%                  \\
\multicolumn{1}{l|}{Janus-Pro-7B \cite{janus}}                    & 57.53\%          & 48.83\%          & \multicolumn{1}{c|}{53.18\%}          & 53.18\%          & 43.48\%          & \multicolumn{1}{c|}{48.33\%}          & \multicolumn{1}{c|}{51.53\%}                  & \multicolumn{1}{c|}{54.83\%}                        & 49.48\%                       & 46.97\%                     & \multicolumn{1}{c|}{48.54\%}                  & 51.63\%                  \\
\multicolumn{1}{l|}{LLaVA-NEXT-8B \cite{llava-next}}                   & 57.43\%          & 55.87\%          & \multicolumn{1}{c|}{56.65\%}          & 55.77\%          & 43.18\%          & \multicolumn{1}{c|}{49.48\%}          & \multicolumn{1}{c|}{54.22\%}                  & \multicolumn{1}{c|}{54.55\%}                        & 44.09\%                       & 31.37\%                     & \multicolumn{1}{c|}{39.33\%}                  & 49.37\%                  \\
\multicolumn{1}{l|}{LLaVA-OneVision-7B \cite{llava-ov}}              & 65.75\%          & 58.83\%          & \multicolumn{1}{c|}{62.29\%}          & 61.13\%          & 51.84\%          & \multicolumn{1}{c|}{56.49\%}          & \multicolumn{1}{c|}{60.32\%}                  & \multicolumn{1}{c|}{58.39\%}                        & 43.71\%                       & 47.00\%                     & \multicolumn{1}{c|}{44.94\%}                  & 54.55\%                  \\
\multicolumn{1}{l|}{LLaVA-OneVision-1.5-8B \cite{llava-ov-1.5}}          & 63.21\%          & 59.95\%          & \multicolumn{1}{c|}{61.58\%}          & 60.44\%          & 44.93\%          & \multicolumn{1}{c|}{52.69\%}          & \multicolumn{1}{c|}{58.56\%}                  & \multicolumn{1}{c|}{58.78\%}                        & 50.74\%                       & 44.43\%                     & \multicolumn{1}{c|}{48.38\%}                  & 55.24\%                  \\
\multicolumn{1}{l|}{InternVL3.5-8B \cite{internvl3.5}}                 & 60.15\%          & 54.37\%          & \multicolumn{1}{c|}{57.26\%}          & 55.57\%          & 50.40\%          & \multicolumn{1}{c|}{52.99\%}          & \multicolumn{1}{c|}{55.81\%}                  & \multicolumn{1}{c|}{61.00\%}                        & 44.39\%                       & 43.43\%                     & \multicolumn{1}{c|}{44.03\%}                  & 53.61\%                  \\
\multicolumn{1}{l|}{Qwen2-VL-7B \cite{qwen2}}                     & 59.98\%          & 59.39\%          & \multicolumn{1}{c|}{59.69\%}          & 54.97\%          & 54.43\%          & \multicolumn{1}{c|}{54.70\%}          & \multicolumn{1}{c|}{57.99\%}                  & \multicolumn{1}{c|}{61.36\%}                        & 48.42\%                       & 49.43\%                     & \multicolumn{1}{c|}{48.80\%}                  & 56.05\%                  \\
\multicolumn{1}{l|}{Qwen2.5-VL-7B \cite{Qwen2.5}}                   & 64.37\%          & 57.24\%          & \multicolumn{1}{c|}{60.81\%}          & 59.24\%          & 54.93\%          & \multicolumn{1}{c|}{57.09\%}          & \multicolumn{1}{c|}{59.54\%}                  & \multicolumn{1}{c|}{61.88\%}                        & 45.69\%                       & 45.73\%                     & \multicolumn{1}{c|}{45.70\%}                  & 55.71\%                  \\
\multicolumn{1}{l|}{Qwen3-VL-8B \cite{qwen3}}                     & 66.80\%          & 59.55\%          & \multicolumn{1}{c|}{63.18\%}          & 60.22\%          & 55.69\%          & \multicolumn{1}{c|}{57.96\%}          & \multicolumn{1}{c|}{61.41\%}                  & \multicolumn{1}{c|}{61.81\%}                        & 43.05\%                       & 42.43\%                     & \multicolumn{1}{c|}{42.82\%}                  & 55.34\%                  \\ \hdashline 
\multicolumn{13}{l}{\textcolor{gray}{\textit{Large-scale open-source MLLMs}}}                                                                                                                                                                                                                                                                                                                                                                                                     \\ \hdashline 
\multicolumn{1}{l|}{LLaVA-OneVision-72B \cite{llava-ov}}             & 66.36\%          & 64.64\%    & \multicolumn{1}{c|}{65.50\%}          & 65.01\%          & 58.11\%          & \multicolumn{1}{c|}{61.56\%}          & \multicolumn{1}{c|}{64.16\%}                  & \multicolumn{1}{c|}{66.92\%}                        & 52.36\%                       & 56.60\%                     & \multicolumn{1}{c|}{53.95\%}                  & 61.68\%                  \\
\multicolumn{1}{l|}{Qwen2-VL-72B \cite{qwen2}}                    & 67.89\%          & 61.58\%          & \multicolumn{1}{c|}{64.74\%}          & 64.12\%          & 60.00\%          & \multicolumn{1}{c|}{62.06\%}          & \multicolumn{1}{c|}{63.83\%}                  & \multicolumn{1}{c|}{64.69\%}                        & 47.23\%                       & 48.10\%                     & \multicolumn{1}{c|}{47.56\%}                  & 58.69\%                  \\
\multicolumn{1}{l|}{Qwen2.5-VL-72B \cite{Qwen2.5}}                  & 67.08\%          & 61.28\%          & \multicolumn{1}{c|}{64.18\%}          & 63.92\%          & 59.80\%          & \multicolumn{1}{c|}{61.86\%}          & \multicolumn{1}{c|}{63.39\%}                  & \multicolumn{1}{c|}{{\ul 67.84\%}}                  & 45.33\%                       & 47.50\%                     & \multicolumn{1}{c|}{46.14\%}                  & 59.13\%                  \\ \hdashline 
\multicolumn{13}{l}{\textcolor{gray}{\textit{Proprietary   MLLMs}}}                                                                                                                                                                                                                                                                                                                                                                                                               \\ \hdashline 
\multicolumn{1}{l|}{Qwen-VL-Max \cite{qwen-vl}}                        & 66.51\%          & 60.96\%          & \multicolumn{1}{c|}{63.74\%}          & 63.59\%          & 58.84\%          & \multicolumn{1}{c|}{61.22\%}          & \multicolumn{1}{c|}{62.88\%}                  & \multicolumn{1}{c|}{67.27\%}                        & 43.24\%                       & 45.72\%                     & \multicolumn{1}{c|}{44.17\%}                  & 58.11\%                  \\
\multicolumn{1}{l|}{GPT-4o \cite{gpt4}}                          & 68.25\%          & {\ul 64.80\%}          & \multicolumn{1}{c|}{66.53\%}          & 64.41\%          & 61.49\%          & \multicolumn{1}{c|}{62.95\%}          & \multicolumn{1}{c|}{65.31\%}                  & \multicolumn{1}{c|}{65.67\%}                        & {\ul 55.97\%}                 & 56.57\%                     & \multicolumn{1}{c|}{56.19\%}                  & {\ul 62.39\%}            \\
\multicolumn{1}{l|}{GPT-5 \cite{gpt5}}                           & 67.40\%          & 60.05\%          & \multicolumn{1}{c|}{63.73\%}          & 63.46\%          & 61.96\%          & \multicolumn{1}{c|}{62.71\%}          & \multicolumn{1}{c|}{63.38\%}                  & \multicolumn{1}{c|}{61.71\%}                        & 55.21\%                       & {\ul 58.99\%}               & \multicolumn{1}{c|}{{\ul 56.63\%}}            & 60.57\%                  \\
\multicolumn{1}{l|}{Claude-3.7-Sonnet \cite{claude}}               & 64.08\%          & 59.04\%          & \multicolumn{1}{c|}{61.56\%}          & 61.93\%          & 61.49\%          & \multicolumn{1}{c|}{61.71\%}          & \multicolumn{1}{c|}{61.61\%}                  & \multicolumn{1}{c|}{67.05\%}                        & 51.58\%                       & 52.07\%                     & \multicolumn{1}{c|}{51.76\%}                  & 60.14\%                  \\
\multicolumn{1}{l|}{Gemini-2.5-Pro \cite{gemini2.5}}                  & \textbf{72.22\%} & 64.59\%          & \multicolumn{1}{c|}{{\ul 68.41\%}}    & 65.21\%          & {\ul 62.09\%}    & \multicolumn{1}{c|}{{\ul 63.65\%}}    & \multicolumn{1}{c|}{{\ul 66.79\%}}            & \multicolumn{1}{c|}{60.46\%}                        & 45.44\%                       & 42.03\%                     & \multicolumn{1}{c|}{44.16\%}                  & 57.14\%                  \\ \hdashline 
\multicolumn{13}{l}{\textcolor{gray}{\textit{Emotion-Oriented   MLLMs}}}                                                                                                                                                                                                                                                                                                                                                                                                          \\ \hdashline 
\multicolumn{1}{l|}{EmoVIT \cite{EmoVIT}}                          & 51.58\%          & 45.00\%          & \multicolumn{1}{c|}{48.29\%}          & 3.48\%           & 7.58\%           & \multicolumn{1}{c|}{5.53\%}           & \multicolumn{1}{c|}{33.77\%}                  & \multicolumn{1}{c|}{29.56\%}                        & 44.27\%                       & 32.70\%                     & \multicolumn{1}{c|}{39.94\%}                  & 34.42\%                  \\
\multicolumn{1}{l|}{AffectGPT \cite{AffectGPT}}                       &   49.13\%
               &    37.76\%
              & \multicolumn{1}{c|}{43.45\%}                 & 50.60\%
           &     37.97\%
             & \multicolumn{1}{c|}{44.29\%}                 & \multicolumn{1}{c|}{43.73\%}                         & \multicolumn{1}{c|}{18.57\%
}                               &           0.56\%
                    &         12.20\%
                    & \multicolumn{1}{c|}{4.92\%}                         &        22.41\%
                  \\
\multicolumn{1}{l|}{Emotion-Qwen \cite{Emotionqwen}}                    & 63.98\%          & 60.31\%          & \multicolumn{1}{c|}{62.15\%}          & 60.24\%          & 61.23\%          & \multicolumn{1}{c|}{60.74\%}          & \multicolumn{1}{c|}{61.67\%}                  & \multicolumn{1}{c|}{61.89\%}                        & 36.29\%                       & 36.67\%                     & \multicolumn{1}{c|}{36.43\%}                  & 53.33\%                  \\
\multicolumn{1}{l|}{R1-Omni-0.5B \cite{R1-Omni}}                         & 47.70\%           & 42.05\%        & \multicolumn{1}{c|}{44.88\%}        & 48.91\%        & 35.79\%        &  \multicolumn{1}{c|}{42.35\%}        & \multicolumn{1}{c|}{44.02\%}                & \multicolumn{1}{c|}{15.53\%}                      & 8.14\%                     & 31.93\%                   & \multicolumn{1}{c|}{17.05\%}                & 25.53\%                \\
\multicolumn{1}{l|}{EmoCaliber \cite{EmoCaliber}}                      & 67.19\%          & 56.99\%          & \multicolumn{1}{c|}{62.09\%}          & {\ul 65.51\%}    & 50.30\%          & \multicolumn{1}{c|}{57.91\%}          & \multicolumn{1}{c|}{60.67\%}                  & \multicolumn{1}{c|}{61.84\%}                        & 35.07\%                       & 29.67\%                     & \multicolumn{1}{c|}{33.05\%}                  & 51.85\%                  \\
\multicolumn{1}{l|}{\textbf{EEmo-Logic} (Ours)}                & {\ul 71.73\%}    & \textbf{68.37\%} & \multicolumn{1}{c|}{\textbf{70.05\%}} & \textbf{66.80\%} & \textbf{64.41\%} & \multicolumn{1}{c|}{\textbf{65.61\%}} & \multicolumn{1}{c|}{\textbf{68.54\%}}         & \multicolumn{1}{c|}{\textbf{67.97\%}}               & \textbf{59.02\%}              & \textbf{67.40\%}            & \multicolumn{1}{c|}{\textbf{62.16\%}}         & \textbf{66.22\%}         \\ \hline
\end{tabular}%
}
\label{tab: 3 tasks}
\vspace{-0.5cm}
\end{table*}

\begin{table}[t]
\centering
\scriptsize
\caption{Performance comparison on the EEmo-Bench Assessment task. Metrics: \textit{SRCC}↑ / \textit{PLCC}↑. }
\renewcommand{\arraystretch}{1.1}   
\setlength{\tabcolsep}{7.5pt}        
\resizebox{0.48\textwidth}{!}{%
\begin{tabular}{l|cccc}
\hline
\multicolumn{1}{c|}{\textbf{Dimension}} & \textbf{Valence}            & \textbf{Arousal}           & \textbf{Dominance}          & \textbf{Overall}            \\ \hline
Deepseek-VL-7B-chat            & 0.85/0.83    & 0.54/0.56         & -0.50/-0.53        & 0.30/0.29          \\
Janus-Pro-7B                   & 0.76/0.76          & 0.54/0.38         & -0.30/-0.23        & 0.33/0.30          \\
LLaVA-OneVision-7B             & 0.67/0.47          & 0.24/0.28         & -0.19/-0.17        & 0.24/0.19          \\
LLaVA-NEXT-8B                  & 0.58/0.57          & 0.38/0.37         & -0.07/-0.05        & 0.30/0.30          \\
mPLUG-Owl3-7B                  & 0.85/0.86    & 0.56/0.53         & -0.22/-0.07        & 0.40/0.44          \\
Qwen2-VL-7B                    & \textbf{0.87/0.88} & 0.59/0.57         & -0.04/-0.04        & 0.47/0.47          \\
Qwen2.5-VL-7B                  & 0.80/0.74          & 0.46/0.46         & 0.32/0.31          & 0.53/0.50          \\
Qwen3-VL-8B                    & 0.84/0.82          & 0.66/0.62   & -0.26/-0.15        & 0.41/0.43          \\
\textbf{EEmo-Logic} (Probability-based)  & 0.85/0.84    & \textbf{0.71/0.70} & 0.46/0.41    & 0.67/0.65    \\
\textbf{EEmo-Logic} (Think)    & 0.84/0.83          & 0.57/0.56         & \textbf{0.78/0.74} & \textbf{0.73/0.71} \\ \hline
\end{tabular}%
}
\label{tab: assessment}
\vspace{-0.5cm}
\end{table}

\begin{table*}[]
\scriptsize
\centering
\renewcommand\arraystretch{1.1}
\renewcommand\tabcolsep{4pt}
\caption{Performance comparison on cross-domain emotion datasets, covering dominant evoked emotion recognition (Artphoto, ArtEmis) and emotion-related aesthetic assessment benchmark subsets (UNIAA-Sent., AesBench-AesE).}
\resizebox{\textwidth}{!}{%
\begin{tabular}{l|cc|cc|c|ccccccccc}
\hline
\multicolumn{1}{c|}{\multirow{3}{*}{\begin{tabular}[c]{@{}c@{}}\textbf{Cross-Domain}\\ \textbf{Dataset}\end{tabular}}} & \multicolumn{2}{c|}{\textbf{Artphoto}}                & \multicolumn{2}{c|}{\textbf{ArtEmis}}                 & \textbf{UNIAA Sent.}               & \multicolumn{9}{c}{\textbf{AesBench   AesE}}                                                                                                                                                                                         \\ \cline{2-15} 
\multicolumn{1}{c|}{}                                                                                & \multirow{2}{*}{\textit{F1↑}} & \multirow{2}{*}{\textit{ACC↑}} & \multirow{2}{*}{\textit{F1↑}} & \multirow{2}{*}{\textit{ACC↑}} & \multirow{2}{*}{\textit{Overall↑}} & \multicolumn{4}{c|}{\textit{Empathy Dimensions}}                                                        & \multicolumn{4}{c|}{\textit{Question Types}}                                                            & \multirow{2}{*}{\textit{overall↑}} \\ \cline{7-14}
\multicolumn{1}{c|}{}                                                                                &                      &                       &                      &                       &                           & \textit{Emotion↑}         & \textit{Interest↑}        & \textit{Uniqueness↑}      & \multicolumn{1}{c|}{\textit{Vibe↑}}            & \textit{Yes-No↑}          & \textit{What↑}            & \textit{How↑}             & \multicolumn{1}{c|}{\textit{Why↑}}             &                           \\ \hline
EmoVIT                                                                                               & 35.53\%              & 39.33\%               & 12.73\%              & 12.98\%               & 1.94\%                    & 29.86\%          & 35.48\%          & 34.48\%          & \multicolumn{1}{c|}{41.93\%}          & 3.35\%           & 33.85\%          & 56.29\%          & \multicolumn{1}{c|}{70.10\%}          & 34.00\%                   \\
AffectGPT                                                                                            & 8.76\%               & 10.17\%               & 8.73\%               & 9.53\%                & 31.68\%                   & 44.99\%          & 80.65\%          & 44.83\%          & \multicolumn{1}{c|}{53.69\%}          & 51.83\%          & 34.84\%          & 45.00\%          & \multicolumn{1}{c|}{68.63\%}          & 48.29\%                   \\
Emotion-qwen                                                                                         & 37.53\%              & 38.33\%               & 25.34\%              & 29.01\%               & \textbf{79.53\%}          & {\ul 71.02\%}    & 80.65\%          & {\ul 68.97\%}    & \multicolumn{1}{c|}{{\ul 81.93\%}}    & \textbf{74.65\%} & 66.86\%          & {\ul 75.57\%}    & \multicolumn{1}{c|}{87.25\%}          & {\ul 74.75\%}             \\
R1-Omni-0.5B                                                                                         & 10.45\%              & 11.91\%               & 4.30\%               & 3.07\%                & 51.08\%                   & 37.23\%          & 41.94\%          & 37.93\%          & \multicolumn{1}{c|}{48.66\%}          & 49.09\%          & 39.80\%          & 36.71\%          & \multicolumn{1}{c|}{31.62\%}          & 41.11\%                   \\
EmoCaliber                                                                                           & \textbf{39.37\%}     & {\ul 40.32\%}         & {\ul 30.59\%}        & {\ul 33.23\%}         & {\ul 76.72\%}             & 70.97\%          & {\ul 83.87\%}    & \textbf{72.41\%} & \multicolumn{1}{c|}{81.07\%}          & {\ul 73.23\%}    & {\ul 68.27\%}    & 74.57\%          & \multicolumn{1}{c|}{{\ul 88.24\%}}    & 74.50\%                   \\
EEmo-Logic (Ours)                                                                                    & {\ul 38.68\%}        & \textbf{42.43\%}      & \textbf{31.58\%}     & \textbf{35.10\%}      & 75.65\%                   & \textbf{72.30\%} & \textbf{87.10\%} & 65.52\%          & \multicolumn{1}{c|}{\textbf{82.14\%}} & 72.92\%          & \textbf{69.97\%} & \textbf{77.00\%} & \multicolumn{1}{c|}{\textbf{89.95\%}} & \textbf{75.68\%}          \\ \hline
\end{tabular}%
}
\label{tab: OOD}
\vspace{-0.3cm}
\end{table*}

\textbf{Hybrid Regression Reward}.
While applying GRPO to VAD regression remains unexplored, parallels exist in mean opinion score (MOS) prediction. Conventional strategies typically employ threshold-based mechanisms, such as binary \cite{Q-Insight} or bell-shaped rewards \cite{VQAThinker} or distance-based $\ell_{1}$-norm-based rewards \cite{VQ-Insight}. 
However, the former often suffer from sparse supervision under significant errors, whereas the latter exhibit diminishing sensitivity during fine-grained convergence. To navigate these trade-offs, we engineer a hybrid regression reward $\mathcal{R}_{reg}^{(i)}$ for the $i$-th response, designed to synergize global optimization stability with local high-precision refinement. The formulation is defined as:

\vspace{-1.5em}
\begin{equation}
    \mathcal{R}_{\mathrm{reg}}^{(i)}=\lambda_{\mathrm{base}}\cdot\max(0,1-\delta)+\lambda_{\mathrm{peak}}\cdot\exp\left(-\frac{\delta^2}{2\sigma^2}\right),
    \label{equ:reg}
\end{equation}
\vspace{-1.5em}

where $\delta=|s-\hat{s}_{i}|$ is the absolute deviation of prediction $\hat{s}_{i}$ from ground truth $s$. This formulation creates a funnel-shaped landscape comprising two zones: 
1) A guidance zone via linear decay $\lambda_{\mathrm{base}}$, offering continuous signals in high-error regions to prevent stagnation.
2) A high precision zone via the gaussian kernel $(\lambda_{\mathrm{peak}},\sigma)$, which amplifies rewards to incentivize fine-grained calibration during convergence. This mechanism steers the policy from coarse alignment toward precise matching (see Fig.~\ref{fig: algorithm architecture}(b)).

\textbf{Emotion Similarity Reward}.
Dominant emotion recognition is pivotal in affective computing. However, current GRPO-based methods typically rely on binary rewards \cite{AffectGPT-R1,R1-Omni}, leading to sparse supervision in large label spaces. While EmoCaliber \cite{EmoCaliber} attempts to mitigate this via VAD mapping, it remains constrained by limited dimensional expressiveness. To densify feedback signals, we propose a dual-perspective similarity reward that fuses VAD vectors with semantic spaces \cite{similarity_score}, a process structured into $3$ distinct phases.
1) Label set alignment: To resolve semantic conflicts across datasets, we constrain training to dataset-specific emotion sets and pre-compute the corresponding intra-set similarity matrices.
2) VAD similarity $S_{vad}^{(i)}$: This metric is derived from pairwise weighted Euclidean distances between VAD anchors.
3) Semantic similarity $S_{emb}^{(i)}$: We append detailed semantic descriptions to each label and encode them via \textit{Qwen3-embedding} \cite{Qwen3-embedding}. 
The results are normalized by dividing by $\mu_{\max}$, the global maximum of inter-emotion similarity, to enhance discriminability.
Consequently, we fuse these metrics to define the composite reward $\mathcal{R}_{sim}^{(i)}$ of the $i$-th response:

\vspace{-1em}
\begin{equation}
    \mathcal{R}_{\text{sim}}^{(i)} = \left[ \lambda_{\text{sim}} S_{\text{vad}}^{(i)} + (1 - \lambda_{\text{sim}}) \left( \frac{S_{\text{emb}}^{(i)}}{\mu_{\max}} \right)^m \right]^p,
    \label{equ:sim}
\end{equation}
\vspace{-0.5em}

where $\lambda_{\text{sim}}$ balances the two modalities, while $m$ and $p$ regulate the reward landscape curvature to enhance discriminability.
The scores distribution is depicted in Fig.~\ref{fig: algorithm architecture}(b).

The final reward is calculated as:

\vspace{-1.5em}
\begin{equation}
    r_i=\lambda_0\cdot\mathcal{R}_{\text{fmt}}^{(i)}+\mathbbm{1}_{\text{rank}}\cdot
    \mathcal{R}_{\text{rank}}^{(i)} + \mathbbm{1}_{\text{reg}}\cdot\mathcal{R}_{\mathrm{reg}}^{(i)} + \mathbbm{1}_{\text{sim}}\cdot\mathcal{R}_{\text{sim}}^{(i)},
    \label{equ:total}
\end{equation}
\vspace{-1.5em}

where $\lambda_0$ scales the format reward, and $\mathbbm{1}_{\text{rank}}$ serves as a gate activating only for compliant ranking responses. Analogous gates $\mathbbm{1}_{\text{reg}}$ and $\mathbbm{1}_{\text{sim}}$ are employed for VAD scoring and dominant emotion classification, respectively.

\section{Experiments}

\subsection{Experimental Settings}

\textbf{Implementation Details}.
We implement the EEmo-Logic training pipeline in PyTorch using \textit{Qwen2.5-VL-7B} \cite{Qwen2.5}. In Stage 1, we fine-tune the model on EEmoDB-QA via LoRA on two $50GB$ $A6000$ GPUs, utilizing a batch size of $8$ and an initial learning rate of $2\times10^{-4}$ with cosine annealing. The LoRA configuration includes rank $r=64$, scaling factor $\alpha=64$, and a dropout rate of $0.05$. Subsequently, this SFT model serves as the cold-start policy for GRPO training on EEmoDB-Assess. 
For this phase, we employ three $140GB$ $H200$ GPUs with a generation group size of $8$ and a KL penalty coefficient $\beta=0.001$. The model is optimized for one epoch with a global batch size $N=128$ and a reduced learning rate of $1\times10^{-6}$.
Hyperparameters are set to $\lambda_{\text{base}}=0.4$ (Eq. \ref{equ:reg}) and $\lambda_{0}=0.2$ (Eq. \ref{equ:total}), with similarity parameters in Eq. \ref{equ:sim} fixed at $\lambda_{\text{sim}}=0.6$, $m=2$, and $p=3$.

\textbf{Compared Methods}.
We benchmark EEmo-Logic against $23$ baselines across $4$ categories, comprising $10$ medium-scale and $3$ large-scale open-source MLLMs, alongside $5$ proprietary MLLMs, and $5$ emotion-oriented MLLMs. Detailed introductions are provided in Appendix \ref{supp: models}.

\textbf{Benchmark Datasets}.
We evaluate EEmo-Logic across \textbf{in-domain} and \textbf{cross-domain} datasets. For the former, we utilize EEmo-Bench \cite{EEmobench}, which represents the first systematic benchmark for image-evoked emotion assessment. Here, we assess performance on $4$ core tasks, including perception, ranking, description, and assessment. For the latter, we examine generalization on the classic DEC recognition task using Artphoto \cite{Artphoto} and ArtEmis \cite{ArtEmis} datasets. Furthermore, acknowledging the correlation between evoked emotions and aesthetic empathy, we include the emotion-specific subsets of UNIAA \cite{UNIAA} and AesBench \cite{AesBench}. Notably, all models are evaluated under a \textbf{zero-shot} setting to ensure fairness.

\subsection{Evaluation on In-Domain Datasets}
As shown in Tabs.~\ref{tab: 3 tasks}, \ref{tab: assessment}, and Fig.~\ref{fig:spotlight}(a), EEmo-Logic achieves state-of-the-art performance across the $4$ tasks in EEmo-Bench. We elaborate on these results as follows:

\textbf{Perception Task}. 
 In terms of accuracy (ACC), EEmo-Logic surpasses competing models across most metrics, trailing only the advanced Gemini-2.5-Pro marginally in single-image judgment. Notably, large-scale QA training confers a $10\%$ improvement over the baseline on ``What-How" queries, underscoring its superior emotional sensitivity. Furthermore, results demonstrate the model's resilience to label noise arising from divergent definitions in source datasets. By leveraging the comprehensive task design in Stage $1$ and fine-grained attribute learning in Stage $2$, EEmo-Logic effectively mitigates these data imperfections, achieving robust performance on the benchmark validated by high-quality human annotations.

\textbf{Ranking Task}.
Emotion ranking measures the ability to discern intensity variations in co-occurring emotions. Adopting the EEmo-Bench \cite{EEmobench} protocol (see Appendix \ref{supp: ranking evaluation}), our two-stage training yields a $6.09\%$ improvement over the baseline, establishing a new state-of-the-art. Analysis indicates that existing models plateau around $60\%$; while they identify primary emotions, they lack the sensitivity to resolve subtle intensity gradients, often resulting in rank reversals. Our GRPO-based approach effectively surmounts this non-trivial bottleneck, capturing fine-grained nuances and validating the efficacy of our strategy.

\begin{table*}[]
\centering
\scriptsize
\setlength{\tabcolsep}{3pt}    
\renewcommand{\arraystretch}{1}   
\caption{Ablation studies on fine-tuning strategies, training stages, and GRPO tasks.}
\resizebox{\textwidth}{!}{%
\begin{tabular}{cccccc|c|c|ccc|c|c|c|c|c}
\hline
 & \multicolumn{5}{c|}{\textbf{Training Strategy}}                                                                                                                                                                                                                                              & \textbf{Perception} & \textbf{Description} & \multicolumn{3}{c|}{\textbf{Assessment}}                                                                                                                                    & \textbf{Ranking}                                                 & \textbf{Artphoto}        & \textbf{ArtEmis}         & \textbf{UNIAA}   & \textbf{AesBench} \\ \cdashline{2-16}
 & \multicolumn{2}{c|}{Stage 1}                                                                       & \multicolumn{3}{c|}{Stage 2}                                                                                                                                                   & Overall    & Overall     & Valence                                              & Arousal                                              & Dominance                                            & Overall                                                 & Overall         & Overall         & Sent.   & AesE     \\ \cdashline{2-16}
No. & Full                      & \multicolumn{1}{c|}{LoRA}                                              & \begin{tabular}[c]{@{}c@{}}VAD\\ Scoring\end{tabular} & \begin{tabular}[c]{@{}c@{}}Emotion\\ Ranking\end{tabular} & \begin{tabular}[c]{@{}c@{}}Dominant\\ Emotion\end{tabular} & \textit{ACC↑}        & \begin{tabular}[c]{@{}c@{}}\textit{Description}\\ \textit{Score↑}\end{tabular}         & \begin{tabular}[c]{@{}c@{}}\textit{SRCC↑}\\ /\textit{PLCC↑}\end{tabular} & \begin{tabular}[c]{@{}c@{}}\textit{SRCC↑}\\ /\textit{PLCC↑}\end{tabular} & \begin{tabular}[c]{@{}c@{}}\textit{SRCC↑}\\ /\textit{PLCC↑}\end{tabular} & \begin{tabular}[c]{@{}c@{}}\textit{Ranking}\\ \textit{Score↑}\end{tabular} & \textit{F1↑/ACC↑}          & \textit{F1↑/ACC↑}          & \textit{ACC↑}     & \textit{ACC↑}      \\ \hline
(1) & \checkmark & \multicolumn{1}{c|}{}                                                  &                                                       &                                                           &                                                            & 64.31\%    & 62.16\%     & NA                                                   & NA                                                   & NA                                                   & 64.44\%                                                 & 33.59\%/33.50\% & 15.00\%/20.47\% & 71.98\% & 73.64\%  \\
                 (2)   &      & \multicolumn{1}{c|}{\checkmark}                         &                                                       &                                                           &                                                            & 68.02\%    & 65.45\%     & NA                                                   & NA                                                   & NA                                                   & 63.01\%                                                 & 35.67\%/37.60\% & 21.64\%/28.02\% & 72.84\% & 74.68\%  \\
                  (3)   &     & \multicolumn{1}{c|}{\checkmark}                         & \checkmark                             &                                                           &                                                            & 68.02\%    &   62.75\%          & \multicolumn{1}{c}{0.85/0.83}                                 & \multicolumn{1}{c}{0.55/0.55}                         & \multicolumn{1}{c|}{0.80/0.79}                                &                                          62.57\%               & 38.76\%/40.94\% & 25.58\%/29.84\% & 72.84\% & 74.93\%  \\
                     (4)    &         & \multicolumn{1}{c|}{\checkmark}                         & \checkmark                             & \checkmark                                 &                                                            & 67.73\%    & 59.76\%     & 0.85/0.85                                            & 0.56/0.55                                            & 0.79/0.77                                            & 66.83\%                                                 & 38.59\%/41.32\% & 30.05\%34.22\%  & 70.91\% & 73.89\%  \\
                     (5)    &         & \multicolumn{1}{c|}{}                                                  & \checkmark                             & \checkmark                                 & \checkmark                                  & 32.02\%    &  49.09\%           & 0.85/0.85                                            & 0.44/0.43                                            & 0.73/0.72                                            & 67.53\%                                                 & 39.62\%/40.07\% & 29.63\%/27.08\% & 47.20\% & 76.18\%  \\
                     (6)    &    \checkmark     & \multicolumn{1}{c|}{}                                                  & \checkmark                             & \checkmark                                 & \checkmark                                  & 65.48\%    &  57.81\%           & 0.85/0.84                                            & 0.58/0.56                                            & 0.76/0.76                                            & 64.02\%                                                 & 38.70\%/40.07\% & 28.74\%/31.30\% & 71.98\% & 73.54\%  \\
\rowcolor[HTML]{D9D9D9} 
                     (7)    &         & \multicolumn{1}{c|}{\cellcolor[HTML]{D9D9D9}\checkmark} & \checkmark                             & \checkmark                                 & \checkmark                                  & 68.54\%    & 62.16\%     & 0.84/0.83                                            & 0.57/0.56                                            & 0.78/0.74                                            & 67.97\%                                                 & 38.68\%/42.43\% & 31.58\%/35.10\% & 75.65\% & 75.68\%  \\ \hline
\end{tabular}%
}
\label{tab: Ablation}
\vspace{-0.2cm}
\end{table*}

\begin{figure*}[] 
    \centering
    \vspace{-2pt}
    \includegraphics[width=\textwidth]{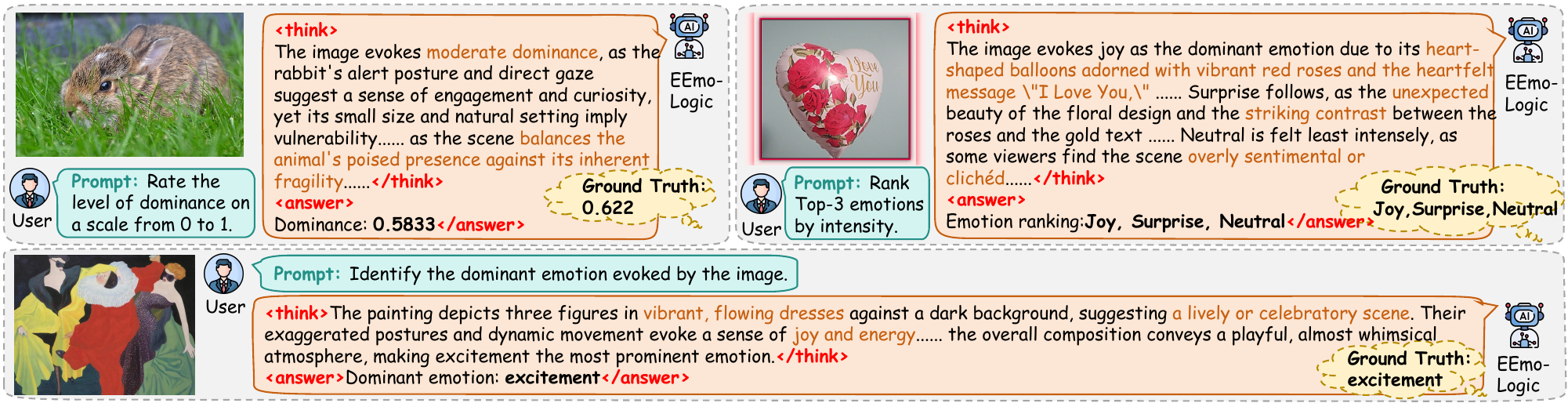}
    \caption{Visualization of EEmo-Logic’s fine-grained emotion assessment and reasoning capabilities in $3$ tasks.}
    \label{fig: case study}
    \vspace{-0.5cm}
\end{figure*}

    


\textbf{Description Task}.
We evaluate open-ended performance using a proprietary LLM judge on customized dimensions \cite{G-Eval}. To ensure alignment with human reasoning and counter verbosity bias \cite{verbosity_bias}, we implement a five-round \textit{DeepSeek} \cite{deepseek} adjudication protocol covering \textit{completeness}, \textit{precision}, \textit{relevance}, and \textit{conciseness} (see Appendix \ref{supp: description evaluation} for details). Tab.~\ref{tab: 3 tasks} demonstrates our model's superior performance across single and multi-image contexts, confirming the high label fidelity of our pipeline. Furthermore, external reasoning constraints enable the model to approximate expert-level emotional cognition, advancing autonomous emotional analysis.

\textbf{Assessment Task}.
As standard models lack reliable numerical regression, we adopt a probability-based protocol \cite{qbench} that maps keywords to VAD levels \cite{EEmobench}. While baselines rely on this method (details in Appendix \ref{supp: softmax}), EEmo-Logic enables direct score prediction. We use spearman rank-order correlation (SRCC) and Pearson Linear Correlation (PLCC) for evaluation. As shown in Tab.~\ref{tab: assessment}, EEmo-Logic delivers robust performance. Most notably, in the dominance dimension, our direct method exceeds the top probability-based competitor (Qwen2.5-VL-7B) by $144\%$ in SRCC. Even under the same protocol, EEmo-Logic performs strongly, particularly in arousal. Additionally, it produces human-aligned reasoning traces and a more balanced score distribution than probability-based methods (discussed in Appendix \ref{supp: distribution}).

\subsection{Evaluation on Cross-Domain Datasets} \label{sec: ood}

To evaluate the generalization ability of EEmo-Logic, we test on $4$ cross-domain datasets, as shown in Tab.~\ref{tab: OOD}.

\textbf{Dominant Emotion Classification.}
EEmo-Logic demonstrates robust overall performance and achieves state-of-the-art results on the ArtEmis dataset. This demonstrates its proficiency in decoding abstract cues like color and composition. While EmoCaliber \cite{EmoCaliber} ranks second due to its focus on evoked emotions, facial expression recognition models fail to generalize because of their dependence on human-specific features and modules. Similarly, video models like R1-Omni \cite{R1-Omni} struggle to transfer to static images. These findings confirm the critical need for specialized image-evoked emotion modeling.

\textbf{Aesthetic Empathy QA Tasks}.
The empathy research module within image aesthetics focuses on understanding evoked emotions through OV definitions and an aesthetic lens. With extensive emotion-related QA pairs, it provides an optimal evaluation testbed. As shown in Tab.~\ref{tab: OOD}, EEmo-Logic delivers robust performance on the emotion subsets of two aesthetic benchmarks. Specifically, in the granular AesBench AesE subset, it achieves top accuracy on emotion questions ($72.30\%$). The model also attains state-of-the-art results on \textit{What}, \textit{How}, and \textit{Why} inquiries, confirming its fine-grained perception and reasoning abilities. Notably, despite lacking training on \textit{Interest} and \textit{Vibe} dimensions, EEmo-Logic exhibits impressive knowledge transfer, effectively bridging affective computing and computational aesthetics.

\subsection{Ablation Study}

\textbf{Analysis of Fine-tuning Strategies}.
Tab.~\ref{tab: Ablation} compares full fine-tuning and LoRA in Stage $1$. As shown in rows ($1$) and ($2$), LoRA excels in perception and description. Since zero-shot evaluation is susceptible to distribution shifts, full fine-tuning risks overfitting to noise. LoRA mitigates this via regularization, promoting robust emotional knowledge transfer capability. Although full fine-tuning holds a slight edge in ranking, the difference is negligible since fine-grained capability is primarily honed during the GRPO stage.
A comparison between Rows ($6$) and ($7$) reveals that, even after GRPO optimization, the full-parameter SFT variant continues to underperform our approach across multiple metrics, thereby further substantiating the superiority of our LoRA-based training strategy.
A detailed discussion and comparison of the experimental settings and training logs are provided in Appendix \ref{supp: sft setting comparison}.

\textbf{Effect of Training Stages and Multi-Task GRPO}.
Comparisons of rows ($2$), ($3$), and ($7$) demonstrate that integrating the QA-focused Stage $1$ with the precision-oriented Stage $2$ yields balanced capabilities, overcoming the generalization limitations of individual stages and enhancing causal reasoning (Fig.~\ref{fig: case study}). Furthermore, rows ($3$), ($4$), and ($7$) indicate that incrementally adding GRPO tasks consistently improves cross-domain performance. While the exploratory nature of GRPO slightly perturbs the SFT framework, causing a marginal decline in description scores, the model retains state-of-the-art status. Notably, incorporating the DEC classification task enriches emotional semantics, validating the robustness of our hierarchical training strategy.
More ablation studies on parameter and method selection are provided in Appendix~\ref{supp: more ablation experiments}.

\section{Conclusion}

In this paper, we introduce \textbf{EEmoDB}, the largest instruction dataset to date for image-evoked emotion understanding, comprising $2$ subsets for general QA and fine-grained label assessment. Building on this, we propose \textbf{EEmo-Logic}, a two-stage framework designed to discern intensity variations, perform human-like reasoning, and evaluate fine-grained emotional attributes. Extensive experiments demonstrate its robust zero-shot generalization across diverse benchmarks. By advancing the comprehensive understanding of image-evoked emotions, this work lays a solid foundation for developing empathic MLLMs.

\section*{Acknowledgements}

This work was supported in part by the National Natural Science Foundation of China under Grant 62522116, Grant 62271312 and Grant 62132006, and in part by STCSM under Grant 22DZ2229005.
We sincerely thank the reviewer team (cNUn, 69qY, DNkB, and Yt1G) for their invaluable feedback to improve our manuscript.

\section*{Impact Statement}

\textbf{Social Impact.} 
Higher-level empathic abilities require models to accurately perceive nuanced intensity differences among the divergent emotions individuals may experience within the same scene, alongside specialized emotional features across various dimensions. Among existing approaches, EEmo-Logic achieves superior performance in these tasks, demonstrating its potential to facilitate more empathetic, user-centric human-computer interactions. We remain committed to advancing this critical line of research in future work.

\textbf{Ethics Statement.}
We clarify that no new images are collected for this work; all data is exclusively sourced from established, publicly available datasets under open academic licenses. Because the original creators have already rigorously screened and removed personally identifiable information (PII) and sensitive content, our aggregated dataset strictly inherits these built-in privacy protections, fully complying with standard academic policies without introducing new ethical risks.
Additionally, we restrict the use of this dataset under the license of CC BY-NC $4.0$, requiring researchers to use our dataset responsibly. Therefore, no ethical concerns are raised in this paper.


\bibliography{reference}

@article{emotion_subjective,
  title={Affective image content analysis: A comprehensive survey},
  author={Zhao, Sicheng and Ding, Guiguang and Huang, Qingming and Chua, Tat-Seng and Schuller, Bj{\"o}rn and Keutzer, Kurt},
  year={2018}
}

@book{human-robot,
  title={Society of mind},
  author={Minsky, Marvin},
  year={1988},
  publisher={Simon and Schuster}
}

@article{advertising,
  title={Multimodal sentiment analysis: A survey and comparison},
  author={Kaur, Ramandeep and Kautish, Sandeep},
  journal={International Journal of Service Science, Management, Engineering, and Technology (IJSSMET)},
  volume={10},
  number={2},
  pages={38--58},
  year={2019},
  publisher={IGI Global Scientific Publishing}
}

@article{human-robot-interaction,
  title={Emotion recognition for human-robot interaction: Recent advances and future perspectives},
  author={Spezialetti, Matteo and Placidi, Giuseppe and Rossi, Silvia},
  journal={Frontiers in Robotics and AI},
  volume={7},
  pages={532279},
  year={2020},
  publisher={Frontiers Media SA}
}

@article{counseling,
  title={Towards emotional support dialog systems},
  author={Liu, Siyang and Zheng, Chujie and Demasi, Orianna and Sabour, Sahand and Li, Yu and Yu, Zhou and Jiang, Yong and Huang, Minlie},
  journal={arXiv preprint arXiv:2106.01144},
  year={2021}
}

@inproceedings{dominant_1,
  title={Building a large scale dataset for image emotion recognition: The fine print and the benchmark},
  author={You, Quanzeng and Luo, Jiebo and Jin, Hailin and Yang, Jianchao},
  booktitle={Proceedings of the AAAI Conference on Artificial Intelligence (AAAI)},
  volume={30},
  number={1},
  year={2016}
}

@inproceedings{dominant_2,
  title={Discovering affective regions in deep convolutional neural networks for visual sentiment prediction},
  author={Sun, Ming and Yang, Jufeng and Wang, Kai and Shen, Hui},
  booktitle={Proceedings of the 2016 IEEE International Conference on Multimedia and Expo (ICME)},
  pages={1--6},
  year={2016},
  organization={IEEE}
}

@article{LDL_1, title={Learning Visual Sentiment Distributions via Augmented Conditional Probability Neural Network}, volume={31}, number={1}, journal={In Proceedings of the AAAI Conference on Artificial Intelligence (AAAI)}, author={Yang, Jufeng and Sun, Ming and Sun, Xiaoxiao}, year={2017}, month={Feb.} }

@ARTICLE{VAD_regression_1,
  author={Mollahosseini, Ali and Hasani, Behzad and Mahoor, Mohammad H.},
  journal={IEEE Transactions on Affective Computing (TAFFC)}, 
  title={AffectNet: A Database for Facial Expression, Valence, and Arousal Computing in the Wild}, 
  year={2019},
  volume={10},
  number={1},
  pages={18-31}}

@ARTICLE{VAD_regression_2,
  author={Zhao, Sicheng and Yao, Hongxun and Gao, Yue and Ji, Rongrong and Ding, Guiguang},
  journal={IEEE Transactions on Multimedia (TMM)}, 
  title={Continuous Probability Distribution Prediction of Image Emotions via Multitask Shared Sparse Regression}, 
  year={2017},
  volume={19},
  number={3},
  pages={632-645}}

@article{SFT,
  title={Visual instruction tuning},
  author={Liu, Haotian and Li, Chunyuan and Wu, Qingyang and Lee, Yong Jae},
  journal={In Proceedings of Advances in Neural Information Processing Systems (NeurIPS)},
  volume={36},
  pages={34892--34916},
  year={2023}
}

@InProceedings{Affection,
    author    = {Achlioptas, Panos and Ovsjanikov, Maks and Guibas, Leonidas and Tulyakov, Sergey},
    title     = {Affection: Learning Affective Explanations for Real-World Visual Data},
    booktitle = {Proceedings of the IEEE/CVF Conference on Computer Vision and Pattern Recognition (CVPR)},
    month     = {June},
    year      = {2023},
    pages     = {6641-6651}
}

@inproceedings{Artphoto,
  title={Affective image classification using features inspired by psychology and art theory},
  author={Machajdik, Jana and Hanbury, Allan},
  booktitle={Proceedings of the 18th ACM International Conference on Multimedia (ACM MM)},
  pages={83--92},
  year={2010}
}

@article{VA,
  title={A Circumplex Model of Affect Journal of Personality and Social Psychology 39},
  author={Russell, JA},
  journal={{\'I}6I-I78},
  year={1980}
}

@article{IAPS,
  title={International affective picture system (IAPS): Technical manual and affective ratings},
  author={Lang, Peter J and Bradley, Margaret M and Cuthbert, Bruce N and others},
  journal={NIMH Center for the Study of Emotion and Attention},
  volume={1},
  number={39-58},
  pages={3},
  year={1997},
  publisher={Gainesville}
}

@article{VAD,
  title={Evidence for a three-factor theory of emotions},
  author={Russell, James A and Mehrabian, Albert},
  journal={Journal of Research in Personality},
  volume={11},
  number={3},
  pages={273--294},
  year={1977},
  publisher={Elsevier}
}

@inproceedings{PCNN,
  title={Joint Image Emotion Classification and Distribution Learning via Deep Convolutional Neural Network.},
  author={Yang, Jufeng and She, Dongyu and Sun, Ming},
  booktitle={Proceedings of the International Joint Conference on Artificial Intelligence (IJCAI)},
  pages={3266--3272},
  year={2017}
}

@InProceedings{EmoSet,
    author    = {Yang, Jingyuan and Huang, Qirui and Ding, Tingting and Lischinski, Dani and Cohen-Or, Danny and Huang, Hui},
    title     = {EmoSet: A Large-scale Visual Emotion Dataset with Rich Attributes},
    booktitle = {Proceedings of the IEEE/CVF International Conference on Computer Vision (ICCV)},
    month     = {October},
    year      = {2023},
    pages     = {20383-20394}
}

@InProceedings{EMOTIC,
author = {Kosti, Ronak and Alvarez, Jose M. and Recasens, Adria and Lapedriza, Agata},
title = {Emotion Recognition in Context},
booktitle = {Proceedings of the IEEE Conference on Computer Vision and Pattern Recognition (CVPR)},
month = {July},
year = {2017}
}

@InProceedings{Emotion6,
author = {Peng, Kuan-Chuan and Chen, Tsuhan and Sadovnik, Amir and Gallagher, Andrew C.},
title = {A Mixed Bag of Emotions: Model, Predict, and Transfer Emotion Distributions},
booktitle = {Proceedings of the IEEE Conference on Computer Vision and Pattern Recognition (CVPR)},
month = {June},
year = {2015}
}

@article{Findingemo,
  title={Findingemo: An image dataset for emotion recognition in the wild},
  author={Mertens, Laurent and Yargholi, Elahe and Op de Beeck, Hans and Van den Stock, Jan and Vennekens, Joost},
  journal={In Proceedings of the Advances in Neural Information Processing Systems (NeurIPS)},
  volume={37},
  pages={4956--4996},
  year={2024}
}

@article{GAPED,
  title={The Geneva affective picture database (GAPED): a new 730-picture database focusing on valence and normative significance},
  author={Dan-Glauser, Elise S and Scherer, Klaus R},
  journal={Behavior Research Methods},
  volume={43},
  number={2},
  pages={468--477},
  year={2011},
  publisher={Springer}
}

@article{OASIS,
  title={Introducing the open affective standardized image set (OASIS)},
  author={Kurdi, Benedek and Lozano, Shayn and Banaji, Mahzarin R},
  journal={Behavior Research Methods},
  volume={49},
  number={2},
  pages={457--470},
  year={2017},
  publisher={Springer}
}

@inproceedings{EEmobench,
  title={Eemo-bench: a benchmark for multi-modal large language models on image evoked emotion assessment},
  author={Gao, Lancheng and Jia, Ziheng and Zeng, Yunhao and Sun, Wei and Zhang, Yiming and Zhou, Wei and Zhai, Guangtao and Min, Xiongkuo},
  booktitle={Proceedings of the 33rd ACM International Conference on Multimedia (ACM MM)},
  pages={7064--7073},
  year={2025}
}

@inproceedings{FI-1,
  title={Cross-modality consistent regression for joint visual-textual sentiment analysis of social multimedia},
  author={You, Quanzeng and Luo, Jiebo and Jin, Hailin and Yang, Jianchao},
  booktitle={Proceedings of the Ninth ACM International Conference on Web Search and Data Mining (WSDM)},
  pages={13--22},
  year={2016}
}

@article{not_distribution,
  title={On the out-of-distribution generalization of multimodal large language models},
  author={Zhang, Xingxuan and Li, Jiansheng and Chu, Wenjing and Hai, Junjia and Xu, Renzhe and Yang, Yuqing and Guan, Shikai and Xu, Jiazheng and Cui, Peng},
  journal={arXiv preprint arXiv:2402.06599},
  year={2024}
}

@article{VAD_norm,
  title={Norms of valence, arousal, and dominance for 13,915 English lemmas},
  author={Warriner, Amy Beth and Kuperman, Victor and Brysbaert, Marc},
  journal={Behavior Research Methods},
  volume={45},
  pages={1191--1207},
  year={2013},
  publisher={Springer}
}

@inproceedings{synonym,
  title={Predicting personalized emotion perceptions of social images},
  author={Zhao, Sicheng and Yao, Hongxun and Gao, Yue and Ji, Rongrong and Xie, Wenlong and Jiang, Xiaolei and Chua, Tat-Seng},
  booktitle={Proceedings of the 24th ACM International Conference on Multimedia (ACM MM)},
  pages={1385--1394},
  year={2016}
}

@article{cluster,
  title={Towards new mappings between emotion representation models},
  author={Landowska, Agnieszka},
  journal={Applied Sciences},
  volume={8},
  number={2},
  pages={274},
  year={2018},
  publisher={MDPI}
}

@article{emotion_circle,
  title={A circumplex model of affect.},
  author={Russell, James A},
  journal={Journal of Personality and Social Psychology},
  volume={39},
  number={6},
  pages={1161},
  year={1980},
  publisher={American Psychological Association}
}

@article{COT,
  title={Chain-of-thought prompting elicits reasoning in large language models},
  author={Wei, Jason and Wang, Xuezhi and Schuurmans, Dale and Bosma, Maarten and Xia, Fei and Chi, Ed and Le, Quoc V and Zhou, Denny and others},
  journal={In Proceedings of the Advances in Neural Information Processing Systems (NeurIPS)},
  volume={35},
  pages={24824--24837},
  year={2022}
}

@article{Emotionqwen,
  title={Emotion-qwen: A unified framework for emotion and vision understanding},
  author={Huang, Dawei and Li, Qing and Yan, Chuan and Cheng, Zebang and Han, Zihao and Huang, Yurong and Li, Xiang and Li, Bin and Wang, Xiaohui and Lian, Zheng and others},
  journal={arXiv preprint arXiv:2505.06685},
  year={2025}
}

@article{Qwen2.5,
  title={Qwen2. 5-vl technical report},
  author={Bai, Shuai and Chen, Keqin and Liu, Xuejing and Wang, Jialin and Ge, Wenbin and Song, Sibo and Dang, Kai and Wang, Peng and Wang, Shijie and Tang, Jun and others},
  journal={arXiv preprint arXiv:2502.13923},
  year={2025}
}

@article{deepseek,
  title={Deepseek-v3 technical report},
  author={Liu, Aixin and Feng, Bei and Xue, Bing and Wang, Bingxuan and Wu, Bochao and Lu, Chengda and Zhao, Chenggang and Deng, Chengqi and Zhang, Chenyu and Ruan, Chong and others},
  journal={arXiv preprint arXiv:2412.19437},
  year={2024}
}

@inproceedings{uniform_distribution,
  title={Delving into deep imbalanced regression},
  author={Yang, Yuzhe and Zha, Kaiwen and Chen, Yingcong and Wang, Hao and Katabi, Dina},
  booktitle={Proceedings of the International Conference on Machine Learning (ICML)},
  pages={11842--11851},
  year={2021},
  organization={PMLR}
}

@article{InstructGPT,
  title={Training language models to follow instructions with human feedback},
  author={Ouyang, Long and Wu, Jeffrey and Jiang, Xu and Almeida, Diogo and Wainwright, Carroll and Mishkin, Pamela and Zhang, Chong and Agarwal, Sandhini and Slama, Katarina and Ray, Alex and others},
  journal={In Proceedings of the Advances in Neural Information Processing Systems (NeurIPS)},
  volume={35},
  pages={27730--27744},
  year={2022}
}

@article{overfit,
  title={Fine-tuning can distort pretrained features and underperform out-of-distribution},
  author={Kumar, Ananya and Raghunathan, Aditi and Jones, Robbie and Ma, Tengyu and Liang, Percy},
  journal={arXiv preprint arXiv:2202.10054},
  year={2022}
}

@article{lora,
  title={Lora: Low-rank adaptation of large language models.},
  author={Hu, Edward J and Shen, Yelong and Wallis, Phillip and Allen-Zhu, Zeyuan and Li, Yuanzhi and Wang, Shean and Wang, Lu and Chen, Weizhu and others},
  journal={In Proceedings of the International Conference on Learning Representations (ICLR)},
  volume={1},
  number={2},
  pages={3},
  year={2022}
}

@article{COT-GRPO,
  title={Unified multimodal chain-of-thought reward model through reinforcement fine-tuning},
  author={Wang, Yibin and Li, Zhimin and Zang, Yuhang and Wang, Chunyu and Lu, Qinglin and Jin, Cheng and Wang, Jiaqi},
  journal={arXiv preprint arXiv:2505.03318},
  year={2025}
}

@article{AffectGPT-R1,
  title={AffectGPT-R1: Leveraging Reinforcement Learning for Open-Vocabulary Multimodal Emotion Recognition},
  author={Lian, Zheng and Zhang, Fan and Zhang, Yazhou and Tao, Jianhua and Liu, Rui and Chen, Haoyu and Li, Xiaobai},
  journal={arXiv preprint arXiv:2508.01318},
  year={2025}
}

@article{GRPO,
  title={Deepseekmath: Pushing the limits of mathematical reasoning in open language models},
  author={Shao, Zhihong and Wang, Peiyi and Zhu, Qihao and Xu, Runxin and Song, Junxiao and Bi, Xiao and Zhang, Haowei and Zhang, Mingchuan and Li, YK and Wu, Yang and others},
  journal={arXiv preprint arXiv:2402.03300},
  year={2024}
}

@article{KRCC,
  title={A new measure of rank correlation},
  author={Kendall, Maurice G},
  journal={Biometrika},
  volume={30},
  number={1-2},
  pages={81--93},
  year={1938},
  publisher={Oxford University Press}
}

@article{Q-Insight,
  title={Q-insight: Understanding image quality via visual reinforcement learning},
  author={Li, Weiqi and Zhang, Xuanyu and Zhao, Shijie and Zhang, Yabin and Li, Junlin and Zhang, Li and Zhang, Jian},
  journal={arXiv preprint arXiv:2503.22679},
  year={2025}
}

@article{Ekman,
  title={An argument for basic emotions},
  author={Ekman, Paul},
  journal={Cognition \& emotion},
  volume={6},
  number={3-4},
  pages={169--200},
  year={1992},
  publisher={Taylor \& Francis}
}

@article{Mikels,
  title={Emotional category data on images from the International Affective Picture System},
  author={Mikels, Joseph A and Fredrickson, Barbara L and Larkin, Gregory R and Lindberg, Casey M and Maglio, Sam J and Reuter-Lorenz, Patricia A},
  journal={Behavior Research Methods},
  volume={37},
  number={4},
  pages={626--630},
  year={2005},
  publisher={Springer}
}

@article{VQ-Insight,
  title={VQ-Insight: Teaching VLMs for AI-Generated Video Quality Understanding via Progressive Visual Reinforcement Learning},
  author={Zhang, Xuanyu and Li, Weiqi and Zhao, Shijie and Li, Junlin and Zhang, Li and Zhang, Jian},
  journal={arXiv preprint arXiv:2506.18564},
  year={2025}
}

@article{VQAThinker,
  title={Vqathinker: Exploring generalizable and explainable video quality assessment via reinforcement learning},
  author={Cao, Linhan and Sun, Wei and Zhang, Weixia and Zhu, Xiangyang and Jia, Jun and Zhang, Kaiwei and Zhu, Dandan and Zhai, Guangtao and Min, Xiongkuo},
  journal={arXiv preprint arXiv:2508.06051},
  year={2025}
}

@article{R1-Omni,
  title={R1-omni: Explainable omni-multimodal emotion recognition with reinforcement learning},
  author={Zhao, Jiaxing and Wei, Xihan and Bo, Liefeng},
  journal={arXiv preprint arXiv:2503.05379},
  year={2025}
}

@article{EmoCaliber,
  title={EmoCaliber: Advancing Reliable Visual Emotion Comprehension via Confidence Verbalization and Calibration},
  author={Wu, Daiqing and Yang, Dongbao and Zhou, Can Ma Yu},
  journal={arXiv preprint arXiv:2512.15528},
  year={2025}
}

@inproceedings{similarity_score,
  title={Refining word embeddings for sentiment analysis},
  author={Yu, Liang-Chih and Wang, Jin and Lai, K Robert and Zhang, Xuejie},
  booktitle={Proceedings of the Conference on Empirical Methods in Natural Language Processing (EMNLP)},
  pages={534--539},
  year={2017}
}

@article{Qwen3-embedding,
  title={Qwen3 Embedding: Advancing Text Embedding and Reranking Through Foundation Models},
  author={Zhang, Yanzhao and Li, Mingxin and Long, Dingkun and Zhang, Xin and Lin, Huan and Yang, Baosong and Xie, Pengjun and Yang, An and Liu, Dayiheng and Lin, Junyang and others},
  journal={arXiv preprint arXiv:2506.05176},
  year={2025}
}

@article{gemini2.5,
  title={Gemini 2.5: Pushing the frontier with advanced reasoning, multimodality, long context, and next generation agentic capabilities},
  author={Comanici, Gheorghe and Bieber, Eric and Schaekermann, Mike and Pasupat, Ice and Sachdeva, Noveen and Dhillon, Inderjit and Blistein, Marcel and Ram, Ori and Zhang, Dan and Rosen, Evan and others},
  journal={arXiv preprint arXiv:2507.06261},
  year={2025}
}

@article{UNIAA,
  title={Uniaa: A unified multi-modal image aesthetic assessment baseline and benchmark},
  author={Zhou, Zhaokun and Wang, Qiulin and Lin, Bin and Su, Yiwei and Chen, Rui and Tao, Xin and Zheng, Amin and Yuan, Li and Wan, Pengfei and Zhang, Di},
  journal={arXiv preprint arXiv:2404.09619},
  year={2024}
}

@article{AesBench,
  title={Aesbench: An expert benchmark for multimodal large language models on image aesthetics perception},
  author={Huang, Yipo and Yuan, Quan and Sheng, Xiangfei and Yang, Zhichao and Wu, Haoning and Chen, Pengfei and Yang, Yuzhe and Li, Leida and Lin, Weisi},
  journal={arXiv preprint arXiv:2401.08276},
  year={2024}
}

@inproceedings{ArtEmis,
  title={Artemis: Affective language for visual art},
  author={Achlioptas, Panos and Ovsjanikov, Maks and Haydarov, Kilichbek and Elhoseiny, Mohamed and Guibas, Leonidas J},
  booktitle={Proceedings of the IEEE/CVF Conference on Computer Vision and Pattern Recognition (CVPR)},
  pages={11569--11579},
  year={2021}
}

@article{deepseek-vl,
  title={Deepseek-vl: towards real-world vision-language understanding},
  author={Lu, Haoyu and Liu, Wen and Zhang, Bo and Wang, Bingxuan and Dong, Kai and Liu, Bo and Sun, Jingxiang and Ren, Tongzheng and Li, Zhuoshu and Yang, Hao and others},
  journal={arXiv preprint arXiv:2403.05525},
  year={2024}
}

@article{janus,
  title={Janus-pro: Unified multimodal understanding and generation with data and model scaling},
  author={Chen, Xiaokang and Wu, Zhiyu and Liu, Xingchao and Pan, Zizheng and Liu, Wen and Xie, Zhenda and Yu, Xingkai and Ruan, Chong},
  journal={arXiv preprint arXiv:2501.17811},
  year={2025}
}

@article{mplug3,
  title={mplug-owl3: Towards long image-sequence understanding in multi-modal large language models},
  author={Ye, Jiabo and Xu, Haiyang and Liu, Haowei and Hu, Anwen and Yan, Ming and Qian, Qi and Zhang, Ji and Huang, Fei and Zhou, Jingren},
  journal={arXiv preprint arXiv:2408.04840},
  year={2024}
}

@article{llava-ov,
  title={Llava-onevision: Easy visual task transfer},
  author={Li, Bo and Zhang, Yuanhan and Guo, Dong and Zhang, Renrui and Li, Feng and Zhang, Hao and Zhang, Kaichen and Zhang, Peiyuan and Li, Yanwei and Liu, Ziwei and others},
  journal={arXiv preprint arXiv:2408.03326},
  year={2024}
}

@article{llava-ov-1.5,
  title={Llava-onevision-1.5: Fully open framework for democratized multimodal training},
  author={An, Xiang and Xie, Yin and Yang, Kaicheng and Zhang, Wenkang and Zhao, Xiuwei and Cheng, Zheng and Wang, Yirui and Xu, Songcen and Chen, Changrui and Zhu, Didi and others},
  journal={arXiv preprint arXiv:2509.23661},
  year={2025}
}

@article{llava-next,
  title={Llava-next-interleave: Tackling multi-image, video, and 3d in large multimodal models},
  author={Li, Feng and Zhang, Renrui and Zhang, Hao and Zhang, Yuanhan and Li, Bo and Li, Wei and Ma, Zejun and Li, Chunyuan},
  journal={arXiv preprint arXiv:2407.07895},
  year={2024}
}

@article{internvl3.5,
  title={Internvl3. 5: Advancing open-source multimodal models in versatility, reasoning, and efficiency},
  author={Wang, Weiyun and Gao, Zhangwei and Gu, Lixin and Pu, Hengjun and Cui, Long and Wei, Xingguang and Liu, Zhaoyang and Jing, Linglin and Ye, Shenglong and Shao, Jie and others},
  journal={arXiv preprint arXiv:2508.18265},
  year={2025}
}

@article{qwen2,
  title={Qwen2-vl: Enhancing vision-language model's perception of the world at any resolution},
  author={Wang, Peng and Bai, Shuai and Tan, Sinan and Wang, Shijie and Fan, Zhihao and Bai, Jinze and Chen, Keqin and Liu, Xuejing and Wang, Jialin and Ge, Wenbin and others},
  journal={arXiv preprint arXiv:2409.12191},
  year={2024}
}

@article{qwen3,
  title={Qwen3 technical report},
  author={Yang, An and Li, Anfeng and Yang, Baosong and Zhang, Beichen and Hui, Binyuan and Zheng, Bo and Yu, Bowen and Gao, Chang and Huang, Chengen and Lv, Chenxu and others},
  journal={arXiv preprint arXiv:2505.09388},
  year={2025}
}

@article{gpt4,
  title={Gpt-4 technical report},
  author={Achiam, Josh and Adler, Steven and Agarwal, Sandhini and Ahmad, Lama and Akkaya, Ilge and Aleman, Florencia Leoni and Almeida, Diogo and Altenschmidt, Janko and Altman, Sam and Anadkat, Shyamal and others},
  journal={arXiv preprint arXiv:2303.08774},
  year={2023}
}

@article{qwen-vl,
  title        = {Qwen-VL: A Versatile Vision-Language Model for Understanding, Localization, Text Reading, and Beyond},
  author       = {Bai, Jinze and Bai, Shuai and Yang, Shusheng and Wang, Shijie and Tan, Sinan and Wang, Peng and Lin, Junyang and Zhou, Chang and Zhou, Jingren},
  journal      = {arXiv preprint arXiv:2308.12966},
  year         = {2023}
}

@article{gpt5,
  title={Openai gpt-5 system card},
  author={Singh, Aaditya and Fry, Adam and Perelman, Adam and Tart, Adam and Ganesh, Adi and El-Kishky, Ahmed and McLaughlin, Aidan and Low, Aiden and Ostrow, AJ and Ananthram, Akhila and others},
  journal={arXiv preprint arXiv:2601.03267},
  year={2025}
}

@misc{claude,
  author       = {Anthropic},
  title        = {Claude 3.7 Sonnet and Claude Code},
  year         = {2025},
  url          = {https://www.anthropic.com/blog},
  note         = {Announcement on Anthropic blog, Feb 24, 2025}
}

@InProceedings{EmoVIT,
    author    = {Xie, Hongxia and Peng, Chu-Jun and Tseng, Yu-Wen and Chen, Hung-Jen and Hsu, Chan-Feng and Shuai, Hong-Han and Cheng, Wen-Huang},
    title     = {EmoVIT: Revolutionizing Emotion Insights with Visual Instruction Tuning},
    booktitle = {Proceedings of the IEEE/CVF Conference on Computer Vision and Pattern Recognition (CVPR)},
    month     = {June},
    year      = {2024},
    pages     = {26596-26605}
}

@article{AffectGPT,
  title={Affectgpt: A new dataset, model, and benchmark for emotion understanding with multimodal large language models},
  author={Lian, Zheng and Chen, Haoyu and Chen, Lan and Sun, Haiyang and Sun, Licai and Ren, Yong and Cheng, Zebang and Liu, Bin and Liu, Rui and Peng, Xiaojiang and others},
  journal={arXiv preprint arXiv:2501.16566},
  year={2025}
}

@article{verbosity_bias,
  title={Judging llm-as-a-judge with mt-bench and chatbot arena},
  author={Zheng, Lianmin and Chiang, Wei-Lin and Sheng, Ying and Zhuang, Siyuan and Wu, Zhanghao and Zhuang, Yonghao and Lin, Zi and Li, Zhuohan and Li, Dacheng and Xing, Eric and others},
  journal={In Proceedings of the Advances in Neural Information Processing Systems (NeurIPS)},
  volume={36},
  pages={46595--46623},
  year={2023}
}

@article{G-Eval,
  title={G-eval: NLG evaluation using gpt-4 with better human alignment},
  author={Liu, Yang and Iter, Dan and Xu, Yichong and Wang, Shuohang and Xu, Ruochen and Zhu, Chenguang},
  journal={arXiv preprint arXiv:2303.16634},
  year={2023}
}

@article{qbench,
  title={Q-bench: A benchmark for general-purpose foundation models on low-level vision},
  author={Wu, Haoning and Zhang, Zicheng and Zhang, Erli and Chen, Chaofeng and Liao, Liang and Wang, Annan and Li, Chunyi and Sun, Wenxiu and Yan, Qiong and Zhai, Guangtao and others},

  journal={arXiv preprint arXiv:2309.14181},
  year={2023}
}

@inproceedings{jiang2020dfew,
  title={Dfew: A large-scale database for recognizing dynamic facial expressions in the wild},
  author={Jiang, Xingxun and Zong, Yuan and Zheng, Wenming and Tang, Chuangao and Xia, Wanchuang and Lu, Cheng and Liu, Jiateng},
  booktitle={Proceedings of the 28th ACM International Conference on Multimedia (ACM MM)},
  pages={2881--2889},
  year={2020}
}

@article{lian2023explainable,
  title={Explainable multimodal emotion recognition},
  author={Lian, Zheng and Sun, Haiyang and Sun, Licai and Gu, Hao and Wen, Zhuofan and Zhang, Siyuan and Chen, Shun and Xu, Mingyu and Xu, Ke and Chen, Kang and others},
  journal={arXiv preprint arXiv:2306.15401},
  year={2023}
}

@inproceedings{liu2024mmbench,
  title={Mmbench: Is your multi-modal model an all-around player?},
  author={Liu, Yuan and Duan, Haodong and Zhang, Yuanhan and Li, Bo and Zhang, Songyang and Zhao, Wangbo and Yuan, Yike and Wang, Jiaqi and He, Conghui and Liu, Ziwei and others},
  booktitle={Proceedings of the European Conference on Computer Vision (ECCV)},
  pages={216--233},
  year={2024},
  organization={Springer}
}

@article{dai2023instructblip,
  title={Instructblip: Towards general-purpose vision-language models with instruction tuning},
  author={Dai, Wenliang and Li, Junnan and Li, Dongxu and Tiong, Anthony and Zhao, Junqi and Wang, Weisheng and Li, Boyang and Fung, Pascale N and Hoi, Steven},
  journal={In Proceedings of the Advances in Neural Information Processing Systems (NeurIPS)},
  volume={36},
  pages={49250--49267},
  year={2023}
}

@inproceedings{lin2014microsoft,
  title={Microsoft coco: Common objects in context},
  author={Lin, Tsung-Yi and Maire, Michael and Belongie, Serge and Hays, James and Perona, Pietro and Ramanan, Deva and Doll{\'a}r, Piotr and Zitnick, C Lawrence},
  booktitle={Proceedings of the European Conference on Computer Vision (ECCV)},
  pages={740--755},
  year={2014},
  organization={Springer}
}

@article{zhou2019semantic,
  title={Semantic understanding of scenes through the ade20k dataset},
  author={Zhou, Bolei and Zhao, Hang and Puig, Xavier and Xiao, Tete and Fidler, Sanja and Barriuso, Adela and Torralba, Antonio},
  journal={International Journal of Computer Vision (IJCV)},
  volume={127},
  number={3},
  pages={302--321},
  year={2019},
  publisher={Springer}
}

@incollection{plutchik1980general,
  title={A general psychoevolutionary theory of emotion},
  author={Plutchik, Robert},
  booktitle={Theories of Emotion},
  pages={3--33},
  year={1980},
  publisher={Elsevier}
}

@article{VAD_generation_validation,
  title   = {Ratings are Overrated!},
  author  = {Yannakakis, Georgios N. and Mart{\'i}nez, H{\'e}ctor P.},
  journal = {Frontiers in ICT},
  volume  = {2},
  pages   = {13},
  year    = {2015},
  doi     = {10.3389/fict.2015.00013}
}

@InProceedings{RICE_VIT,
    author    = {Xie, Yin and Yang, Kaicheng and An, Xiang and Wu, Kun and Zhao, Yongle and Deng, Weimo and Ran, Zimin and Wang, Yumeng and Feng, Ziyong and Miles, Roy and Elezi, Ismail and Deng, Jiankang},
    title     = {Region-based Cluster Discrimination for Visual Representation Learning},
    booktitle = {Proceedings of the IEEE/CVF International Conference on Computer Vision (ICCV)},
    month     = {October},
    year      = {2025},
    pages     = {1793-1803}
}

@article{llama3,
  title={The llama 3 herd of models},
  author={Grattafiori, Aaron and Dubey, Abhimanyu and Jauhri, Abhinav and Pandey, Abhinav and Kadian, Abhishek and Al-Dahle, Ahmad and Letman, Aiesha and Mathur, Akhil and Schelten, Alan and Vaughan, Alex and others},
  journal={arXiv preprint arXiv:2407.21783},
  year={2024}
}

@inproceedings{Q-former,
  title={Blip-2: Bootstrapping language-image pre-training with frozen image encoders and large language models},
  author={Li, Junnan and Li, Dongxu and Savarese, Silvio and Hoi, Steven},
  booktitle={Proceedings of the International Conference on Machine Learning (ICML)},
  pages={19730--19742},
  year={2023},
  organization={PMLR}
}

@article{zhao2025humanomni,
  title={Humanomni: A large vision-speech language model for human-centric video understanding},
  author={Zhao, Jiaxing and Yang, Qize and Peng, Yixing and Bai, Detao and Yao, Shimin and Sun, Boyuan and Chen, Xiang and Fu, Shenghao and Wei, Xihan and Bo, Liefeng and others},
  journal={arXiv preprint arXiv:2501.15111},
  year={2025}
}

@book{deviation,
  title={Statistical inference},
  author={Casella, George and Berger, Roger},
  year={2024},
  publisher={Chapman and Hall/CRC}
}

@article{transformer,
  title={Attention is all you need},
  author={Vaswani, Ashish and Shazeer, Noam and Parmar, Niki and Uszkoreit, Jakob and Jones, Llion and Gomez, Aidan N and Kaiser, {\L}ukasz and Polosukhin, Illia},
  journal={In Proceedings of Advances in Neural Information Processing Systems (NeurIPS)},
  volume={30},
  year={2017}
}
\bibliographystyle{icml2026}

\newpage
\appendix
\onecolumn
\section{Overview}

This supplementary material provides comprehensive details regarding the datasets, methodology, experiments, and results discussed in the main text.
The content is organized as follows:
Sec.~\ref{supp sec: emotion datasets} benchmarks EEmoDB against established datasets in the AICA domain and offers detailed descriptions of the source data and evaluation protocols. 
Sec.~\ref{supp sec: dataset construction} delineates the dataset construction pipeline, with a specific focus on emotion label refinement and QA generation procedures. 
Sec.~\ref{supp sec: method theory} elaborates on the algorithmic formulations and theoretical underpinnings.
Sec.~\ref{supp sec: implementation details} describes implementation specifics, including the compared models, evaluation protocols, and metrics. 
Finally, Sec.~\ref{supp sec: more results} presents additional qualitative visualizations and an in-depth discussion of the results.

\section{Detailed Information of Emotion Datasets} \label{supp sec: emotion datasets}

Tab.~\ref{tab:dataset} compares representative AICA datasets, instruction sets, and benchmarks. The two subsets of EEmoDB explicitly capture the diversity and subjectivity of evoked emotions, providing comprehensive coverage of both CES and DES. In addition, EEmoDB offers large-scale QA data, paired samples, and emotion reasoning instances, making it well-suited for training MLLMs to achieve comprehensive and fine-grained image-evoked emotion understanding. We describe the seven source datasets and $5$ evaluation datasets and benchmarks as follows:

\textbf{Abstract-8} \cite{Artphoto} is an image dataset designed for emotion classification in abstract paintings. The dataset comprises a collection of abstract artworks expressed through purely visual elements such as color, texture, and composition, without containing recognizable figurative content. All images are annotated with eight basic emotion categories, including $4$ positive and $4$ negative emotions. The dataset is constructed based on a psychological emotion taxonomy framework, with images sourced from open art platforms and annotated through a multi-annotator process. 

\textbf{GAPED} \cite{GAPED} is a standardized image dataset designed for emotion induction research. It includes six categories of affective stimuli: snakes, spiders, human rights violations, animal mistreatment, neutral images, and positive images. All images have undergone standardized preprocessing and are accompanied by continuous ratings across $4$ dimensions: valence, arousal, moral norm acceptability, and legal norm acceptability. The dataset employs a multi-annotator labeling process with statistical consistency validation, ensuring cross-participant comparability of the ratings.

\textbf{Emotion6} \cite{Emotion6} is a standardized dataset for emotion distribution modeling in images. It is constructed based on Ekman's six basic emotion categories \cite{Ekman}, along with a ``neutral" category serving as an affective baseline. Each image is annotated with continuous valence-arousal ratings and multi-label emotion distribution data. The dataset employs a balanced category design with an equal number of images per class, and excludes images containing obvious facial expressions or emotion-related text.

\textbf{OASIS} \cite{OASIS} is an open-access affective image dataset containing images from multiple thematic categories, including humans, animals, scenes, and objects. Each image is associated with normative ratings on the dimensions of valence and arousal, obtained through a large-scale online rating procedure. The dataset also includes normative rating records and metadata for each image.

\textbf{EMOTIC} \cite{EMOTIC} is a large-scale image dataset designed for individual-level emotion recognition in unconstrained environments. It consists of contextual images collected from MSCOCO~\cite{lin2014microsoft}, ADE20K~\cite{zhou2019semantic}, and the web. Each visible person in every image is annotated at a fine-grained level using two complementary systems: a set of $26$ discrete emotion categories, and continuous ratings along the dimensions of valence, arousal, and dominance. 

\textbf{Affection} \cite{Affection} is a large-scale dataset of affective explanations for real-world images. It collects human emotional responses to images along with corresponding textual justifications. The annotation follows a two-stage process: annotators first select a dominant emotion for each image, then provide a written explanation based on the visual content. The resulting language exhibits a combination of emotional orientation, abstract expression, and subjective perspective.

\textbf{FindingEmo} \cite{Findingemo} is an image dataset constructed for emotion understanding in social scenes, comprising publicly sourced images depicting natural interactions among multiple individuals. The dataset employs Plutchik’s Wheel of Emotions framework~\cite{plutchik1980general} for systematic annotation, providing both basic emotion categories and fine-grained emotion descriptions, along with multi-dimensional human annotations including valence, arousal, and participant age groups. All annotations are collected through a structured crowdsourcing pipeline with corresponding quality verification mechanisms.

\textbf{EEmo-Bench} \cite{EEmobench} is a benchmark dataset for systematically evaluating the ability of MLLMs to comprehend image-evoked emotions. It contains $1,960$ images collected from Flickr, spanning diverse content categories including animals, humans, and scenes. The annotation scheme combines manual ranking of the top $3$ evoked emotions per image based on Ekman's basic emotion theory, with quantitative Valence-Arousal-Dominance scores on a self-assessment manikin (SAM) $9$-point scale \cite{IAPS}. To facilitate comprehensive assessment, $6,773$ question-answer pairs are constructed across $4$ core tasks: perception, ranking, description, and assessment.

\textbf{Artphoto} \cite{Artphoto} is a dataset for affective image recognition research, consisting of $807$ professional photographs collected from an art-sharing platform. The dataset is constructed via a retrieval-based method using eight discrete emotion categories as search terms, with emotional labels self-assigned by the uploading artists. The images exhibit systematic control by creators over visual dimensions such as color, composition, lighting, and texture to convey distinct emotional expressions.

\textbf{ArtEmis} \cite{ArtEmis} is a large-scale dataset of emotional explanations for visual artworks, comprising over $455,000$ emotion-explanation pairs based on $80,031$ artworks from WikiArt. Annotators are asked to select a dominant emotion from eight basic categories and provide free-text justifications, linking visual content, affective experience, and linguistic explanation. The annotations are characterized by subjectivity and linguistic richness, including metaphors, abstract concepts, and personal expressions.

\textbf{UNIAA} \cite{UNIAA} is a dataset for evaluating the aesthetic perception capability of MLLMs. It contains $5,354$ images, each accompanied by a multiple-choice question across six aesthetic dimensions: content and theme, composition, color, light, focus, and sentiment. The questions are formulated in $3$ types: ``Yes-or-No", ``What", and ``How". The dataset is divided into in-domain and in-the-wild splits. In this work, we adopt the Sentiment dimension from UNIAA to evaluate models’ emotion understanding ability.

\textbf{AesBench} \cite{AesBench} is an expert-annotated dataset for evaluating the image aesthetics perception capability of MLLMs. It contains a large-scale, multi-source expert-labeled aesthetics perception database covering $3$ main categories: natural images, artistic images, and AI-generated images. All annotations are provided by professional aesthetics experts, offering multi-turn question-answer pairs and detailed interpretation texts across $4$ dimensions: aesthetic perception, empathy, assessment, and interpretation. In this study, we employ its aesthetic empathy question-answering subset for testing.

\begin{table*}[]
\scriptsize
\centering
\setlength{\tabcolsep}{2.7pt}
\caption{Dataset comparison. ``DE'' denotes the Dimensional Emotion Space, while ``OV'' indicates that the emotion category set is defined based on an open-vocabulary lexicon.}
\begin{tabular}{lcccccccc}
\hline
\textbf{Dataset}             & \multicolumn{1}{l}{\textbf{Images}} & \multicolumn{1}{l}{\textbf{Emotions}} & \multicolumn{1}{l}{\textbf{Emotion Diversity}} & \multicolumn{1}{l}{\textbf{DES}} & \multicolumn{1}{l}{\textbf{QA pairs}} & \multicolumn{1}{l}{\textbf{Pairwise Samples}} & \multicolumn{1}{l}{\textbf{Reasoning}} & \multicolumn{1}{c}{\textbf{Annotation Manner}} \\ \hline
Artphoto \cite{Artphoto}      & 807                        & 8                            & \xmark                                   & \xmark                      & \xmark                           & \xmark                                   & \xmark                            & Human                                 \\
Abstract-8 \cite{Artphoto}    & 280                        & 8                            & \xmark                                    & \xmark                      & \xmark                           & \xmark                                   & \xmark                            & Human                                 \\
GAPED \cite{GAPED}         & 730                        & \xmark                           & \xmark                                    & VA                      & \xmark                           & \xmark                                   & \xmark                            & Human                                 \\
Emotion6 \cite{Emotion6}      & 1980                       & 7                            & \checkmark                                   & VA                      & \xmark                           & \xmark                                   & \xmark                            & Human                                 \\
FI \cite{dominant_1}            & 23,308                     & 9                            & \xmark                                    & \xmark                      & \xmark                           & \xmark                                   & \xmark                            & Human                                 \\
IESN \cite{synonym}         & 1,012,901                  & 8                            & \xmark                                    & VAD                     & \xmark                           & \xmark                                   & \xmark                            & Social             \\
OASIS \cite{OASIS}         & 900                        & \xmark                           & \xmark                                    & VA                      & \xmark                           & \xmark                                   & \xmark                            & Human                                 \\
EMOTIC \cite{EMOTIC}        & 18,302                     & 26                           & \xmark                                    & VAD                     & \xmark                           & \xmark                                   & \xmark                            & Human                                 \\
FlickrLDL \cite{LDL_1}    & 10,700                     & 11                           & \checkmark                                   & \xmark                      & \xmark                           & \xmark                                   & \xmark                            & Human                                 \\
TwitterLDL \cite{LDL_1}   & 10,045                     & 8                            & \checkmark                                   & \xmark                      & \xmark                           & \xmark                                   & \xmark                            & Human                                 \\
ArtEmis \cite{ArtEmis}       & 80,031                     & 9                            & \checkmark                                   & \xmark                      & \xmark                           & \xmark                                   & \checkmark                           & Human                                 \\
Affection \cite{Affection}     & 85,007                     & 9                            & \checkmark                                   & \xmark                      & \xmark                           & \xmark                                   & \checkmark                           & Human                                 \\
EmoSet \cite{EmoSet}        & 118,102                    & 8                            & \xmark                                    & \xmark                      & \xmark                           & \xmark                                   & \xmark                            & Human                                 \\
FindingEmo \cite{Findingemo}    & 21,964                     & 32                           & \xmark                                    & VA                      & \xmark                           & \xmark                                   & \xmark                            & Human                                 \\
AesBench AesE \cite{AesBench} & 2,800                      & OV                           & \xmark                                    & \xmark                      & 2,800                        & \xmark                                   & \checkmark                           & Human                                 \\
UNIAA Sent. \cite{UNIAA}   & 464                        & OV                           & \xmark                                    & \xmark                      & 464                          & \xmark                                   & \checkmark                           & Human                                 \\
EmoVIT \cite{EmoVIT}        & 51,200                     & OV                           & \xmark                                    & \xmark                      & 62,361                       & \xmark                                   & \checkmark                           & Model-led+Human-assisted              \\
EEmo-Bench \cite{EEmobench}     & 1,960                      & 7                            & \checkmark                                   & VAD                     & 6,773                        & 1,008                                & \checkmark                           & Human                                 \\
\textbf{EEmoDB-QA  (Ours)}  & 122,332                    & 7                            & \checkmark                                   & VAD                     & 1,197,288                    & 355,611                              & \checkmark                           & Model-led+Human-assisted              \\
\textbf{EEmoDB-Assess (Ours)}    & 25,315                     & 7(52)                        & \checkmark                                   & VAD                     & 36,000                       & \xmark                                   & \xmark                            & Model-led+Human-assisted              \\ \hline
\end{tabular}
\label{tab:dataset}
\end{table*}

\section{Dataset Construction Details} \label{supp sec: dataset construction}

\subsection{Details of VAD-based Emotion Clustering} \label{supp: cluster}

We first delineate the characteristics of the EMOTIC \cite{EMOTIC} dataset, which provides multi-annotator labels comprising both discrete emotion categories and continuous VAD scores (we normalize the VAD scores for convenience). A critical limitation, however, is that annotators do not explicitly quantify intensity differences among selected emotions; instead, the labels follow a fixed, pre-defined order rather than a preference-based ranking.
To resolve this ambiguity, we adopt a uniform contribution hypothesis: for a given image, we assume each emotion selected by an annotator contributes equally to the associated VAD scores. Mathematically, we distribute a total probability mass of $1$ uniformly across the selected categories. Consequently, when an annotator selects fewer emotions, each receives a higher probability mass, effectively serving as a proxy for higher emotional intensity or preference, or higher \textbf{concentration} of viewpoints. While this heuristic may introduce noise at the individual level, such variance is effectively mitigated through large-scale aggregation, yielding robust global intensity estimates.

Building on these probability distributions, we bridge the gap between categorical and dimensional representations by establishing a linear mapping from the $26$-category discrete space to the three-dimensional VAD space inspired by \cite{cluster}. Specifically, we hypothesize that the continuous VAD scores can be modeled as a linear combination of discrete emotion probabilities. Let $p_e\in\mathbb{R}^{26}$ denote the normalized probability vector over the discrete categories (e.g., ``engagement'', ``anticipation'', ``happiness''), and $s_{\text{vad}}\in\mathbb{R}^{3}$ represent the corresponding VAD vector. We formulate this relationship as a multivariate linear regression problem without an intercept term:

\begin{equation}
    s_{\text{vad}}=\mathbf{W}p_e+\mathbf{\epsilon}
\end{equation}

where $\mathbf{W}\in\mathbb{R}^{3\times26}$ is the projection matrix (coefficient matrix) to be estimated, and $\mathbf{\epsilon}$ is the residual error. In this formulation, the $j$-th column of $\mathbf{W}$ can be interpreted as the VAD centroid (or coordinates) of the $j$-th discrete emotion category.

To ensure the reliability and generalizability of the estimated mappings, we employ a repeated $k$-fold cross-validation scheme (specifically, $10$-fold cross-validation repeated $10$ times). Regression coefficients are estimated on training partitions, and the final VAD scores for each emotion category are derived by averaging these coefficients across all folds and repetitions. This rigorous process minimizes sampling bias from random splitting, yielding a robust estimation of the affective mapping.

Finally, we identify anchor emotions from the EMOTIC set that best align with the six Ekman categories and the ``neutral'' state. We determine an appropriate clustering radius based on Euclidean distances between VAD vectors and refine the results through manual semantic verification to establish the final mapping. Notably, by bridging CES and DES, this strategy effectively preserves the annotators' original labeling characteristics and emotional preferences, thereby minimizing semantic bias in the emotion representation. The mapping result is shown in Tab.~\ref{tab: emotic emotion}.

\begin{table}[]
\centering
\scriptsize
\setlength{\tabcolsep}{4pt}    
\caption{Clustering-based mapping results from EMOTIC emotions to Ekman emotion categories.}
\begin{tabular}{llllll}
\hline
\textbf{Mapped emotion} & \textbf{Anchor emotion} & \textbf{Valence anchor} & \textbf{Arousal anchor} & \textbf{Dominance anchor} & \textbf{Members}                                                                                                         \\ \hline
Joy              & Pleasure       & 0.6605         & 0.5017         & 0.6320           & \begin{tabular}[c]{@{}l@{}}Anticipation, Happiness, Excitement, Confidence, Pleasure, Affection, \\ Esteem, Sympathy, Yearning, Sensitivity\end{tabular} \\
Surprise         & Surprise       & 0.5069         & 0.5077         & 0.5326           & Engagement, Surprise, Embarrassment                                                                             \\
Anger            & Anger          & 0.2160         & 0.6575         & 0.5032           & Anger                                                                                                           \\
Disgust          & Aversion       & 0.3651         & 0.4591         & 0.5024           & Disquietment, Doubt/Confusion, Annoyance, Disapproval,   Aversion                                               \\
Sadness          & Sadness        & 0.2098         & 0.3250         & 0.3707           & Sadness, Suffering, Pain                                                                                        \\
Fear             & Fear           & 0.4093         & 0.7010         & 0.4415           & Fear                                                                                                            \\
Neutral            & Peace          & 0.5283         & 0.3603         & 0.6027           & Disconnection, Peace, Fatigue                                                                                   \\ \hline
\end{tabular}
\label{tab: emotic emotion}
\end{table}

\subsection{Validation of VAD Generation Quality} \label{supp: vad generation}

\textbf{Usage and Validation of Discrete VAD.}
To rigorously assess the synthesized VAD scores, we adopt a discretized validation strategy, where continuous scales are quantized into $3$ ordinal levels (high, medium, and low) based on tertiles. We select $1,800$ samples via stratified sampling, and $5$ domain experts independently categorize the perceived intensity in a blind evaluation. By computing classification accuracy, this protocol mitigates the subjective noise of exact regression while ensuring semantic alignment. 
The underlying rationale for this approach is that the scores assigned by individual experts are inherently discrete, given that human judgment exhibits greater stability in categorical classification than in continuous value estimation \cite{VAD_generation_validation}. Although integrating scores from multiple subjects into MOSs yields robust continuous values, obtaining such a large number of expert opinions for each individual sample is highly resource-intensive. 
Additionally, it should be noted that we evaluate our model's discrete and continuous VAD capabilities in Tab.~\ref{tab: 3 tasks} \& \ref{tab: assessment}, respectively. Because existing datasets for discrete VAD QA are relatively small, we apply the IESN approach \cite{synonym} in Stage $1$ to expand the QA training set. Therefore, this discrete validation method specifically verifies the effectiveness of this data generation method by comparing the generated three-level categories against human classification results.
The results demonstrate agreement rates of $94.00\%$, $93.83\%$, and $91.17\%$ for Valence, Arousal, and Dominance, respectively, confirming that our lexicon-based VAD generation process aligns closely with human perception.
Furthermore, we compute Krippendorff’s $\alpha$ for the VAD dimensions, yielding $0.8174$, $0.6432$, and $0.6879$, respectively. Despite the inherent subjectivity of evoked emotion understanding, these scores demonstrate robust annotation consistency.

\textbf{Usage and Validation of Continuous VAD.}
For the continuous VAD prediction task (corresponding to Stage $2$), training relies primarily on gold-standard human MOSs from existing datasets; our generated data using the IESN approach serves merely as a minor supplement to balance the distribution.
As illustrated in Fig.~\ref{fig: Affection distribution}(a), the number of words contributing to the VAD score calculation varies for each image. Assuming individual word scores may contain inherent noise, an increased word count mitigates these errors, yielding more robust outcomes \cite{deviation}. Consequently, when constructing EEmoDB-Assess, we prioritized images with a higher density of VAD-related terms. By introducing samples in descending order of word count to balance the score distribution, we ensure the high quality and reliability of the dataset.
To further validate these continuous scores, we conduct an additional human evaluation ($600$ samples, $6$ annotators, $9$-point SAM scale). The results yield SRCC values of $0.7060$, $0.6221$, and $0.6946$, and Krippendorff’s $\alpha$ values of $0.6558$, $0.5700$, and $0.5737$ for VAD, respectively. This quantitatively proves the reliability of the generated scores. Given their supplementary role in the Stage $2$ training corpus, this quality is highly acceptable. 

\subsection{Reliability of Reasoning Annotations and QA Correctness} \label{supp: QA validation}

To directly address the concern (partially discussed in Sec.~\ref{sec: refine}), we would like to first clarify the rigorous logic behind our automatic annotation pipeline. The reliability of our generated QA pairs and emotion reasoning is guaranteed by three core factors:
1) \textbf{Validated Foundations.} All source datasets are established affective image content analysis benchmarks. Their core emotion labels are rigorously annotated and validated by multiple humans.
2) \textbf{Verified Expansion.} When unifying semantic spaces and expanding label dimensions (e.g., VAD), we incorporate additional human verification to ensure rigorous accuracy.
3) \textbf{Constrained LLM Generation.} We strictly confine the LLM's role to structural reorganization. It solely utilizes human-verified ground truth to construct QA pairs and distills existing expert comments into structured reasoning trajectories. Consequently, because the underlying content originates entirely from validated human annotations, our pipeline fundamentally precludes hallucinations. This ensures high-quality outputs without necessitating redundant manual verification.

To further validate the quality of the model-generated QA pairs, we uniformly sample $400$ instances from the Perception task, encompassing both single-image and image-pair scenarios. An evaluation by three human annotators yields an accuracy of 
$82.88\%$ (via majority voting). This robust human performance strongly substantiates the reliability and high quality of our generated QA pairs. 

\subsection{Prompts for Description and Reasoning Generation} \label{Supp: reasoning}

\begin{figure}[]
    \centering
    \begin{minipage}{0.9\textwidth}
        \begin{subfigure}[b]{0.48\textwidth} 
            \centering
            \includegraphics[width=\textwidth]{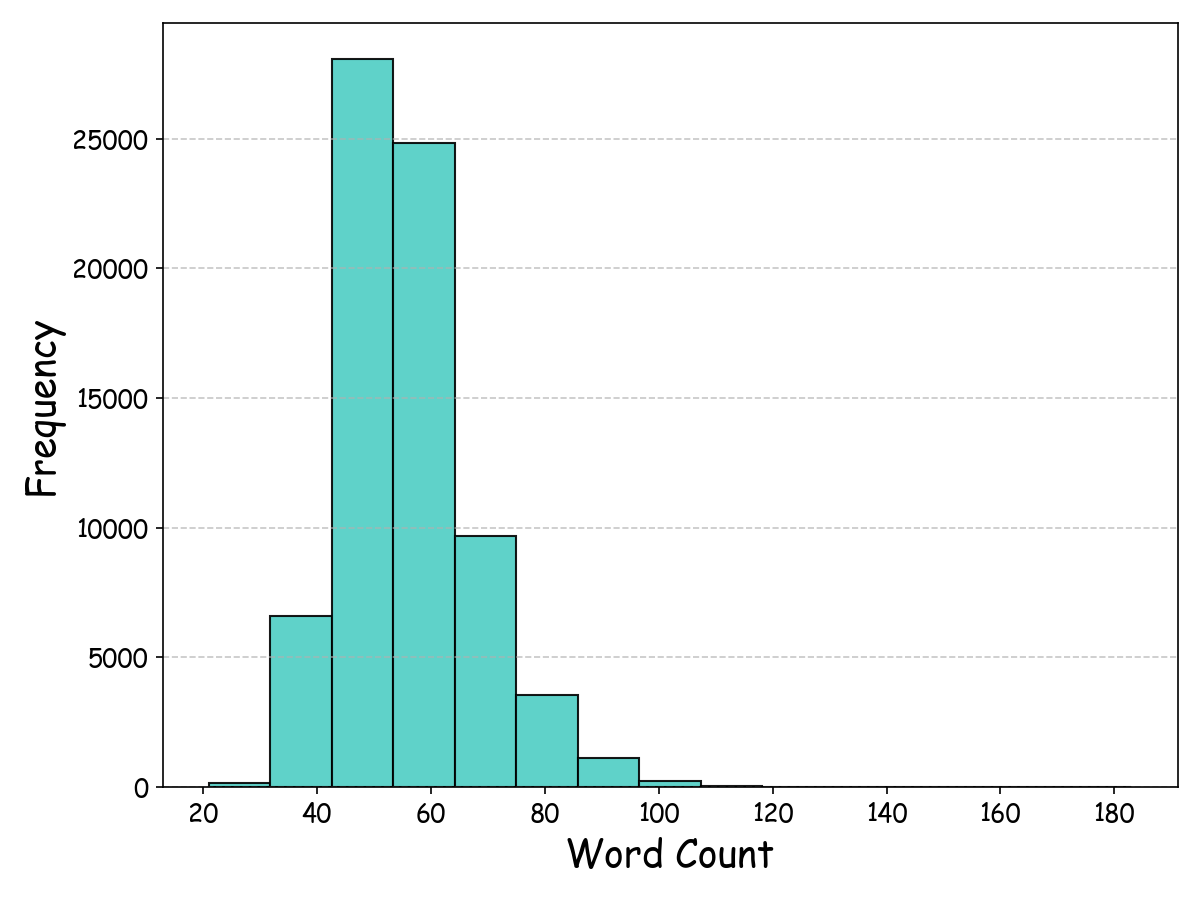} 
            \caption{Number of words within VAD lexicon per image}
            \label{supp fig: adj distribution}
        \end{subfigure}
        \hfill 
        \begin{subfigure}[b]{0.48\textwidth}
            \centering
            \includegraphics[width=\textwidth]{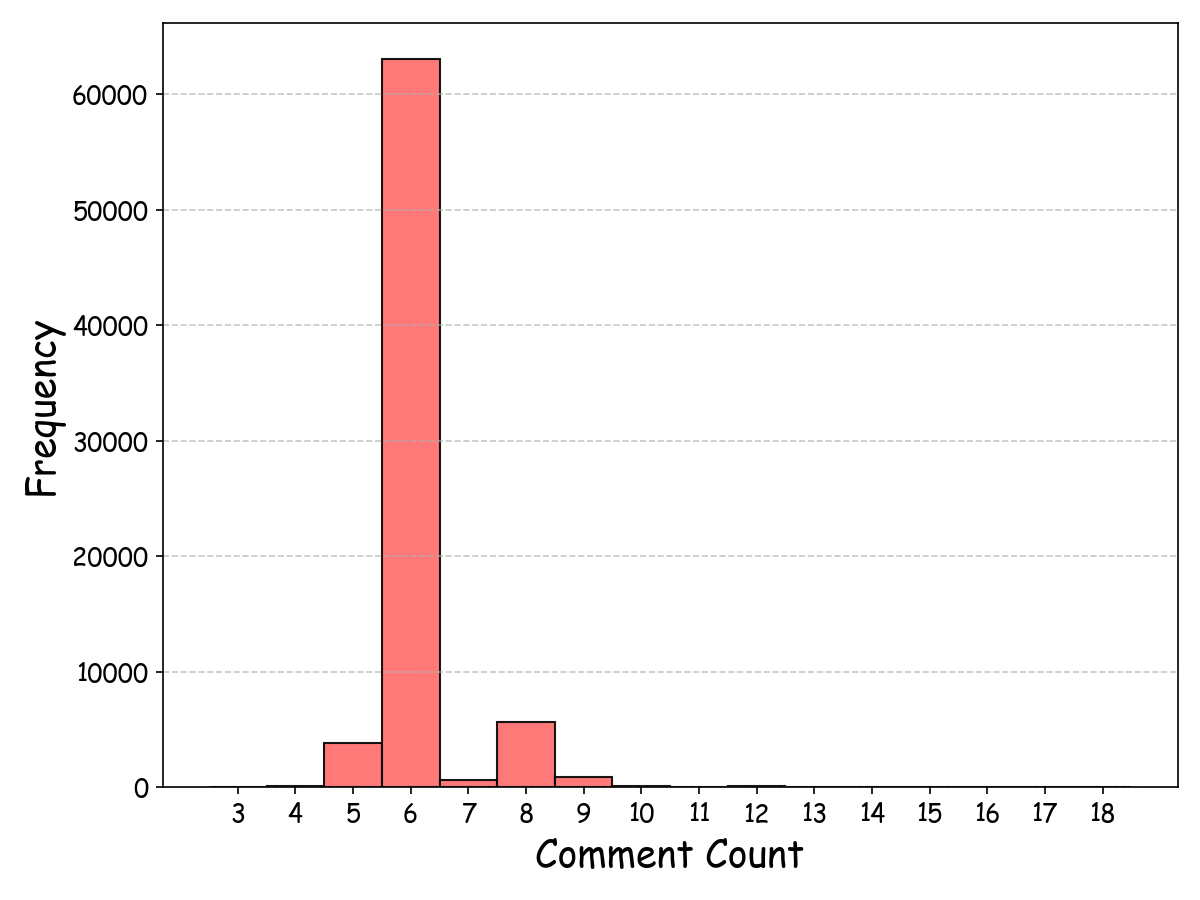}
            \caption{Number of Comments per image}
            \label{supp fig: comments distribution}
        \end{subfigure}
        \hfill 
    \end{minipage}

    \caption{Label statistics of the Affection dataset. (a) Distribution of identified VAD keywords within comments per image, referenced against $13,915$ English lemma norms \cite{VAD_norm} for score generation. (b) Distribution of the number of expert comments available for each image.}
    \label{fig: Affection distribution}
\end{figure}

Although the emotion-reasoning texts are generated via an automated pipeline, their emotional grounding remains strictly anchored in the expert annotations of the Affection dataset \cite{Affection}. 
As shown in Fig.~\ref{fig: Affection distribution}(b), the majority of images in the dataset have more than $5$ expert comments and emotion labels, reflecting diverse emotional causes and providing a rich corpus for emotional analysis.
By utilizing \textit{Qwen2.5} \cite{Qwen2.5} to produce concise image content descriptions and explicitly specifying key reasoning steps and constraints, we effectively leverage the robust CoT reasoning and summarization capabilities of \textit{DeepSeek-V3} \cite{deepseek}. This strategy enables the large-scale generation of emotion reasoning texts that maintain expert-level quality and human-like fidelity. The specific prompt is detailed below.

For description, we apply the following prompt:

{\small \textit{User: \\ Describe the main objects and events in the image. Focus on the most important elements and keep the description concise.}}

For Reasoning, we align each \textbf{emotion} (after the emotion mapping process described in Sec.~\ref{sec: refine}) with the corresponding human \textbf{comment}. Together with the \textbf{description} texts obtained in the previous step, these inputs are summarized by Deepseek using the following prompt:

{\small \textit{System: \\You are a helpful assistant and an expert in emotional reasoning and textual analysis. Your role is to generate concise, logical reasoning narratives explaining how a described scene evokes a sequence of emotions.  \\
Constraints:
                - Keep the output concise and informative, with a maximum of 200 words.
                - Summarize commentary lists rather than restating them verbatim.
                - Use clear causal connectors (e.g., “because,” “therefore,” “which leads to”).
                - Output a single coherent paragraph with no bullet points.  \\
Clarifications:
                - The order of emotions is ranked purely by the number of people who report feeling them, in descending order.  
                  This ranking reflects the diversity of emotions the image evokes, not a sequence of emotional transitions.  
                - When referring to commentary, do not say “the comment says...”.  
                  Instead, interpret and express them as “people think/believe/feel that...”.}}

{\small \textit{User: \\You are given descriptive inputs about an image, including the content described in the picture, the emotions it evokes, and comments related to the reasons for each emotion, displayed in []. \\
    Inputs :
    - Image description: $<$description$>$
    - Emotions and comments (in order):
      1. $<$emotion 1$>$ - comment: $<$comment 1$>$,
      2. $<$emotion 2$>$ - comment: $<$comment 2$>$, ... \\
    Task:
    Synthesize the key points and produce a reasoning-style explanation of how the image elicits these emotions in the given intensity order. Explicitly link concrete visual elements from the description to each emotion and (when relevant) to the intensity difference between emotions. \\
    Output:
    A $\leq$ 200-word reasoning narrative mapping visual cues → ordered emotions.}}

\subsection{Classification of the Question Template}

\textbf{Perceptual QA for Single Image}.
We design $20$ templates, which can be grouped into $3$ categories according to their focus:
1) \textbf{Emotion verification}. Queries the presence or dominance of specific emotions (e.g., assessing whether arousal is low or identifying joy as the dominant emotion). 
2) \textbf{Comparative identification}. Fosters the ability to distinguish subtle nuances and perform exclusionary reasoning (e.g., determining if fear outweighs sadness or comparing valence levels). 
3) \textbf{Visual attribution}. Requires grounding emotions in either local visual details or global composition (e.g., pinpointing the visual causes of evoked emotion).

\textbf{Perceptual QA for Image Pairs}.
While single-image understanding is fundamental, robust affective intelligence requires the ability to perform relational reasoning across multiple visual stimuli. We design $18$ templates, which are grouped into two categories based on their focus:
1) \textbf{Consistency verification}, where the model validates shared affective traits and attributional patterns across image pairs, requiring a sound understanding of both images (e.g., whether both evoke anger or exhibit high dominance);
2) \textbf{Comparative assessment}, which demands fine-grained discrimination in the VAD space and comparison of emotional differences (e.g., which image evokes higher arousal or which has surprise as the dominant emotion).

\textbf{Descriptive QA}.
To foster the application of emotional knowledge in long-form reasoning, we design $7$ description templates categorized by analysis scope: 1) \textbf{Single-image analysis}: Entails articulating perceptions across the $5$ label dimensions, elucidating causal mechanisms of elicitation, and analyzing emotion sources (direct vs. indirect). 2) \textbf{Paired-image comparison}: Necessitates delineating similarities and differences across dimensions, alongside a comparative analysis of VAD levels with causal justification.
Answers to these questions are summarized and generated by Deepseek-V3 based on the provided emotion labels and auxiliary information, including image descriptions and emotion reasoning texts, producing responses that follow human-like reasoning logic. The specific prompt used is provided in Appendix \ref{sec: description QA prompt}.

\subsection{Emotion Ranking Generation Method} \label{supp: ranking method}

As source datasets lack explicit intensity rankings, we derive ground-truth labels via dataset-specific strategies, discussed in two cases:

\textbf{1) Distribution-based Extraction.} For datasets providing probability distributions (e.g., Emotion6 \cite{Emotion6}) or frequency statistics (e.g., Abstract-8 \cite{Artphoto}, Affection \cite{Affection}), we first apply the category mapping from Sec.~\ref{sec: refine}. We then identify the top $3$ emotions based on their probability or frequency. To ensure ranking validity, we exclusively retain samples exhibiting distinct probability gradients among these top $3$ emotions.

\textbf{2) Progressive Comparison Strategy.} For datasets like EMOTIC \cite{EMOTIC}, where annotators provide multiple unranked labels to capture diversity, we derive intensity rankings through a three-stage progressive comparison procedure following category mapping, detailed as follows:

\begin{itemize}
    \item Selection Frequency: We aggregate annotation frequencies for each emotion, sorting them in descending order to identify the top $3$ candidates.
    \item Emotional Concentration: To resolve ties from the first step, we calculate a score based on emotional concentration. Specifically, for an annotator selecting $n$ emotions, each emotion receives a weight of $\frac{1}{n
    }$. We sum these weights across annotators, thereby prioritizing emotions from more focused labels.
    \item Original Intensity Order: Should ties persist, we leverage the sequence of annotation, assigning higher weights to earlier-listed emotions under the assumption of primacy. 
\end{itemize}

Samples remaining tied after this three-step process are discarded. Ultimately, this procedure yields $3,659$ valid samples, constituting the curated data for EEmoDB-Assess.

\subsection{Prompt for Descriptive Answer Generation} \label{sec: description QA prompt}

Here, we present the specific prompt used to generate open-ended QA answers in EEmoDB-QA with Deepseek, based on the \textbf{emotions}, \textbf{comments}, \textbf{description}, and \textbf{reasoning}, as well as the scores of \textbf{valence}, \textbf{arousal}, \textbf{dominance} annotations from the Affection \cite{Affection} dataset. The prompts can be categorized into the following cases:

(1) Generating detailed reasoning narratives for a single image, including the evoked emotion ranking and VAD-level interpretation.

{\small \textit{System: \\
You are an expert in emotional reasoning and textual analysis. 
You will receive an image description, a ranked list of emotions, commentary subtitles associated with each emotion, and 
VAD (valence, arousal, dominance) scores given by human emotional experts. \\
Rules:
1. Use only the provided textual inputs; do not invent details beyond them.
2. The order of emotions reflects prevalence (descending number of people who felt them), not an emotional transition. Do not describe a shift from one emotion to another.
3. When using commentary subtitles, do not quote them directly as ``comments say...". Instead, summarize them into collective perspectives (e.g., ``people think/feel/believe that...").
4. Output must be in English, as a single coherent paragraph, without bullet points.
5. Keep the output concise yet descriptive, maximum 200 words.
6. For VAD scores (normalized to 0–1):
   - Map numbers into qualitative levels: low $\leq$ 0.4, medium if 0.4 $<$ x $<$ 0.6, high $\geq$ 0.6. Do not include specific numerical values in the description.
   - Present both the VAD level and a short rationale tied to the description based on the given description and comments.}}

{\small \textit{User: \\
You are given descriptive inputs about an image, including the content described in the picture, the emotions it evokes, and comments related to the reasons for each emotion, displayed in [].
    You are also given VAD (valence, arousal, dominance) scores given by human emotional experts. \\
    Inputs :
    - Image description: $<$description$>$
    - Emotions (in order):
      1. $<$emotion 1$>$ - comment: $<$comment 1$>$,
      2. $<$emotion 2$>$ - comment: $<$comment 2$>$, ...
    - VAD scores (normalized to 1): Valence=$<$valence rating$>$, Arousal=$<$arousal rating$>$, Dominance=$<$dominance rating$>$ \\
Task:
Write one paragraph in English that:
1) Describes the image content concisely, based on the description;
2) States the $3$ main emotions (in the given order), clarifying they represent diverse audience reactions rather than a progression;
3) For each emotion, summarize what people generally believe or feel according to the associated commentary subtitles;
4) Explain the VAD levels by linking them to visual elements in the description, giving the qualitative level (low/medium/high), with a short reason for each.}}

(2) For open-ended questions that ask for reasoning about the causes of the main evoked emotions, we directly use the \textit{Reasoning} annotations; the corresponding generation prompt is provided in Appendix \ref{Supp: reasoning}.

(3) Generating reasoning answers that determine whether the source of the image-evoked emotion is direct or indirect.

{\small \textit{System: \\
You are an expert in affective image analysis.  
Your task is to determine whether the emotion **evoked** by a given image is **more directly** or **more indirectly** triggered, or if **both direct and indirect factors** are equally significant in evoking the emotion, based on the provided image description and reasoning text.\\
Definitions:
1. **More direct emotional evocation**: The emotion is primarily evoked by the immediate visual effects in the image, such as the image's aesthetic elements (e.g., posture, facial expression, color scheme, etc.). The viewer's emotional response is triggered directly by what is visually presented.
2. **More indirect emotional evocation**: The emotion is primarily evoked by the viewer's associations, cultural knowledge, past experiences, or symbolic meanings. The viewer’s emotional response is triggered through cognitive processes beyond just the immediate visual effects.
3. **Both direct and indirect emotional evocation**: Both the image's visual elements and the viewer's cognitive associations or knowledge play significant roles in evoking the emotion. \\
Rules:
1. Use only the provided image description and reasoning text to analyze how the emotion is evoked.
2. **More direct** should be selected when the emotion is predominantly evoked by the immediate visual effects in the image.
3. **More indirect** should be selected when the emotion is predominantly evoked through the viewer's associations or cognitive processes.
4. **Both direct and indirect** should be selected when both the visual elements and the viewer's associations play significant roles in evoking the emotion.
5. The output must follow the format:
   - ``The emotional evocation of this image is more direct/indirect/both direct and indirect."
   - Provide an analysis explaining why the emotional evocation is classified as direct, indirect, or both, grounded in the image description and reasoning text.
   - **Avoid quoting the comments directly**. Instead, use phrasing such as ``people generally feel..." or ``some people may feel...", based on the reasoning and the description. \\
Output format:
- A clear, concise sentence stating ``more direct", ``more indirect", or ``both direct and indirect".
- A brief explanation based on the description and reasoning text, without directly quoting the comments.
- End with a concise summary sentence.}}

{\small \textit{User: \\
Inputs:
- Image description: $<$description$>$
- Emotions (in order):
      1. $<$emotion 1$>$ - comment: $<$comment 1$>$,
      2. $<$emotion 2$>$ - comment: $<$comment 2$>$, ... \\
Task:
Determine if the emotional evocation of the image is **more direct**, **more indirect**, or **both direct and indirect**.
- **More direct** means the emotion is primarily evoked by the visual elements of the image itself (e.g., posture, facial expression, color scheme, etc.).
- **More indirect** means the emotion is primarily evoked through the viewer’s associations, common knowledge, or past experiences (e.g., symbolism, cultural associations, etc.).
- **Both direct and indirect** means the emotion is evoked by both the visual elements of the image and the viewer’s cognitive associations or knowledge. \\
Output:
- **First, state the conclusion**: whether the emotional evocation is **more direct**, **more indirect**, or **both direct and indirect**.
- **Then, provide an analysis** explaining why the emotional evocation is direct, indirect, or both, based on the description and reasoning.
- **Avoid using direct quotes** from the comments. Instead, use generalizations like ``people generally feel..." or ``some people may feel..." based on the description and reasoning text.
- **End with a short summary** of how the emotion is evoked.}}

(4) Generating comparative analyses of emotion rankings and VAD levels between two images, highlighting their similarities and differences.

{\small \textit{System: \\
You are an expert in affective computing and image emotion analysis. 
Your task is to compare two images based on their provided emotional and VAD information.
Inputs for each image include:
- A list of $3$ emotions in descending order of intensity (emotion ranking).
- Normalized VAD scores: Valence, Arousal, Dominance (ranging 0–1).
- A short reasoning text explaining the emotional causes. \\
Rules:
1. Use only the provided inputs; do not invent additional emotions or reasons.
2. Compare the emotion rankings of the two images, emphasizing similarities or differences in dominant and secondary emotions.
3. Compare the VAD dimensions (Valence, Arousal, Dominance):
   - Map normalized scores into qualitative levels:
     * Valence: negative (0–0.4), neutral (0.4–0.6), positive (0.6–1.0).
     * Arousal: low (0–0.4), moderate (0.4–0.6), high (0.6–1.0).
     * Dominance: low (0–0.4), moderate (0.4–0.6), high (0.6–1.0).
   - If both images fall in the same interval for a dimension, describe them as similar in that aspect.
   - If they differ in interval, highlight the difference.
4. The reasoning texts should be briefly integrated into the comparison to provide explanatory context.
5. Output style: 
   - Analytical yet natural language.
   - At least one cohesive paragraph.
   - English only.
6. Do not output lists or bullet points; use prose comparison.
7. **Do not mention the numerical scores** in your output. Only describe the qualitative level (e.g., “low arousal,” “positive,” “high dominance”).
8. Keep the output concise yet descriptive, maximum 200 words.}}

{\small \textit{User: \\
Inputs:
- Image one:
  * Emotions (ranked): $<$emotion ranking 1$>$
  * Valence: $<$valence rating 1$>$, Arousal: $<$arousal rating 1$>$, Dominance: $<$dominance rating 1$>$
  * Reasoning: $<$reasoning 1$>$
- Image two:
  * Emotions (ranked): $<$emotion ranking 2$>$
  * Valence: $<$valence rating 2$>$, Arousal: $<$arousal rating 2$>$, Dominance: $<$dominance rating 2$>$
  * Reasoning: $<$reasoning 2$>$ \\
Task:
Compare the two images in terms of:
1. Their ranked emotions (highlighting similarities or differences in dominant and secondary emotions).
2. Their VAD dimensions, mapped into qualitative levels (Valence → negative/neutral/positive; Arousal and Dominance → low/moderate/high).
3. Identify where they are similar and where they differ, using the thresholds 0.4 and 0.6.
4. Incorporate reasoning texts to support the interpretation of emotional differences.
Output a cohesive English paragraph summarizing the similarities and differences.}}

(5) Comparing a specific VAD dimension between two images and providing a causal explanation for the observed difference. Here, ``\textbf{vad}" denotes one selected dimension from valence, arousal, or dominance.

{\small \textit{System: \\
You are an expert in affective computing and emotion analysis. 
Your task is to answer questions about the comparison of two images’ emotional responses based on provided normalized VAD scores (Valence, Arousal, Dominance) and reasoning texts. \\
Rules:
1. Each image has one VAD dimension value (0–1) and an associated reasoning text.
2. Convert the numerical VAD score into qualitative categories:
   - Valence: negative (0–0.4), neutral (0.4–0.6), positive (0.6–1.0).
   - Arousal: low (0–0.4), moderate (0.4–0.6), high (0.6–1.0).
   - Dominance: low (0–0.4), moderate (0.4–0.6), high (0.6–1.0).
3. Always structure your answer as:
   - First, give the **direct conclusion** (e.g., ``The emotional response to Image 2 shows a stronger sense of dominance.").
   - Then, provide **analysis** for each image, linking the description and reasoning to its VAD category and interpretation.
   - Finally, give a **short summary sentence** that restates the comparison.
4. Use clear, academic-style English. 
5. Do not invent details beyond the given description and reasoning.
6. Output must be in natural prose, not bullet points.
7. **Do not mention the numerical scores** in your output. Only describe the qualitative level (e.g., “low arousal,” “positive,” “high dominance”).
8. Keep the output concise yet descriptive, maximum 200 words.}}

{\small \textit{User: \\
Inputs:
- Image one:
  * $<$vad$>$ score: $<$valence/arousal/dominance rating 1$>$
  * Reasoning: {reasoning 1}
- Image two:
  * $<$vad$>$ score: $<$valence/arousal/dominance rating 2$>$
  * Reasoning: {reasoning 2}
Question: $<$question$>$ \\
Task:
1. Compare the two images according to the given $<$vad$>$ scores and reasoning texts.
2. Translate the scores into qualitative levels (negative/neutral/positive for valence, low/moderate/high for arousal and dominance).
3. Answer the question by:
   - Giving a clear conclusion first.
   - Providing an analytical explanation for both images using only qualitative levels, not numerical scores.
   - Ending with a summary sentence comparing the two images.}}

\subsection{Presentation Issues and Potential Risks of Data Generation Pipeline.} \label{sec: potentail risks}

Regarding potential risks, we offer a brief analysis below:
1) \textbf{Cross-Cultural Ambiguity.} Unifying diverse datasets introduces cultural variance, potentially rendering certain emotion labels ambiguous across different backgrounds.
2) \textbf{Distractor Plausibility.} For complex ``why/how" questions, distractors may represent plausible secondary factors rather than strictly incorrect options, introducing minor ambiguity.
Despite these inherent risks, our model achieves robust performance across all tasks on manually labeled test sets like EEmo-Bench \cite{EEmobench}. This resilience validates our training strategy and indicates that the large-scale SFT setting effectively dilutes the impact of isolated ambiguous samples.

\section{Supplementary Methodological Details} \label{supp sec: method theory}

\subsection{Details of Loss Function in Stage 1}

In Stage $1$, we adopt SFT and apply LoRA to both the image encoder and the LLM layers. Model parameter 
$\theta$ is optimized using the standard cross-entropy loss with details shown as follows. 
Given a dataset $D={(x,y)}$ consisting of input instructions $x$ and corresponding target responses $y=(y_1,y_2,…,y_T)$, the training objective is to minimize the negative log-likelihood of the target tokens conditioned on the input and preceding context:

\begin{equation}
    \mathcal{L}_{SFT}(\theta)=-\mathbb{E}_{(x,y)\sim\mathcal{D}}\left[\sum_{t=1}^T\log P(y_t\mid x,y_{<t};\theta)\right],
\end{equation}

where $y_{<t}$ denotes the sequence of tokens preceding $y_t$, and $P(y_t|\cdot)$ represents the probability distribution over the vocabulary at step $t$.

\subsection{Training Protocol of GRPO in Stage 2} \label{supp: GRPO}

In Stage $2$, we primarily introduce innovations in the reward design, while the training pipeline and underlying principles remain consistent with the standard GRPO framework \cite{GRPO}. The details are described as follows. 
During training, for a given input image $\mathcal{I}$ and prompt $u$, GRPO generates $N$ distinct responses ${v_1, v_2,...,v_N}$ using the previous policy $\pi_{\theta_{\text{old}}}$, parameterized by the weights $\theta_{\text{old}}$ from the previous epoch. Each response is evaluated with our designed reward function, yielding reward scores ${r_1, r_2,...,r_N}$. GRPO then computes the normalized advantage score $A_i$ for the $i$-th response as the deviation of its reward from the group mean, scaled by the standard deviation:

\begin{equation}
    \hat{A}_i=\frac{r_i-\mu(r_j))}{\sigma(r_j))},
\end{equation}

where $\mu(\cdot)$ and $\sigma(\cdot)$ represent the mean and standard deviation operators, respectively. GRPO then computes the likelihood ratio of each response under the new policy $\pi_{\theta_{\text{new}}}$ and the old policy $\pi_{\theta_{\text{old}}}$, clipping it to the interval $[1 - \delta, 1 + \delta]$ to avoid excessive updates and ensure stable training. Finally, the policy is updated to increase the likelihood of responses with higher relative advantage while penalizing large deviations from the reference policy through a KL divergence term. The optimization objective for GRPO is defined as:

\begin{equation}
\begin{aligned}
\mathcal{L}_{GRPO}(\theta)  =\mathbb{E}_{[u\sim \mathcal{U},v_{i}\sim\pi_{\theta_{\mathrm{old}}}(v|u)]}\Big\{\min\Big[\rho_{i}\hat{A}_{i},  \operatorname{clip}(\rho_{i},1-\delta,1+\delta)\hat{A}_{i}\Big]-\beta\cdot\mathbb{D}_{\mathrm{KL}}[\pi_{\theta}\|\pi_{\mathrm{ref}}]\Big\} ,
\end{aligned}
\end{equation}

where $\rho_{i} = \frac{\pi_{\theta}(v_{i}|\boldsymbol{u}_{i})}{\pi_{\theta_{\mathrm{old}}}(v_{i}|\boldsymbol{u}_{i})}$ denotes the update ratio between the new and old policies for response $v_i$, $\delta$ controls update stability, and $\beta$ balances the KL-regularization relative to the reference model. $\mathcal{Q}$ represents the question set.

\subsection{Generating Semantic Descriptions for Calculating Semantic Similarity} \label{supp: semantic}

To better calculate the semantic similarity $S_{\text{emb}}$ of different emotions within the same emotion set in the emotion similarity reward (Sec.~\ref{sec: grpo}), we refine and expand the definition of each emotion. First, we use the VAD lexicon to find the VAD scores corresponding to each emotion and categorize them into different levels to describe the coarse-grained emotional intensity. Next, we employ Gemini-2.5-Pro \cite{gemini2.5} to generate stimuli and feature descriptions for each emotion. Finally, we summarize these two aspects to form the semantic description of each emotion, as shown in Tab.~\ref{tab: emotion defination}.

\begin{table}[!ht]
\caption{Semantic descriptions of emotion categories employed for calculating the semantic similarity reward $S_{\text{emb}}$.}
\resizebox{\textwidth}{!}{%
\begin{tabular}{lll}
\hline
Emotion         & Definition                                                                                                                                                                                  &  \\ \hdashline
\multicolumn{2}{l}{1. Plutchik Emotion (24)}                                                                                                                                                                  &  \\ \hdashline
Trust           & A positive, moderate energy state of reliance and safety. Reliance on the integrity or strength of someone. Willingness to be vulnerable.                                                   &  \\
Admiration      & A positive, moderate energy state of looking up to someone. Recognition of superior qualities in another. Respect and wonder.                                                               &  \\
Distraction     & A neutral, low energy state of fragmented focus. Fragmented attention. Inability to maintain focus on the primary task due to interference.                                                 &  \\
Joy             & A positive, moderate energy state with a sense of control. Response to immediate success or delightful stimuli. Smiling, laughter, and sudden uplift in energy. Engagement with the moment. &  \\
Boredom         & A negative, low energy state with stagnant focus. Perception of the environment as static and lacking stimulation. Restless desire for engagement but inability to find it.                 &  \\
Terror          & A negative, high energy state of total overwhelm and helplessness. Overwhelming reaction to imminent mortal danger. Paralysis, screaming, or collapse of rational thought.                  &  \\
Amazement       & A positive, high energy state characterized by being overwhelmed yet impressed. Response to something unexpected. Momentary suspension of action. Wide eyes and jaw drop.                   &  \\
Disgust         & A negative, moderate energy state of rejection. Visceral rejection of something toxic, contaminated, or morally offensive. Nausea and urge to turn away.                                    &  \\
Rage            & A negative, high energy state of explosive volatility. Violent reaction to provocation. Loss of impulse control, screaming, and destructive urges.                                          &  \\
Vigilance       & A neutral, low energy state of watchful control. Sustained attention to detect potential signals or threats. Guarded behavior and scanning the environment.                                 &  \\
Anticipation    & A neutral, moderate energy state of expectancy. Looking forward to a future event. Mental preparation and expectation of an outcome.                                                        &  \\
Anger           & A negative, high energy state of antagonism. Reaction to perceived injustice, interference, or threat. Increased heart rate, glaring, and impulse to attack or defend.                      &  \\
Loathing        & A negative, moderate energy state of intense revulsion and hatred. Deep-seated and enduring disgust mixed with hatred. Desire to eliminate or completely avoid the object.                  &  \\
Grief           & A negative, low energy state of deep sorrow and distress. Intense suffering caused by the death of a loved one or a major tragedy. Aching, sobbing, and profound emptiness.                 &  \\
Acceptance      & A positive, moderate energy state of acknowledging reality. Cognitive consent to reality without resistance. Acknowledging facts without attempting to change them.                         &  \\
Fear            & A negative, high energy state of immediate threat response. Reaction to imminent danger. Adrenaline rush, freezing or fleeing, and focused attention on the threat.                         &  \\
Pensiveness     & A negative, moderate energy state of internal reflection. State of deep, serious reflection. Withdrawal of attention from the outside world.                                                &  \\
Serenity        & A positive, low energy state with balanced composure. Deep inner stillness and clarity of mind. Unruffled composure regardless of external chaos.                                           &  \\
Surprise        & A positive, high energy state of sudden reaction. Reaction to an unexpected event. Momentary interruption of thought processing. Startle response.                                          &  \\
Interest        & A positive, moderate energy state of focused engagement. Focus of attention on a novel stimulus. Desire to explore, learn, or understand more.                                              &  \\
Sadness         & A negative, low energy state of withdrawal and helplessness. Response to irrevocable loss or failure. Heaviness, tears, lethargy, and withdrawal from social interaction.                   &  \\
Apprehension    & A negative, moderate energy state of uneasy anticipation. Feeling that something bad might happen. Low-level dread or foreboding.                                                           &  \\
Annoyance       & A negative, moderate energy state of irritation. Reaction to a nuisance or repetitive disturbance. Mild anger and desire for cessation.                                                     &  \\
Ecstasy         & A positive, high energy state with balanced control. Overwhelming peak experience where self-awareness dissolves. Trance-like state induced by extreme sensory intensity.                   &  \\ \hdashline
\multicolumn{2}{l}{2. EMOTIC (26)}                                                                                                                                                                            &  \\ \hdashline
Pain            & A negative, high energy state of physical or emotional suffering. Signal of damage or distress. Sharp focus on the source of hurt.                                                          &  \\
Sadness         & A negative, low energy state of withdrawal and helplessness. Response to irrevocable loss or failure. Heaviness, tears, lethargy, and withdrawal from social interaction.                   &  \\
Embarrassment   & A negative, moderate energy state of social awkwardness. Reaction to a minor social gaffe or exposure. Blushing and nervous smiling.                                                        &  \\
Fatigue         & A negative, low energy state of depletion. Physical or mental exhaustion. Lack of resources to continue activity. Heaviness and slowness.                                                   &  \\
Excitement      & A positive, high energy state with a feeling of empowerment. Anticipation of a stimulating future event. High physical energy, eagerness, and focused attention on upcoming rewards.        &  \\
Engagement      & A positive, moderate energy state of flow. State of being fully occupied and absorbed in an activity. Active participation.                                                                 &  \\
Confidence      & A positive, high energy state of self-assurance. Belief in one's own abilities and judgment. Posture is upright and gaze is steady.                                                         &  \\
Aversion        & A negative, moderate energy state of avoidance. Strong dislike or disinclination towards something. Moving away or turning head.                                                            &  \\
Happiness       & A positive, high energy state of feeling dominant and secure. Sustained evaluation that life conditions are favorable. General sense of stability, satisfaction, and well-being.            &  \\
Yearning        & A negative, moderate energy state of longing. Deep desire for something lost or unattainable. Aching sense of incompleteness.                                                               &  \\
Affection       & A positive, moderate energy state of warmth and fondness. Feeling of liking and caring for another. Gentle touch and soft expression.                                                       &  \\
Sensitivity     & A neutral, moderate energy state of heightened responsiveness. Quick detection of subtle changes in external stimuli or emotional cues.                                                     &  \\
Fear            & A negative, high energy state of immediate threat response. Reaction to imminent danger. Adrenaline rush, freezing or fleeing, and focused attention on the threat.                         &  \\
Sympathy        & A positive, moderate energy state of shared feeling. Understanding and concern for another's suffering. Desire to alleviate their pain.                                                     &  \\
Suffering       & A negative, moderate energy state of undergoing pain or distress. The subjective experience of pain or hardship. Endurance of unpleasant conditions over time.                              &  \\
Doubt/Confusion & A negative, moderate energy state of uncertainty and hesitation. Lack of certainty or clarity. Inability to interpret signals.                                                              &  \\
Disapproval     & A negative, moderate energy state of negative judgment. Evaluation that something is wrong or unacceptable. Frowning and shaking head.                                                      &  \\
Anticipation    & A neutral, moderate energy state of expectancy. Looking forward to a future event. Mental preparation and expectation of an outcome.                                                        &  \\
Anger           & A negative, high energy state of antagonism. Reaction to perceived injustice, interference, or threat. Increased heart rate, glaring, and impulse to attack or defend.                      &  \\
Esteem          & A positive, moderate energy state of respect and value. High regard for oneself or others. Feeling of worth and dignity.                                                                    &  \\
Surprise        & A positive, high energy state of sudden reaction. Reaction to an unexpected event. Momentary interruption of thought processing. Startle response.                                          &  \\
Disquietment    & A negative, moderate energy state of restless unease. Feeling that something is not right. Inability to settle or relax.                                                                    &  \\
Pleasure        & A positive, high energy state of gratification and control. Sensory gratification from physical stimuli (taste, touch) or aesthetic beauty. Immediate comfort and enjoyment.                &  \\
Disconnection   & A negative, low energy state of detachment. Sense of detachment from reality. Numbness, dissociation, or feeling separated by a wall.                                                       &  \\
Peace           & A positive, moderate energy state with a strong sense of security. Absence of conflict, disturbance, or war. Social harmony and freedom from external commotion.                            &  \\
Annoyance       & A negative, moderate energy state of irritation. Reaction to a nuisance or repetitive disturbance. Mild anger and desire for cessation.                                                     &  \\ \hdashline
\multicolumn{2}{l}{3. Ekman Emotion + Neutral (7)}                                                                                                                                                            &  \\ \hdashline
Fear            & A negative, high energy state of immediate threat response. Reaction to imminent danger. Adrenaline rush, freezing or fleeing, and focused attention on the threat.                         &  \\
Joy             & A positive, moderate energy state with a sense of control. Response to immediate success or delightful stimuli. Smiling, laughter, and sudden uplift in energy. Engagement with the moment. &  \\
Anger           & A negative, high energy state of antagonism. Reaction to perceived injustice, interference, or threat. Increased heart rate, glaring, and impulse to attack or defend.                      &  \\
Sadness         & A negative, low energy state of withdrawal and helplessness. Response to irrevocable loss or failure. Heaviness, tears, lethargy, and withdrawal from social interaction.                   &  \\
Surprise        & A positive, high energy state of sudden reaction. Reaction to an unexpected event. Momentary interruption of thought processing. Startle response.                                          &  \\
Disgust         & A negative, moderate energy state of rejection. Visceral rejection of something toxic, contaminated, or morally offensive. Nausea and urge to turn away.                                    &  \\
Neutral         & A neutral, low energy state of equilibrium. Baseline state. Absence of strong activation or valence. Calmness.                                                                              &  \\ \hdashline
\multicolumn{2}{l}{4. Mikels Emotion (8)}                                                                                                                                                                     &  \\ \hdashline
Disgust         & A negative, moderate energy state of rejection. Visceral rejection of something toxic, contaminated, or morally offensive. Nausea and urge to turn away.                                    &  \\
Contentment     & A positive, moderate energy state with a sense of stability. Long-term acceptance of one's lot in life. Gratitude for stability and lack of friction.                                       &  \\
Sadness         & A negative, low energy state of withdrawal and helplessness. Response to irrevocable loss or failure. Heaviness, tears, lethargy, and withdrawal from social interaction.                   &  \\
Awe             & A positive, low energy state of feeling small but safe. Reaction to vastness or grandeur that transcends current understanding. Goosebumps and quiet reverence.                             &  \\
Anger           & A negative, high energy state of antagonism. Reaction to perceived injustice, interference, or threat. Increased heart rate, glaring, and impulse to attack or defend.                      &  \\
Excitement      & A positive, high energy state with a feeling of empowerment. Anticipation of a stimulating future event. High physical energy, eagerness, and focused attention on upcoming rewards.        &  \\
Amusement       & A positive, moderate energy state with a sense of safety. Response to humor or incongruity. Lighthearted laughter and playfulness without serious stakes.                                   &  \\
Fear            & A negative, high energy state of immediate threat response. Reaction to imminent danger. Adrenaline rush, freezing or fleeing, and focused attention on the threat.                         &  \\ \hline
\end{tabular}%
}
\label{tab: emotion defination}
\end{table}

\section{Implementation Details} \label{supp sec: implementation details}

\subsection{Detailed Information of Compared Models} \label{supp: models}
\textbf{Deepseek-VL-chat} \cite{deepseek-vl} is a vision-language model that integrates an efficient visual encoder with a large-scale language model through aligned multimodal training. The model demonstrates strong capabilities in processing high-resolution visual inputs and performing complex semantic understanding, making it particularly suitable for dialogue scenarios requiring fine-grained visual reasoning.

\textbf{Janus-Pro} \cite{janus} is a unified multimodal model capable of handling both visual understanding and content generation tasks simultaneously. Its core design addresses task conflict in a single encoder by decoupling visual encoding into distinct pathways for understanding and generation. Built on a unified Transformer \cite{transformer} architecture, Janus-pro demonstrates significantly enhanced multimodal understanding and instruction-following capabilities.

\textbf{mPLUG-Owl3} \cite{mplug3} is a vision-language model designed for long-sequence multimodal inputs. Its core innovation is the Hyper Attention mechanism, which enables deep fusion of visual and linguistic representations within a unified semantic space. This architecture allows visual embeddings to be processed independently of the text token sequence, thereby significantly enhancing the model's inference throughput when handling multiple images and long videos.

\textbf{LLaVA-OneVision} \cite{llava-ov} is a unified general-purpose vision-language model. By utilizing a single visual encoder and projection layer, it demonstrates competitive performance across $3$ core domains: single-image understanding, multi-image reasoning, and video question answering, thereby extending the capability boundaries of open-source multimodal models in cross-modal applications.

\textbf{LLaVA-OneVision-1.5} \cite{llava-ov-1.5} is an open-source implementation of the LLaVA-OneVision series, notable for providing a fully reproducible pipeline. By integrating an upgraded RICE-ViT visual encoder \cite{RICE_VIT} and a three-stage training protocol, LLaVA-OneVision-1.5 achieves performance superior to models of comparable scale on multiple multimodal benchmarks, alongside a substantially reduced training cost.

\textbf{LLaVA-NEXT} \cite{llava-next} represents a major update and open-source implementation of the LLava series. Adhering to a minimalist design, it integrates more capable base models like LLaMA-3 \cite{llama3} for enhanced overall performance. Its core innovation is a unified interleaved data format and training framework, enabling a single model to master diverse modalities-from single and multiple images to video and 3D data-and achieve state-of-the-art results among open-source models on relevant benchmarks.

\textbf{InternVL3.5} \cite{internvl3.5} is a high-performance open-source variant of the InternVL series. It adopts a native multimodal pre-training approach, jointly processing text and multimodal data from the beginning of pre-training. This methodology enables the model to achieve strong performance on tasks such as multidisciplinary reasoning (MMMU) and multi-image/video comprehension among current open-source multimodal models.

\textbf{Qwen2-VL} \cite{qwen2} is a large-scale vision-language model capable of processing images, videos, and text. The architecture incorporates key technologies including dynamic resolution handling and multimodal rotary position embedding (M-RoPE), endowing the model with strong capabilities in optical character recognition, long-video comprehension, and multilingual dialogue, alongside support
for building robust embodied agents.

\textbf{Qwen2.5-VL} \cite{Qwen2.5} is the next-generation flagship model in the Qwen vision-language series. Its architectural enhancements, including the introduction of absolute time encoding, substantially improve the modeling of video temporal dynamics and the parsing of long-document structures, establishing a leading position on multiple vision-language understanding benchmarks.

\textbf{Qwen3-VL} \cite{qwen3} is a large vision-language model constructed upon the new Qwen3 language foundation. It inherits the series' strengths in areas like dynamic resolution processing, and through enhanced foundational language and visual representation capabilities, achieves significant overall performance gains in tasks requiring complex cross-lingual reasoning, fine-grained visual grounding, and multi-turn interaction.

\textbf{GPT-4o} \cite{gpt4} is a natively MLLM, designed to process and generate text, image, and audio inputs in an end-to-end manner. It achieves deep cross-modal feature fusion through a shared Transformer \cite{transformer} architecture and responds at near real-time speeds. The model establishes significant advances in visual and audio understanding tasks while maintaining performance comparable to its predecessor on conventional text and coding benchmarks.

\textbf{GPT-5} \cite{gpt5} is the latest generation of large language models from OpenAI. At its core lies a native multimodal system that seamlessly integrates text, images, and audio. Distinctively, it introduces a real-time task routing mechanism that dynamically allocates tasks between a fast-response mode and an in-depth thinking mode based on complexity, thereby unifying intelligent reasoning with high efficiency. This architecture enables significant performance leaps in tasks involving complex reasoning and cross-modal understanding.

\textbf{Qwen-VL-Max} \cite{qwen-vl} is the preeminent closed-source flagship model of the Qwen series, integrating massive-scale parameters and a Mixture-of-Experts (MoE) design within a unified architecture. Qwen-VL-Max exhibits superior capabilities in multimodal understanding and generation, ultra-long-context handling, and complex reasoning. It achieves state-of-the-art results across a wide range of tasks, from conventional text and code benchmarks to advanced applications requiring deep document analysis and long-video understanding.

\textbf{Claude-3.7-Sonnet} \cite{claude} is a high-performance model that employs a shared fast-slow dual-system architecture, designed to balance immediate responsiveness with deep-thinking and high-confidence outputs. This architectural approach enables it to demonstrate markedly enhanced capabilities when handling tasks that require complex reasoning, long-context analysis, and code generation.

\textbf{Gemini-2.5-Pro} \cite{gemini2.5} is a next-generation MLLM with a context window of up to 2 million tokens. This enables the processing of extensive inputs, such as hours of video, large codebases, or lengthy documents, in a single pass. Leveraging a MoE architecture, Gemini-2.5-Pro shows improved performance across complex reasoning, cross-modal understanding, and long-document analysis tasks.

\textbf{Emotion-Qwen} \cite{Emotionqwen} is a large multimodal model designed to unify fine-grained emotion understanding and general vision-language reasoning. To address the limitations of existing LMMs in emotion tasks and catastrophic forgetting from specialized fine-tuning, it employs a MoE-based Hybrid Compressor that dynamically balances emotion-specific and general feature processing via a gating network. The model integrates a Facial Emotion Capture (FEC) module and is trained with a three-stage pre-training, multi-LoRA fine-tuning strategy and the VER dataset, achieving leading and balanced performance on both emotion benchmarks like DFEW~\cite{jiang2020dfew} and EMER~\cite{lian2023explainable}, and general benchmarks like MMBench~\cite{liu2024mmbench}.

\textbf{EmoVIT} \cite{EmoVIT} is a multimodal large model designed for emotion understanding through visual instruction tuning. Based on the InstructBLIP~\cite{dai2023instructblip} architecture, it leverages large-scale emotion-aligned visual instruction data generated via GPT-4~\cite{gpt4}, covering categorical, conversational, and reasoning types. The model employs a selective fine-tuning strategy, updating only the instruction-aware Q-Former module \cite{Q-former} while keeping both the visual encoder and the language model frozen. This approach significantly enhances performance on emotion classification, affective reasoning, and humor comprehension tasks, and demonstrates superior instruction sensitivity and robustness across multiple benchmark datasets.

\textbf{EmoCaliber} \cite{EmoCaliber} is a MLLM designed for reliable visual emotion comprehension, with its core innovation being explicit confidence verbalization to address perceptual subjectivity. Built upon the Qwen2.5-VL~\cite{Qwen2.5} architecture, it follows a three-stage progressive training framework: first, SFT endows the model with structured reasoning capabilities; second, guided by semantic-probabilistic priors, the model learns to output confidence estimates; third, reinforcement learning (GRPO)~\cite{GRPO} jointly calibrates confidence expression and emotion prediction. Evaluated on the unified benchmark VECBench, EmoCaliber outperforms existing methods in both emotion classification and confidence calibration, providing an effective solution for trustworthy visual emotion comprehension.

\textbf{AffectGPT} \cite{AffectGPT} is an emotion-oriented MLLM designed for emotion understanding from audio, video, and text inputs. It introduces a pre-fusion operation to enhance multimodal integration outside of the LLM, employing a Q-Former-based \cite{Q-former} strategy that preserves temporal information and an attention-based strategy for efficient feature fusion. The model is trained on the MER-Caption dataset, which is constructed with a model-led, human-assisted annotation strategy. AffectGPT provides a unified framework for emotion-centric MLLMs, emphasizing multimodal characteristics in affective computing.

\textbf{R1-Omni} \cite{R1-Omni} is an omnimodal large language model optimized via Reinforcement Learning with Verifiable Reward (RLVR) for explainable emotion recognition. Built upon HumanOmni \cite{zhao2025humanomni}, the model integrates visual and audio cues, generating outputs containing both reasoning processes and emotion predictions. Training follows a two-stage strategy: initial fine-tuning on EMER dataset \cite{lian2023explainable} for basic reasoning, followed by RLVR optimization with GRPO on emotion recognition datasets. R1-Omni represents a reinforcement learning approach to multimodal emotion understanding with explicit reasoning mechanisms.

\subsection{Evaluation Criteria of Ranking Task} \label{supp: ranking evaluation}

For the ranking task, we adopt the \textbf{emotion ranking score} proposed in EEmo-Bench \cite{EEmobench} to quantify prediction performance, jointly accounting for the hit rate of main emotions and the ranking consistency among correctly identified emotions. The detailed computation of this score is presented as follows.
Let $E_{\text{GT}}^i=[e_1^i, e_2^i, e_3^i]$ denotes the ground-truth emotion ranking for the $i$-th image, and $E_{\text{MLLM}}^i=[e_1^{i\prime}, e_2^{i\prime}, e_3^{i\prime}]$ represent the ranking predicted by the MLLM. Both sequences are defined within the permutation space $E_{\text{GT}}^i,E_{\text{MLLM}}\in A_8^3 (e_{\text{ANG}}, e_{\text{DIS}}, e_{\text{FEA}}, e_{\text{JOY}}, e_{\text{NEU}}, e_{\text{SAD}}, e_{\text{SUR}}, \text{None})$. Accordingly, the ranking score $S_{\text{rank}}^i$ of each image is formulated as follows:

\begin{equation}
    \begin{split}
        S_{\text{rank}}^{i}= (\sum_{k=1}^{3}\sum_{j=1}^{3}w_{k}\mathbb{I} (e_{k}^{i},e_{j}^{i\prime}))_{\text{Scaled}}
    +\mathcal{W} (E_{\text{MLLM}}^{i},E_{\text{GT}}^{i})\cdot (\mathcal{K} (\mathcal{SP} (E_{\text{MLLM}}^{i},E_{\text{GT}}^{i}))_{\text{Scaled}},
    \end{split}
\end{equation}

where $\mathbb{I} (e_{k}^{i},e_{j}^{i})$, $\{w_{k}| k=1,2,3\}$, $\mathcal{SP} (\cdot)$, $\mathcal{K} (\cdot)$, and $\mathcal{W} (\cdot)$
share the definition in Equ. (\ref{equ:rank}),
$ (\cdot)_{\text{Scaled}}$ denotes the scaling function used to map values to a range of $50$. 
Subsequently, the final ranking result is obtained by averaging $S_{\text{rank}}^i$ across all images.

\subsection{Evaluation Criteria of Description Task} \label{supp: description evaluation}

We employ a \textit{five-round Deepseek-assisted} evaluation protocol to quantify MLLM performance on the description task. Specifically, we define $4$ evaluation dimensions: \textbf{completeness, precision, relevance, and conciseness}. Notably, we decompose the aggregate evaluation scheme used in EEmo-Bench \cite{EEmobench} into these finer-grained criteria, allowing for clearer score differentiation and preventing extreme outliers in a single dimension from skewing the overall assessment. Furthermore, we introduce Conciseness to mitigate the ``verbosity bias" inherent in LLMs \cite{verbosity_bias}, where longer responses often artificially inflate completeness scores. This dimension effectively counterbalances such bias, promoting responses that align more closely with human-like reasoning styles. The specific scoring prompts and rules are detailed below.

For dimensions that require LLM-based evaluation, we first provide \textit{Deepseek} \cite{deepseek} with the following prompt to establish the overall evaluation criteria, together with the open-ended \textbf{question}, the model-generated \textbf{response}, and the human-annotated \textbf{golden answer}.

{\small \textit{User: \\
You are an expert evaluator for open-ended emotional response analysis. Your task is to evaluate the quality of the model-generated emotional response $<$response$>$ to the question $<$question$>$
with respect to the human-written reference (golden answer) $<$golden answer$>$,  
based on the following $3$ metrics: Completeness, Preciseness, and Relevance.  
Each metric should be scored as 0, 1, or 2 according to the detailed criteria below.  
Please output ONLY the three final scores in the specified format.  
}}

For the \textbf{completeness} dimension, we further apply the following prompt to the LLM for $5$ consecutive rounds to obtain scores, which are then averaged to produce the Completeness score $S_{\text{comp}}$. 

{\small \textit{
1. Completeness: \\ 
Evaluate whether the model’s answer completely captures the emotional reasoning, emotional dimensions (Valence, Arousal, Dominance if applicable), and key explanatory elements contained in the golden answer. \\
**Scoring criteria:**
- **Score 2:** The answer completely or almost completely covers all main emotional insights, reasoning steps, and key ideas expressed in the golden answer. The reasoning flow is coherent and sufficiently detailed.
- **Score 1:** The answer includes only part of the emotional content or reasoning in the golden answer, or expresses similar ideas but lacks some emotional nuance or reasoning elements.
- **Score 0:** The answer omits most of the key emotional content or reasoning, or is largely incomplete or off-topic compared with the golden answer. \\
**Output format:**  
Completeness: [0|1|2]}}

\begin{table*}[]
\centering
\scriptsize
\setlength{\tabcolsep}{3.3pt}        
\caption{Details comparison on perception task.}
\begin{tabular}{lccccccccccc}
\hline
\multicolumn{1}{c|}{Sub-categories}            & \multicolumn{4}{c|}{Perceptual Dimension}                                                      & \multicolumn{6}{c|}{Content Category}                                                                                                                                                                                                                                                                                   & \multirow{2}{*}{Overall↑} \\ \cline{1-11}
\multicolumn{1}{c|}{Model}                     & Emotion↑         & Valence↑         & Arousal↑         & \multicolumn{1}{c|}{Dominance↑}       & Animal↑           & \begin{tabular}[c]{@{}c@{}}Abstract↑\\ /Cartoon Image↑\end{tabular} & \begin{tabular}[c]{@{}c@{}}Daily Life \\ Scene↑\end{tabular} & Human↑            & \begin{tabular}[c]{@{}c@{}}Stationary\\ Object↑\end{tabular} & \multicolumn{1}{c|}{\begin{tabular}[c]{@{}c@{}}Natural \\ Landscape↑\end{tabular}} &                          \\ \hline
\multicolumn{1}{c|}{Random guess}              & 39.38\%          & 44.62\%          & 44.72\%          & \multicolumn{1}{c|}{44.44\%}          & 41.67\%          & 41.67\%                                                           & 41.67\%                                                     & 41.67\%          & 41.67\%                                                     & \multicolumn{1}{c|}{41.67\%}                                                      & 41.83\%                  \\ \hline
\multicolumn{12}{l}{\textcolor{gray}{\textit{Medium-scale   open-source MLLMs}}}                                                                                                                                                                                                                                                                                                                                                                                                                                       \\ \hdashline
\multicolumn{1}{l|}{Deepseek-VL-7B-chat}       & 51.30\%          & 59.41\%          & 56.36\%          & \multicolumn{1}{c|}{43.93\%}          & 52.82\%          & 54.98\%                                                           & 57.45\%                                                     & 55.08\%          & 52.31\%                                                     & \multicolumn{1}{c|}{59.11\%}                                                      & 51.98\%                  \\
\multicolumn{1}{l|}{Janus-Pro-7B}              & 50.20\%          & 56.28\%          & 54.53\%          & \multicolumn{1}{c|}{57.67\%}          & 53.08\%          & 50.96\%                                                           & 55.30\%                                                     & 50.24\%          & 55.00\%                                                     & \multicolumn{1}{c|}{55.37\%}                                                      & 51.53\%                  \\
\multicolumn{1}{l|}{mPLUG-owl3-7B}             & 54.06\%          & 67.89\%          & 58.41\%          & \multicolumn{1}{c|}{48.01\%}          & 51.88\%          & 52.68\%                                                           & 59.17\%                                                     & 54.61\%          & 58.65\%                                                     & \multicolumn{1}{c|}{60.31\%}                                                      & 56.14\%                  \\
\multicolumn{1}{l|}{LLaVA-OneVision-7B}        & 57.62\%          & 72.18\%          & 64.87\%          & \multicolumn{1}{c|}{51.87\%}          & 59.38\%          & 61.88\%                                                           & 64.04\%                                                     & 63.83\%          & 62.50\%                                                     & \multicolumn{1}{c|}{61.84\%}                                                      & 60.32\%                  \\
\multicolumn{1}{l|}{LLaVA-OneVision-1.5-8B}    & 59.17\%          & 69.67\%          & 59.48\%          & \multicolumn{1}{c|}{46.31\%}          & 57.51\%          & 60.54\%                                                           & 64.61\%                                                     & 59.34\%          & 62.50\%                                                     & \multicolumn{1}{c|}{66.50\%}                                                      & 58.56\%                  \\
\multicolumn{1}{l|}{LLaVA-NEXT-8B}             & 53.13\%          & 63.39\%          & 53.34\%          & \multicolumn{1}{c|}{47.90\%}          & 55.90\%          & 52.87\%                                                           & 60.60\%                                                     & 49.17\%          & 59.04\%                                                     & \multicolumn{1}{c|}{64.91\%}                                                      & 54.22\%                  \\
\multicolumn{1}{l|}{InternVL3.5-8B}            & 54.50\%          & 65.48\%          & 59.59\%          & \multicolumn{1}{c|}{46.54\%}          & 51.14\%          & 56.51\%                                                           & 60.60\%                                                     & 56.45\%          & 57.50\%                                                     & \multicolumn{1}{c|}{62.69\%}                                                      & 55.81\%                  \\
\multicolumn{1}{l|}{Qwen2-VL-7B}               & 58.62\%          & 66.42\%          & 56.36\%          & \multicolumn{1}{c|}{48.24\%}          & 57.77\%          & 60.34\%                                                           & 60.89\%                                                     & 59.46\%          & 60.00\%                                                     & \multicolumn{1}{c|}{60.14\%}                                                      & 57.99\%                  \\
\multicolumn{1}{l|}{Qwen2.5-VL-7B}             & 58.22\%          & 59.56\%          & 53.15\%          & \multicolumn{1}{c|}{49.72\%}          & 54.69\%          & 61.88\%                                                           & 64.47\%                                                     & 61.82\%          & 62.88\%                                                     & \multicolumn{1}{c|}{59.97\%}                                                      & 59.54\%                  \\
\multicolumn{1}{l|}{Qwen3-VL-8b}               & 60.05\%          & 74.37\%          & 65.62\%          & \multicolumn{1}{c|}{51.70\%}          & 56.49\%          & 64.42\%                                                           & 64.70\%                                                     & 64.45\%          & 63.95\%                                                     & \multicolumn{1}{c|}{66.27\%}                                                      & 61.41\%                  \\ \hline
\multicolumn{12}{l}{\textcolor{gray}{\textit{Large-scale   open-source MLLMs}}}                                                                                                                                                                                                                                                                                                                                                                                                                                        \\ \hdashline
\multicolumn{1}{l|}{LLaVA-OneVision-72B}       & 64.65\%          & 74.27\%          & 64.22\%          & \multicolumn{1}{c|}{52.78\%}          & 62.47\%          & 64.37\%                                                           & 69.77\%                                                     & 62.41\%          & 67.69\%                                                     & \multicolumn{1}{c|}{67.80\%}                                                      & 64.16\%                  \\
\multicolumn{1}{l|}{Qwen2-VL-72B}              & 63.75\%          & 74.48\%          & 64.98\%          & \multicolumn{1}{c|}{51.87\%}          & 61.53\%          & 67.24\%                                                           & 66.33\%                                                     & 60.05\%          & 69.69\%                                                     & \multicolumn{1}{c|}{67.12\%}                                                      & 63.83\%                  \\
\multicolumn{1}{l|}{Qwen2.5-VL-72B}            & 63.81\%          & 75.00\%          & 65.30\%          & \multicolumn{1}{c|}{48.92\%}          & 59.38\%          & 63.60\%                                                           & 66.76\%                                                     & 61.23\%          & 69.23\%                                                     & \multicolumn{1}{c|}{67.46\%}                                                      & 63.39\%                  \\ \hline
\multicolumn{12}{l}{\textcolor{gray}{\textit{Proprietary MLLMs}}}                                                                                                                                                                                                                                                                                                                                                                                                                                                      \\ \hdashline
\multicolumn{1}{l|}{GPT-4o}                    & \textbf{65.41\%} & 75.52\%          & 65.62\%          & \multicolumn{1}{c|}{54.37\%}          & 63.27\%          & 64.75\%                                                           & 69.48\%                                                     & 64.66\%          & 68.27\%                                                     & \multicolumn{1}{c|}{69.85\%}                                                      & 65.31\%                  \\
\multicolumn{1}{l|}{GPT-5}                     & 64.18\%          & 74.24\%          & 59.30\%          & \multicolumn{1}{c|}{53.08\%}          & {\ul 63.40\%}    & 62.98\%                                                           & 67.62\%                                                     & 53.54\%          & 67.88\%                                                     & \multicolumn{1}{c|}{{\ul 70.92\%}}                                                      & 63.38\%                  \\
\multicolumn{1}{l|}{Qwen-VL-Max}               & 63.06\%          & 74.87\%          & 65.33\%          & \multicolumn{1}{c|}{47.71\%}          & 59.65\%          & 64.04\%                                                           & 65.57\%                                                     & 61.10\%          & 68.09\%                                                     & \multicolumn{1}{c|}{66.44\%}                                                      & 62.88\%                  \\
\multicolumn{1}{l|}{Claude-3.7-Sonnet}         & 60.69\%          & 73.82\%          & 61.42\%          & \multicolumn{1}{c|}{52.05\%}          & 57.37\%          & 57.97\%                                                           & 64.85\%                                                     & 58.27\%          & 62.74\%                                                     & \multicolumn{1}{c|}{69.85\%}                                                      & 61.61\%                  \\
\multicolumn{1}{l|}{Gemini-2.5-Pro}            & 63.63\%          & \textbf{80.52\%} & \textbf{73.28\%} & \multicolumn{1}{c|}{55.96\%}    & 63.27\%          & \textbf{70.33\%}                                                  & \textbf{70.30\%}                                            & {\ul 68.52\%}    & \textbf{72.12\%}                                            & \multicolumn{1}{c|}{67.52\%}                                               & {\ul 66.79\%}            \\ \hline
\multicolumn{12}{l}{\textcolor{gray}{\textit{Emotion-Oriented   MLLMs}}}                                                                                                                                                                                                                                                                                                                                                                                                                                               \\ \hdashline
\multicolumn{1}{l|}{Emotion-Qwen}              & 64.06\%          & 69.98\%          & 57.76\%          & \multicolumn{1}{c|}{{\ul 64.02\%}}          & 60.46\%          & 62.45\%                                                           & 63.04\%                                                     & 62.53\%          & 60.19\%                                                     & \multicolumn{1}{c|}{64.12\%}                                                      & 61.67\%                  \\
\multicolumn{1}{l|}{EmoVIT}                    & 34.29\%          & 36.58\%          & 35.78\%          & \multicolumn{1}{c|}{31.59\%}          & 44.64\%          & 51.53\%                                                           & 47.56\%                                                     & 46.34\%          & 48.65\%                                                     & \multicolumn{1}{c|}{53.40\%}                                                      & 33.77\%                  \\
\multicolumn{1}{l|}{AffectGPT}                 & 42.20\%          & 41.42\%          & 47.52\%          & \multicolumn{1}{c|}{46.08\%}          & 40.75\%          & 45.02\%                                                           & 43.55\%                                                     & 44.56\%          & 42.69\%                                                     & \multicolumn{1}{c|}{44.39\%}                                                      & 43.73\%                  \\
\multicolumn{1}{l|}{R1-Omni-0.5B}              & 41.21\%          & 47.80\%          & 50.05\%          & \multicolumn{1}{c|}{43.64\%}          & 45.04\%          & 43.87\%                                                           & 46.13\%                                                     & 46.51\%          & 47.49\%                                                     & \multicolumn{1}{c|}{39.42\%}                                                      & 44.02\%                  \\
\multicolumn{1}{l|}{EmoCaliber}                & 58.24\%          & 76.24\%          & 64.07\%          & \multicolumn{1}{c|}{49.31\%}          & 56.53\%          & 62.96\%                                                           & 66.62\%                                                     & 60.59\%          & 61.15\%                                                     & \multicolumn{1}{c|}{66.04\%}                                                      & 60.67\%                  \\
\multicolumn{1}{l|}{\textbf{EEmo-Logic (Ours)}} & {\ul 65.06\%}    & {\ul 78.45\%}    & {\ul 67.24\%}    & \multicolumn{1}{c|}{\textbf{72.64\%}} & \textbf{67.96\%} & {\ul 69.35\%}                                                     & {\ul 70.20\%}                                               & \textbf{69.74\%} & {\ul 70.00\%}                                               & \multicolumn{1}{c|}{\textbf{73.64\%}}                                             & \textbf{68.54\%}         \\ \hline
\end{tabular}
\label{tab: Supp perception}
\end{table*}

For the \textbf{precision} dimension, we administer the following prompt to the LLM over $5$ consecutive rounds. The resulting scores are then aggregated and averaged to derive the final Completeness score, denoted as $S_{\text{prec}}$.

{\small \textit{
2. Precision: \\
Evaluate how precisely the emotional reasoning in response aligns with the golden answer in terms of correctness, logic, and interpretive accuracy. \\
**Scoring criteria:**
- **Score 2:**  
  The answer is logically consistent with the golden answer.  
  For example, it correctly describes emotional polarity and intensity, 
  and the comparative judgments between images (if present) are accurate.  
  The reasoning chain follows a similar logic or interpretive framework as the golden answer.  
  While the length or phrasing may differ, no contradictory statements, invented reasoning, or factual inconsistencies appear.  
  Answers with similar length and coverage to the golden answer can be considered stronger examples of precision.
- **Score 1:**  
  The answer is mostly consistent but contains minor logical deviations or vague expressions.  
  Some reasoning steps may differ, or comparisons (e.g., between Image 1 and Image 2) may lack clarity.  
  The emotional direction (e.g., polarity in VAD) remains generally correct, though less detailed or partially inconsistent.  
  The reasoning chain may partially diverge or oversimplify the logic in the golden answer.
- **Score 0:**  
  The answer contains clearly incorrect or contradictory reasoning relative to the golden answer.  
  Examples include reversed VAD polarity (e.g., calling a low valence “positive”), wrong comparative conclusions between images, or fabricated reasoning absent from the reference.  
  Logical flow significantly differs or introduces invented interpretations (“hallucinations”) not supported by the reference reasoning.  
  Such deviations indicate imprecision and warrant the lowest score. \\
**Output format:**  
Preciseness: [0|1|2] }}

For \textbf{relevance}, we query the LLM across $5$ rounds using the prompt below. The outputs are averaged to yield the final score $S_{\text{rele}}$.

{\small \textit{
3. Relevance \\
Evaluate whether the model’s response is **relevant to the emotions evoked by the image** and aligned with the emotional focus of the question. \\
**Scoring criteria:**
- **Score 2:**  
  The answer is strongly connected to the **emotions evoked by the image**,  
  addressing aspects such as **VAD dimensions (Valence, Arousal, Dominance)** or **Ekman’s six basic emotions** (anger, fear, disgust, joy, sadness, surprise and neutral).  
  It also aligns with the question’s emotional focus.  
  The analysis may describe emotional expression in the image, but the reasoning ultimately returns to how the image **evokes emotions or emotional resonance in viewers**.  
  The emotional interpretation is directly relevant and contextually grounded.
- **Score 1:**  
  The answer is partly relevant — it discusses emotions but drifts between describing the **subject’s expression** and **the viewer’s emotional response**,  
  without fully connecting the two.  
  The emotional content may partially match the question’s focus.  
  The reasoning touches on affective content but lacks a clear emphasis on **viewer-evoked emotion**.
- **Score 0:**  
  The answer is weakly or not relevant — it primarily describes visual features, character actions, or facial expressions  
  without relating them to how the image **evokes emotions in the viewer**.  
  Responses that misinterpret the asking question,  
  or fail to align with the question’s emotional emphasis, are considered irrelevant. \\
**Output format:**  
Relevance: [0|1|2] }}

\begin{table*}[]
\centering
\scriptsize
\setlength{\tabcolsep}{2pt}        
\caption{Details comparison on description task.}
\begin{tabular}{lccccccccccc}
\hline
\multicolumn{1}{c|}{Sub-categories}            & \multicolumn{5}{c|}{Single}                                                                                                                                                                                                                                                                                                  & \multicolumn{5}{c|}{Pair}                                                                                                                                                                                                                                                                                                 & \multirow{2}{*}{Overall↑} \\ \cline{1-11}
\multicolumn{1}{c|}{Model}                     & \begin{tabular}[c]{@{}c@{}}Detailed\\ Description↑\end{tabular} & \begin{tabular}[c]{@{}c@{}}Reasoning\\ Description↑\end{tabular} & \begin{tabular}[c]{@{}c@{}}Direct\\ -Indirect↑\end{tabular} & \multicolumn{1}{c|}{\begin{tabular}[c]{@{}c@{}}Conflicts\\ Description↑\end{tabular}} & \multicolumn{1}{c|}{Overall↑}          & \begin{tabular}[c]{@{}c@{}}Detailed\\ Comparison↑\end{tabular} & \begin{tabular}[c]{@{}c@{}}Valence\\ Comparison↑\end{tabular} & \begin{tabular}[c]{@{}c@{}}Arousal\\ Comparison↑\end{tabular} & \multicolumn{1}{c|}{\begin{tabular}[c]{@{}c@{}}Dominance\\ Comparison↑\end{tabular}} & \multicolumn{1}{c|}{Overall↑}          &                          \\ \hline
\multicolumn{12}{l}{\textcolor{gray}{\textit{Medium-scale   open-source MLLMs}}}                                                                                                                                                                                                                                                                                                                                                                                                                                                                                                                                                                                                                                                                       \\ \hdashline 
\multicolumn{1}{l|}{Deepseek-VL-7B-chat}       & 46.60\%                                                        & 42.21\%                                                         & 44.89\%                                                    & \multicolumn{1}{c|}{41.93\%}                                                         & \multicolumn{1}{c|}{43.95\%}          & 39.11\%                                                       & 53.07\%                                                      & 46.67\%                                                      & \multicolumn{1}{c|}{41.85\%}                                                        & \multicolumn{1}{c|}{45.10\%}          & 44.38\%                  \\
\multicolumn{1}{l|}{Janus-Pro-7B}              & 54.00\%                                                        & 49.24\%                                                         & 49.13\%                                                    & \multicolumn{1}{c|}{44.12\%}                                                         & \multicolumn{1}{c|}{49.48\%}          & 47.97\%                                                       & 50.67\%                                                      & 49.17\%                                                      & \multicolumn{1}{c|}{45.37\%}                                                        & \multicolumn{1}{c|}{46.97\%}          & 48.54\%                  \\
\multicolumn{1}{l|}{mPLUG-owl3-7B}             & 48.47\%                                                        & 48.14\%                                                         & {\ul 56.30\%}                                              & \multicolumn{1}{c|}{44.82\%}                                                         & \multicolumn{1}{c|}{48.98\%}          & 49.62\%                                                       & 60.40\%                                                      & 53.83\%                                                      & \multicolumn{1}{c|}{50.74\%}                                                        & \multicolumn{1}{c|}{53.50\%}          & 50.67\%                  \\
\multicolumn{1}{l|}{LLaVA-OneVision-7B}        & 46.87\%                                                        & 42.90\%                                                         & 44.35\%                                                    & \multicolumn{1}{c|}{40.09\%}                                                         & \multicolumn{1}{c|}{43.71\%}          & 47.85\%                                                       & 49.47\%                                                      & 49.50\%                                                      & \multicolumn{1}{c|}{41.11\%}                                                        & \multicolumn{1}{c|}{47.00\%}          & 44.94\%                  \\
\multicolumn{1}{l|}{LLaVA-OneVision-1.5-8B}    & 51.13\%                                                        & 49.83\%                                                         & 53.59\%                                                    & \multicolumn{1}{c|}{49.04\%}                                                         & \multicolumn{1}{c|}{50.74\%}          & 50.38\%                                                       & 56.27\%                                                      & 47.50\%                                                      & \multicolumn{1}{c|}{36.11\%}                                                        & \multicolumn{1}{c|}{44.43\%}          & 48.38\%                  \\
\multicolumn{1}{l|}{LLaVA-NEXT-8B}             & 48.00\%                                                        & 37.86\%                                                         & 45.43\%                                                    & \multicolumn{1}{c|}{45.79\%}                                                         & \multicolumn{1}{c|}{44.09\%}          & 39.49\%                                                       & 29.87\%                                                      & 24.83\%                                                      & \multicolumn{1}{c|}{27.04\%}                                                        & \multicolumn{1}{c|}{31.37\%}          & 39.33\%                  \\
\multicolumn{1}{l|}{InternVL3.5-8B}            & 46.53\%                                                        & 44.76\%                                                         & 39.89\%                                                    & \multicolumn{1}{c|}{44.74\%}                                                         & \multicolumn{1}{c|}{44.39\%}          & 46.20\%                                                       & 47.60\%                                                      & 47.33\%                                                      & \multicolumn{1}{c|}{40.93\%}                                                        & \multicolumn{1}{c|}{43.43\%}          & 44.03\%                  \\
\multicolumn{1}{l|}{Qwen2-VL-7B}               & 50.80\%                                                        & 47.52\%                                                         & 53.26\%                                                    & \multicolumn{1}{c|}{42.54\%}                                                         & \multicolumn{1}{c|}{48.42\%}          & 45.95\%                                                       & 60.93\%                                                      & 48.83\%                                                      & \multicolumn{1}{c|}{42.04\%}                                                        & \multicolumn{1}{c|}{49.43\%}          & 48.80\%                  \\
\multicolumn{1}{l|}{Qwen2.5-VL-7B}             & 49.53\%                                                        & 44.07\%                                                         & 41.85\%                                                    & \multicolumn{1}{c|}{45.79\%}                                                         & \multicolumn{1}{c|}{45.69\%}          & 43.42\%                                                       & 46.93\%                                                      & 52.33\%                                                      & \multicolumn{1}{c|}{40.56\%}                                                        & \multicolumn{1}{c|}{45.73\%}          & 45.70\%                  \\
\multicolumn{1}{l|}{Qwen3-VL-8b}               & 47.27\%                                                        & 44.97\%                                                         & 34.67\%                                                    & \multicolumn{1}{c|}{41.84\%}                                                         & \multicolumn{1}{c|}{43.05\%}          & 46.08\%                                                       & 45.20\%                                                      & 44.17\%                                                      & \multicolumn{1}{c|}{30.00\%}                                                        & \multicolumn{1}{c|}{42.43\%}          & 42.82\%                  \\ \hline
\multicolumn{12}{l}{\textcolor{gray}{\textit{Large-scale   open-source MLLMs}}}                                                                                                                                                                                                                                                                                                                                                                                                                                                                                                                                                                                                                                                                        \\ \hdashline 
\multicolumn{1}{l|}{LLaVA-OneVision-72B}       & 52.73\%                                                        & 51.93\%                                                         & 52.17\%                                                    & \multicolumn{1}{c|}{52.54\%}                                                         & \multicolumn{1}{c|}{52.36\%}          & 51.90\%                                                       & {\ul 68.40\%}                                                & 55.00\%                                                      & \multicolumn{1}{c|}{50.56\%}                                                        & \multicolumn{1}{c|}{56.60\%}          & 53.95\%                  \\
\multicolumn{1}{l|}{Qwen2-VL-72B}              & 50.60\%                                                        & 45.31\%                                                         & 45.87\%                                                    & \multicolumn{1}{c|}{46.32\%}                                                         & \multicolumn{1}{c|}{47.23\%}          & 49.37\%                                                       & 49.07\%                                                      & 49.33\%                                                      & \multicolumn{1}{c|}{45.37\%}                                                        & \multicolumn{1}{c|}{48.10\%}          & 47.56\%                  \\
\multicolumn{1}{l|}{Qwen2.5-VL-72B}            & 49.33\%                                                        & 45.10\%                                                         & 38.37\%                                                    & \multicolumn{1}{c|}{45.96\%}                                                         & \multicolumn{1}{c|}{45.33\%}          & 46.71\%                                                       & 49.33\%                                                      & 49.17\%                                                      & \multicolumn{1}{c|}{46.67\%}                                                        & \multicolumn{1}{c|}{47.50\%}          & 46.14\%                  \\ \hline
\multicolumn{12}{l}{\textcolor{gray}{\textit{Proprietary MLLMs}}}                                                                                                                                                                                                                                                                                                                                                                                                                                                                                                                                                                                                                                                                                      \\ \hdashline 
\multicolumn{1}{l|}{GPT-4o}                    & \textbf{61.13\%}                                               & 54.97\%                                                         & 53.37\%                                                    & \multicolumn{1}{c|}{52.54\%}                                                         & \multicolumn{1}{c|}{{\ul 55.97\%}}    & {\ul 55.95\%}                                                 & 58.67\%                                                      & 57.33\%                                                      & \multicolumn{1}{c|}{55.74\%}                                                        & \multicolumn{1}{c|}{56.57\%}          & 56.19\%                  \\
\multicolumn{1}{l|}{GPT-5}                     & 59.32\%                                                        & 46.76\%                                                         & \textbf{60.87\%}                                           & \multicolumn{1}{c|}{{\ul 56.05\%}}                                                   & \multicolumn{1}{c|}{55.21\%}          & 49.74\%                                                       & 57.57\%                                                      & {\ul 65.76\%}                                                & \multicolumn{1}{c|}{\textbf{73.33\%}}                                               & \multicolumn{1}{c|}{{\ul 58.99\%}}    & {\ul 56.63\%}            \\
\multicolumn{1}{l|}{Qwen-VL-Max}               & 46.73\%                                                        & 43.79\%                                                         & 35.57\%                                                    & \multicolumn{1}{c|}{41.75\%}                                                         & \multicolumn{1}{c|}{43.24\%}          & 44.43\%                                                       & 46.22\%                                                      & 46.10\%                                                      & \multicolumn{1}{c|}{47.74\%}                                                        & \multicolumn{1}{c|}{45.72\%}          & 44.17\%                  \\
\multicolumn{1}{l|}{Claude-3.7-Sonnet}         & 49.60\%                                                        & {\ul 55.17\%}                                                   & 51.85\%                                                    & \multicolumn{1}{c|}{49.39\%}                                                         & \multicolumn{1}{c|}{51.58\%}          & \textbf{56.46\%}                                              & 50.27\%                                                      & 51.17\%                                                      & \multicolumn{1}{c|}{52.59\%}                                                        & \multicolumn{1}{c|}{52.07\%}          & 51.76\%                  \\
\multicolumn{1}{l|}{Gemini-2.5-Pro}            & 47.67\%                                                        & 46.88\%                                                         & 34.84\%                                                    & \multicolumn{1}{c|}{49.20\%}                                                         & \multicolumn{1}{c|}{45.44\%}          & 43.42\%                                                       & 45.20\%                                                      & 38.33\%                                                      & \multicolumn{1}{c|}{36.67\%}                                                        & \multicolumn{1}{c|}{42.03\%}          & 44.16\%                  \\ \hline
\multicolumn{12}{l}{\textcolor{gray}{\textit{Emotion-Oriented   MLLMs}}}                                                                                                                                                                                                                                                                                                                                                                                                                                                                                                                                                                                                                                                                               \\ \hdashline 
\multicolumn{1}{l|}{Emotion-Qwen}              & 47.27\%                                                        & 32.41\%                                                         & 44.57\%                                                    & \multicolumn{1}{c|}{20.09\%}                                                         & \multicolumn{1}{c|}{36.29\%}          & 45.57\%                                                       & 38.53\%                                                      & 32.50\%                                                      & \multicolumn{1}{c|}{30.00\%}                                                        & \multicolumn{1}{c|}{36.67\%}          & 36.43\%                  \\
\multicolumn{1}{l|}{EmoVIT}                    & 46.47\%                                                        & 41.86\%                                                         & 48.59\%                                                    & \multicolumn{1}{c|}{40.96\%}                                                         & \multicolumn{1}{c|}{44.27\%}          & 30.00\%                                                       & 36.13\%                                                      & 35.00\%                                                      & \multicolumn{1}{c|}{25.00\%}                                                        & \multicolumn{1}{c|}{32.70\%}          & 39.94\%                  \\
\multicolumn{1}{l|}{AffectGPT}                 & 1.33\%                                                         & 0.14\%                                                          & 0.22\%                                                     & \multicolumn{1}{c|}{0.35\%}                                                          & \multicolumn{1}{c|}{0.56\%}           & 22.03\%                                                       & 9.33\%                                                       & 7.67\%                                                       & \multicolumn{1}{c|}{5.56\%}                                                         & \multicolumn{1}{c|}{12.20\%}          & 4.92\%                   \\
\multicolumn{1}{l|}{R1-Omni-0.5B}              & 10.93\%                                                        & 8.90\%                                                          & 3.59\%                                                     & \multicolumn{1}{c|}{7.19\%}                                                          & \multicolumn{1}{c|}{8.14\%}           & 27.47\%                                                       & 33.47\%                                                      & 37.00\%                                                      & \multicolumn{1}{c|}{28.70\%}                                                        & \multicolumn{1}{c|}{31.93\%}          & 17.05\%                  \\
\multicolumn{1}{l|}{EmoCaliber}                & 41.80\%                                                        & 37.17\%                                                         & 34.24\%                                                    & \multicolumn{1}{c|}{24.21\%}                                                         & \multicolumn{1}{c|}{35.07\%}          & 32.41\%                                                       & 40.40\%                                                      & 27.33\%                                                      & \multicolumn{1}{c|}{14.44\%}                                                        & \multicolumn{1}{c|}{29.67\%}          & 33.05\%                  \\
\multicolumn{1}{l|}{\textbf{EEmo-Logic (Ours)}} & {\ul 61.00\%}                                                  & \textbf{60.34\%}                                                & 55.00\%                                                    & \multicolumn{1}{c|}{\textbf{57.98\%}}                                                & \multicolumn{1}{c|}{\textbf{59.02\%}} & 51.90\%                                                       & \textbf{85.33\%}                                             & \textbf{69.50\%}                                             & \multicolumn{1}{c|}{{\ul 71.85\%}}                                                  & \multicolumn{1}{c|}{\textbf{67.40\%}} & \textbf{62.16\%}         \\ \hline
\end{tabular}
\label{tab: Supp description}
\end{table*}

For \textbf{conciseness}, we empirically observe that the LLM exhibits ambiguity and insensitivity to the prompt, often assigning identical scores to all responses. Therefore, we introduce a length-sensitive scoring function, denoted as $S_{\text{conc}}$. 
Let $L_{\text{gen}}$ and $L_{\text{gt}}$ represent the word counts of the generated response and the ground truth, respectively. The score is determined by the ratio of $L_{\text{gt}}$ to $L_{\text{gt}}$, rewarding responses that maintain an information density comparable to expert annotations:

\begin{equation}
    S_{\text{conc}} = \begin{cases} 
2, & \text{if } \frac{2}{3} L_{\text{gt}} \le L_{\text{gen}} \le 2 L_{\text{gt}} \\
1, & \text{if } \frac{1}{3} L_{\text{gt}} \le L_{\text{gen}} < \frac{2}{3} L_{\text{gt}} \quad \text{or} \quad 2 L_{\text{gt}} < L_{\text{gen}} \le 4 L_{\text{gt}} \\
0, & \text{otherwise}
\end{cases}
\end{equation}

Finally, we average and normalize the scores across the $4$ dimensions to obtain the final score $S_{\text{desc}}$ for each model response.

\begin{equation}
    S_{\text{desc}} = \frac{1}{2} \cdot \frac{1}{4} (S_{\text{comp}} + S_{\text{prec}} + S_{\text{rele}} + S_{\text{conc}}).
\end{equation}


\subsection{Probability-based VAD Scoring Method} \label{supp: softmax}

To enable a fair comparison of EEmo-Logic on the assessment task, we adopt the probability-based score generation method introduced in EEmo-Bench \cite{EEmobench} to convert the outputs of medium-scale open-source models into numerical scores. The formal definition is given as follows. 
First, emotional level keywords are defined for each VAD attribute: \textit{positive, neutral, negative} for \textbf{valence}; \textit{high, moderate, low} for \textbf{arousal}; and \textit{powerful, moderate, helpless} for \textbf{dominance}. These keywords are then mapped to a unified scale of \textit{high, medium, low}. Subsequently, softmax pooling is applied to the logits of the extracted keywords to derive probabilistic scores for each rating level:

\begin{equation}
    p_{l}=\frac{e^{x_{\text{SCORE\_TOKEN}}^l}}{\sum_{l}^{l\in\mathcal{L}}e^{x_{\text{SCORE\_TOKEN}}^l}},
\end{equation}

where $\mathcal{L}$ denotes the set of level keywords (\textit{high, medium, low}), while $x_{\text{SCORE\_TOKEN}}^l$ and $p_{l}$ represent the logits and probabilities corresponding to each level, respectively. Finally, the predicted rating $r_{\text{VAD}}$ is computed via a weighted average of $p_{l}$:

\begin{equation}
    r_{\text{VAD}} = \sum_{l}^{l\in\mathcal{L}}w_l^r\cdot p_l,
\end{equation}

where $w_l^r$ represents the numerical weight assigned to each rating level, with the values $w_l^r=\{1, 0.5, 0\}$ corresponding to \textit{high}, \textit{medium}, and \textit{low}, respectively.

\subsection{Comparison between LoRA Fine-tuning and Full Fine-tuning} \label{supp: sft setting comparison}

\begin{table*}[]
\centering
\footnotesize
\renewcommand{\arraystretch}{1.2}   
\setlength{\tabcolsep}{5pt}        
\caption{Comparison of experimental settings between LoRA SFT and FFT.}
\begin{tabular}{lllllll}
\hline
\textbf{Settings} & \textbf{LR}       & \textbf{Epoch} & \textbf{Global Batch Size} & \textbf{Weight Decay} & \textbf{LR Scheduler} & \textbf{Warmup Ratio} \\ \hline
LoRA SFT & 2.00E-04 & 1     & 128               & 0.1          & cosine       & 0.03         \\
FFT      & 1.00E-05 & 1     & 128               & 0.1          & cosine       & 0.03         \\ \hline
\end{tabular}
\label{tab: sft setting}
\end{table*}

\begin{table*}[]
\centering
\footnotesize
\renewcommand{\arraystretch}{1.2}   
\setlength{\tabcolsep}{5pt}        
\caption{Comparison of training processes between LoRA SFT and FFT.}
\begin{tabular}{lllllll}
\hline

Training Progress & 0\%       & 20\% & 40\% & 60\% & 80\% & 100\% \\ \hline
Training Loss of LoRA SFT & 1.098 & 0.415     & 0.365               & 0.374         & 0.397       & 0.370         \\
Training Loss of FFT      & 1.083 & 0.408     & 0.358               & 0.367          & 0.389       & 0.361         \\ \hline
\end{tabular}
\label{tab: sft process}
\end{table*}

We provide a detailed comparison of hyperparameters and training losses. As shown in Tab.~\ref{tab: sft setting} \& \ref{tab: sft process}, all settings remain identical except for the optimal learning rate (LR), since full fine-tuning (FFT) strictly requires a smaller LR to prevent representation collapse. Although FFT achieves a slightly lower training loss, it performs significantly worse on the unseen EEmo-Bench in Tab.~\ref{tab: Ablation}. This discrepancy indicates that FFT overfits; it memorizes the synthetic training data distribution and suffers from catastrophic forgetting of pre-trained knowledge. Conversely, the superior test performance of LoRA demonstrates that its low-rank constraint acts as an effective implicit regularization mechanism. Thus, LoRA reasonably outperforms FFT by preventing overfitting to synthetic noise while preserving general reasoning capabilities.

\section{More Results Comparison} \label{supp sec: more results}

\begin{figure}[]
    \centering
    \begin{subfigure}[b]{0.33\textwidth} 
        \centering
        \includegraphics[width=\textwidth]{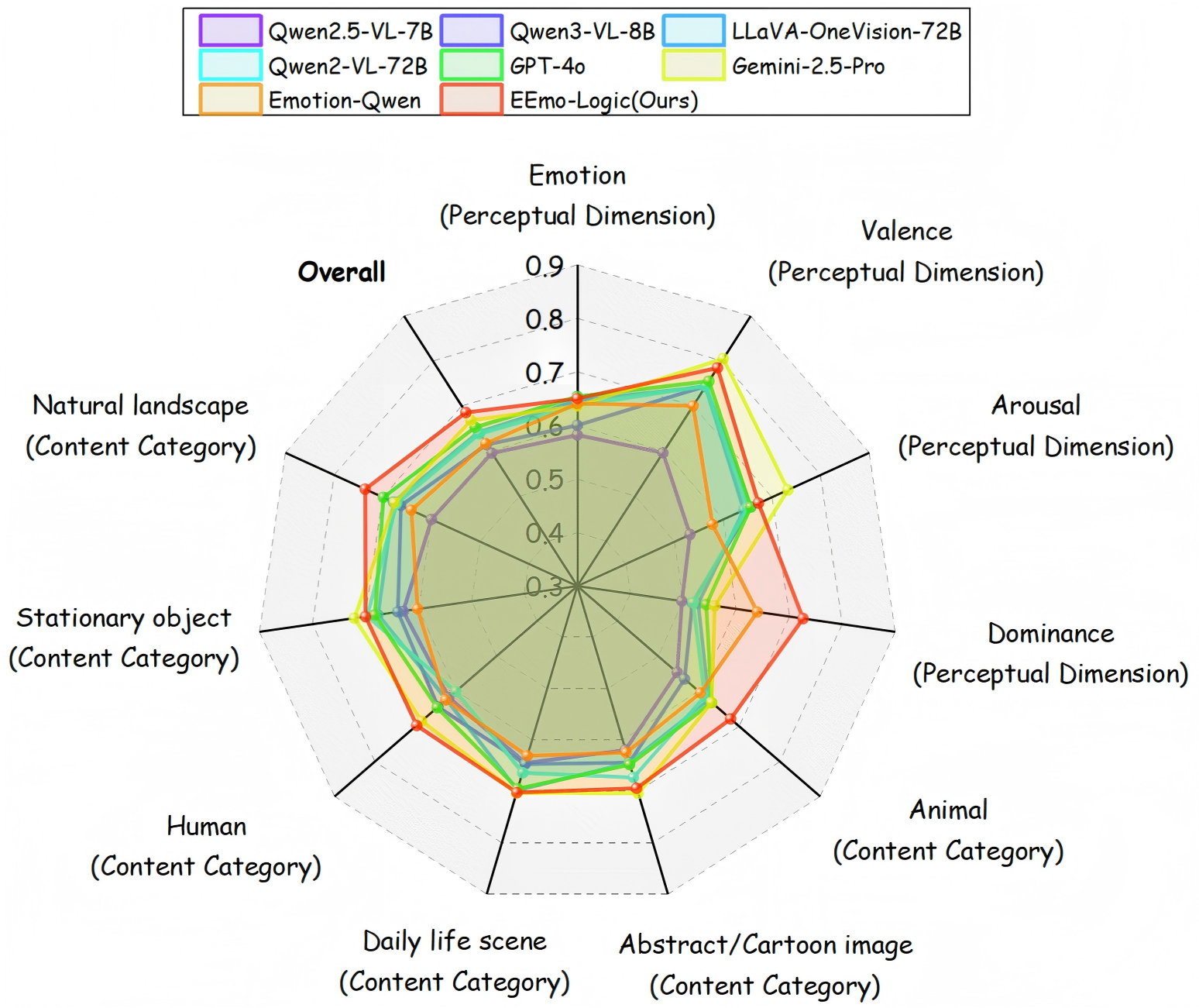} 
        \caption{Perception}
        \label{supp fig: perception}
    \end{subfigure}
    \hfill 
    \begin{subfigure}[b]{0.34\textwidth}
        \centering
        \includegraphics[width=\textwidth]{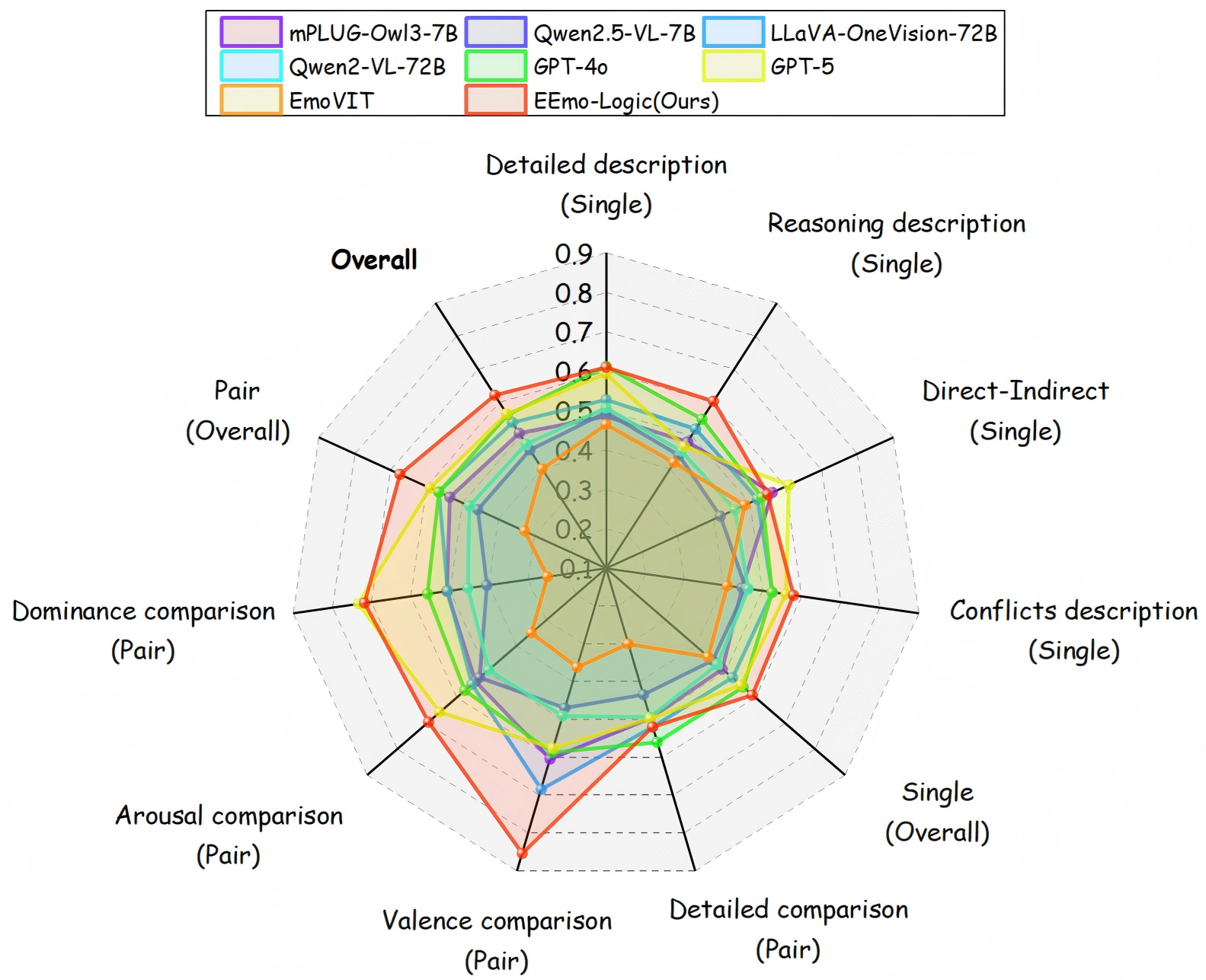}
        \caption{Description}
        \label{supp fig: description}
    \end{subfigure}
    \hfill 
    \begin{subfigure}[b]{0.29\textwidth}
        \centering
        \includegraphics[width=\textwidth]{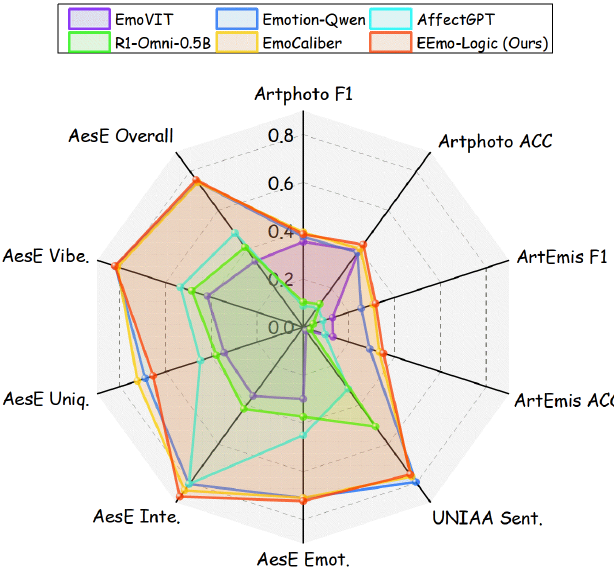}
        \caption{Cross-Domain Dataset}
        \label{supp fig: ood}
    \end{subfigure}
    
    \caption{Comparison of EEmo-Logic with other MLLMs on the perception and description tasks in EEmo-Bench, as well as on cross-domain datasets.}
    \label{fig:three_graphs}
\end{figure}

\begin{table*}[]
\centering
\footnotesize
\renewcommand{\arraystretch}{1.2}   
\setlength{\tabcolsep}{5pt}        
\caption{Validation of GRPO's emotional reasoning capabilities from textual and semantic perspectives.}
\begin{tabular}{l|cccccc}
\hline
\textbf{Verification Method} & \textbf{Ranking} & \textbf{Valence} & \textbf{Arousal} & \textbf{Dominance} & \textbf{Artphoto} & \textbf{Artemis} \\ \hline
Textual Matching       & 99.28\% & 84.95\% & 93.93\% & 84.37\%   & 98.76\%  & 95.21\% \\
Semantic Matching   & 71.11\% & 85.46\% & 96.63\% & 83.16\%   & 97.52\%  & 94.90\% \\ \hline
\end{tabular}
\label{tab: reasoning matching}
\end{table*}

\subsection{Detail Analysis on Perception Task}

As shown in Tab \ref{tab: Supp perception} and Fig.~\ref{fig:three_graphs}(a), we adopt the task decomposition defined in EEmo-Bench \cite{EEmobench} based on perceptual dimensions and content categories. This setup allows us to comprehensively compare EEmo-Logic with other models using an established standard.

\textbf{Performance across Perceptual Dimensions}.
The results indicate that EEmo-Logic performs comparably to current state-of-the-art commercial models across all perceptual dimensions. Notably, in \textit{dominance} analysis, EEmo-Logic significantly outperforms all competitors, surpassing the second-best model by $8.62\%$. This achievement highlights its superior alignment with human perception in capturing deep, specialized emotional nuances, reflecting a robust empathic capability.
Furthermore, a comparison with the original EEmo-Bench study reveals a growing emphasis on emotion understanding in recent research. For instance, Gemini-2.5-Pro \cite{gemini2.5} leads in \textit{valence} and \textit{arousal} tasks, while Qwen3-VL-8B \cite{qwen3} sets the standard among medium-scale open-source models. These rapid developments underscore the critical role of comprehensive visual evoked-emotion understanding in advancing human-like intelligence and validate the significance of emotion-centric models like EEmo-Logic.

\textbf{Performance across Content categories}.
Following two-stage training with complementary emphases on emotion understanding, EEmo-Logic achieves improvements exceeding $5\%$ across all content categories relative to the base model (Qwen2.5-VL-7B \cite{Qwen2.5}). Notably, it secures substantial gains in \textit{animal} and \textit{natural landscape} scenes, with increases of $13.27\%$ and $13.67\%$ respectively, and demonstrates a clear performance advantage over other proprietary MLLMs. The proficiency in the \textit{animal} category indicates that EEmo-Logic not only understands emotions in human-centric images but also captures animal emotions in a human-analogous manner, leveraging commonsense knowledge about human–animal affective associations to better align with human perception. Similarly, the performance in \textit{natural landscapes} reveals an enhanced sensitivity to global color styles and lighting compositions, highlighting promising potential for aesthetic analysis.
In other categories, EEmo-Logic parallels the performance of Gemini-2.5-Pro \cite{gemini2.5} without a pronounced lead. This parity likely stems from unavoidable heterogeneity in emotion definitions and cultural biases within source datasets, particularly for \textit{abstract/cartoon images} and \textit{stationary objects}, where subjective variance is high. Unlike proprietary MLLMs pretrained on massive corpora, learning these fine-grained cultural nuances remains a challenge with small-scale data. We remain committed to advancing this line of research.

\subsection{Detail Analysis on Description Task}

For the description task on EEmo-Bench \cite{EEmobench}, we compare EEmo-Logic with other models across its specifically defined question types to infer fine-grained differences in emotion understanding and descriptive ability. As illustrated in Tab.~\ref{tab: Supp description} and Fig.~\ref{fig:three_graphs}(b), we show the results for single-image and paired-image scenarios and provide observations as follows.

\textbf{Evaluation in Single-Image Scenarios}.
The results indicate that EEmo-Logic achieves state-of-the-art performance on both \textit{emotion causal reasoning} and \textit{conflicting emotion description} tasks. Success in the former validates the effectiveness of our automated emotion reasoning generation in EEmoDB, which is grounded in expert comments. By rooting emotion analyses in expert interpretations, we substantially reduce annotation noise caused by model hallucinations while leveraging the summarization and CoT capabilities of proprietary LLMs to produce large-scale, high-quality synthetic data.
The latter demonstrates that EEmo-Logic genuinely captures emotion intensity differences and inter-individual diversity. For instance, it recognizes that a sunset scene may evoke melancholic sadness for some viewers while eliciting relaxed joy for others. This reflects a comprehensive and nuanced understanding of emotion. 
However, performance in emotion source attribution (\textit{direct-indirect}) is partially limited. The absence of human annotations necessitates reliance on \textit{Deepseek-generated} judgments, which inevitably introduces model-specific bias and hallucinations. We plan to explore more reliable model-assisted data generation strategies to address this limitation in future work.

\textbf{Evaluation in Paired-Image Scenarios}.
EEmo-Logic exhibits robust performance in inter-image VAD comparison tasks. This success stems from our two-stage coarse-to-fine training strategy, which ensures high accuracy in DES understanding and fosters CoT processes that closely parallel human reasoning. Consequently, the descriptive responses generated by the model possess high information density. Although EEmo-Logic falls slightly behind proprietary MLLMs like Claude-3.7-Sonnet \cite{claude} in discerning emotional similarities and differences, it remains the most competitive candidate among open-source models.

\begin{figure}[]
    \centering
    
    
    \begin{subfigure}[b]{0.19\textwidth}
        \centering
        \includegraphics[width=\linewidth]{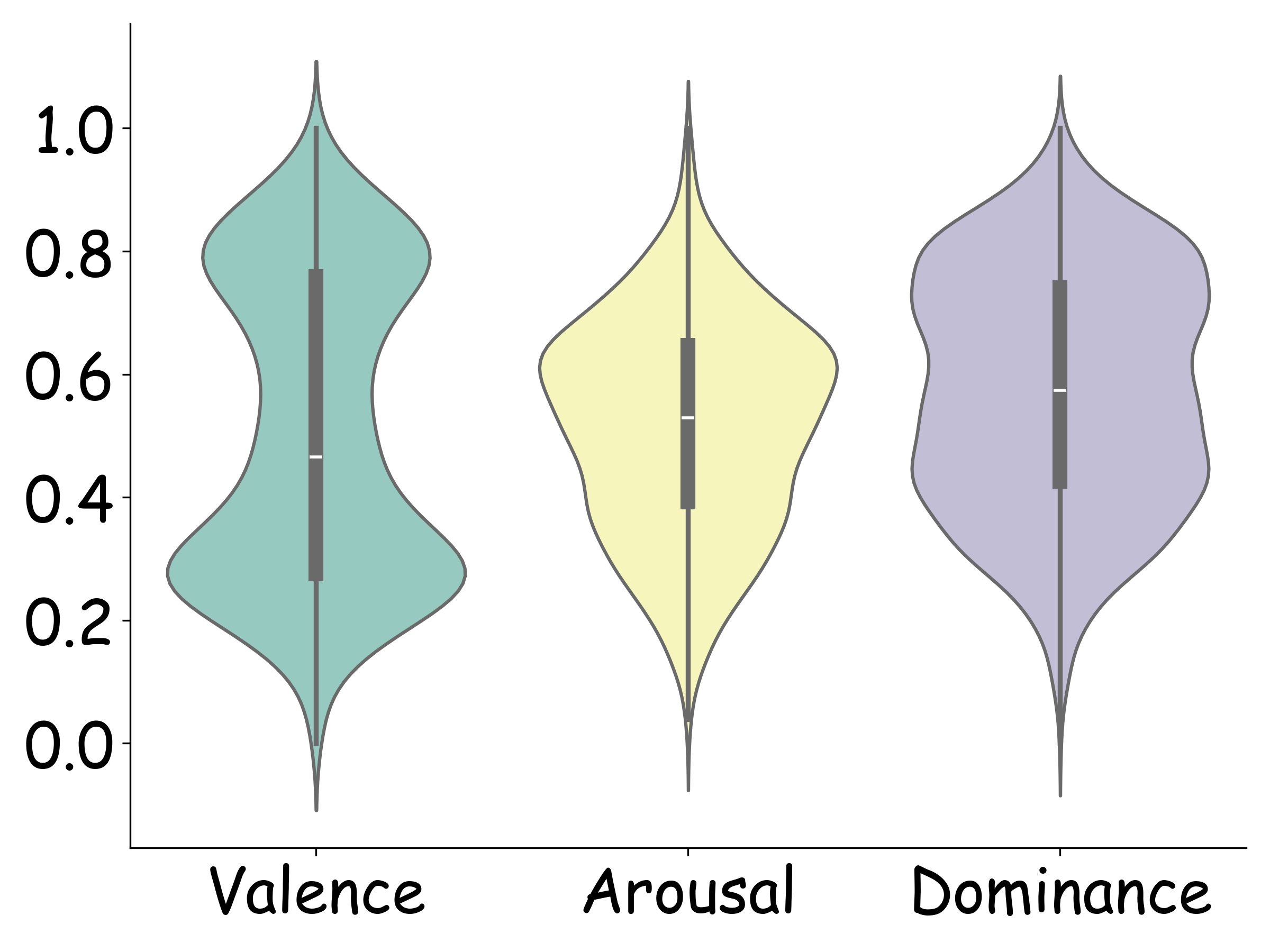}
        \caption{Ground-truth}
    \end{subfigure}
    \hfill
    \begin{subfigure}[b]{0.19\textwidth}
        \centering
        \includegraphics[width=\linewidth]{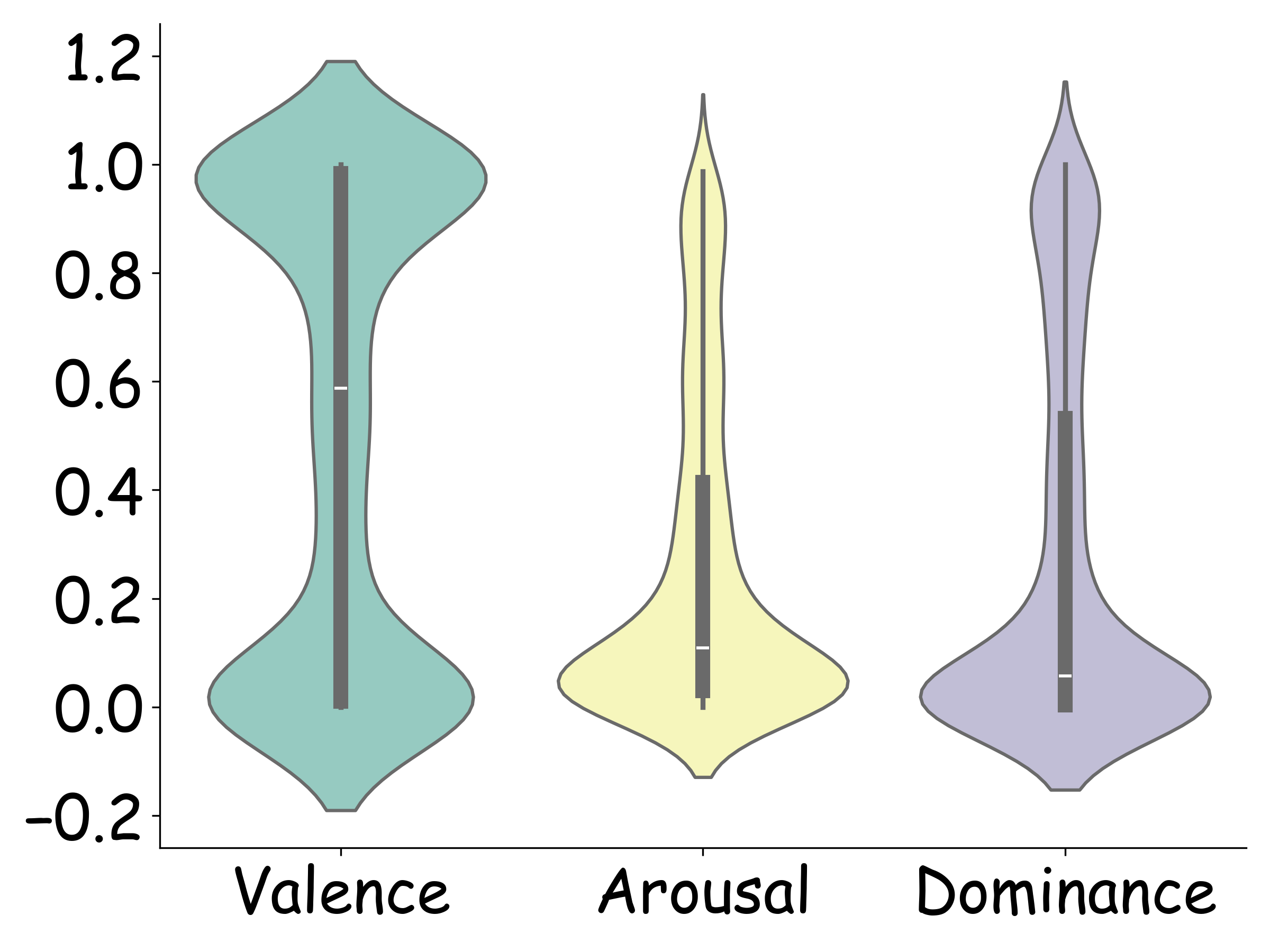}
        \caption{Deepseek-VL-7B}
    \end{subfigure}
    \hfill
    \begin{subfigure}[b]{0.19\textwidth}
        \centering
        \includegraphics[width=\linewidth]{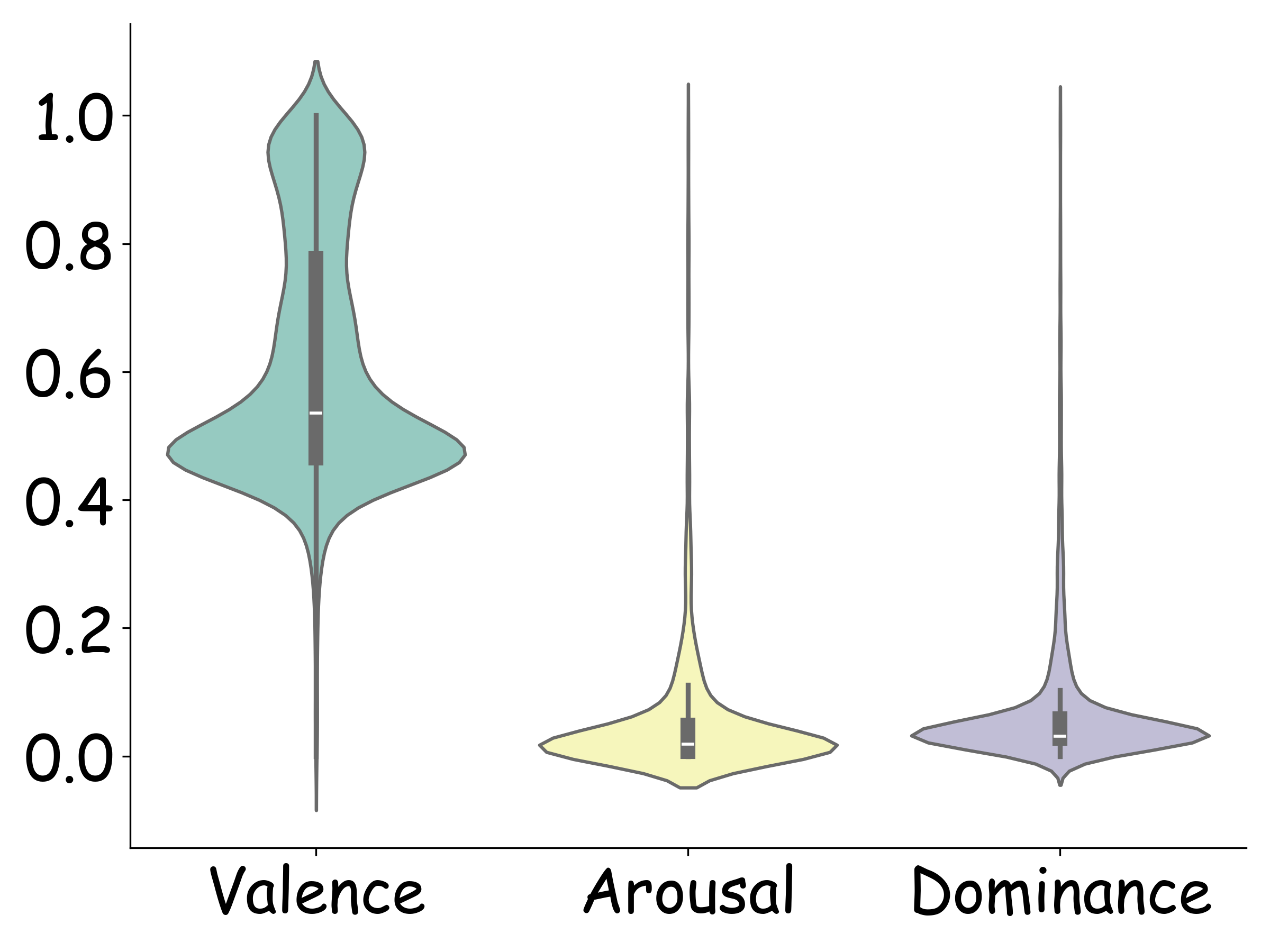}
        \caption{Janus-7B-Pro}
    \end{subfigure}
    \hfill
    \begin{subfigure}[b]{0.19\textwidth}
        \centering
        \includegraphics[width=\linewidth]{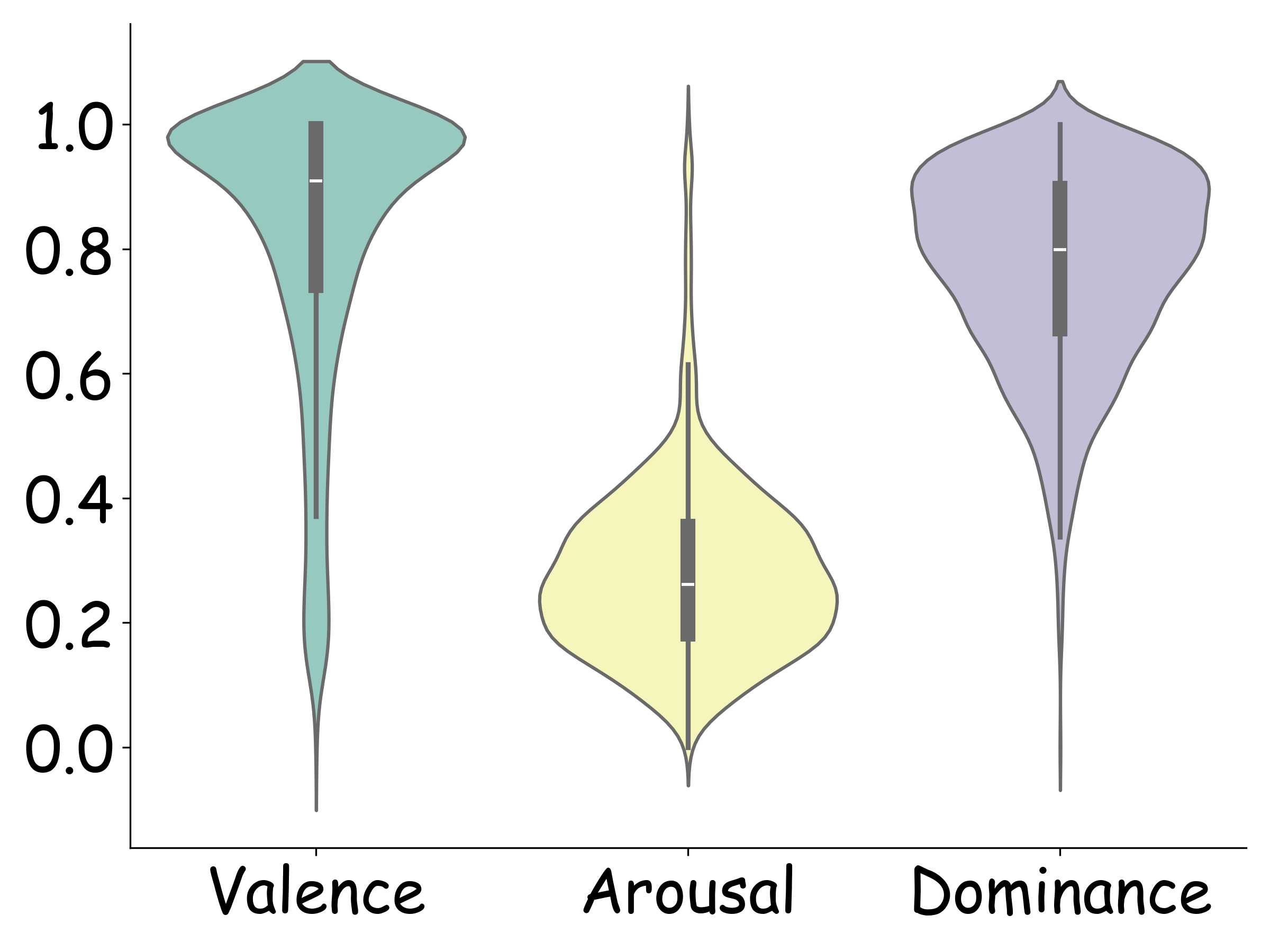}
        \caption{LLaVA-OneVision-7B}
    \end{subfigure}
    \hfill
    \begin{subfigure}[b]{0.19\textwidth}
        \centering
        \includegraphics[width=\linewidth]{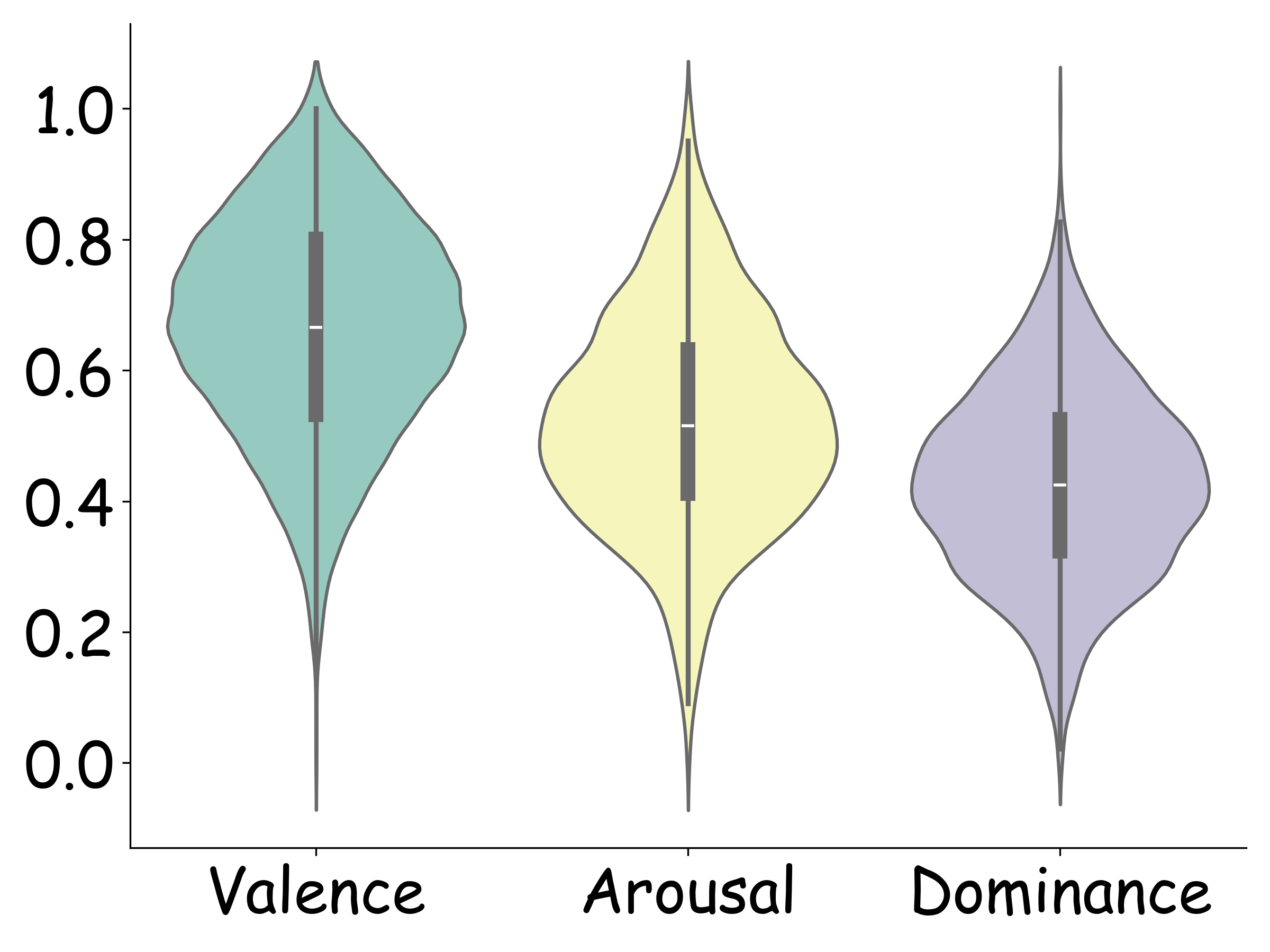}
        \caption{LLaVA-NEXT-8B}
    \end{subfigure}
    
    \par\bigskip 
    
    
    \begin{subfigure}[b]{0.19\textwidth}
        \centering
        \includegraphics[width=\linewidth]{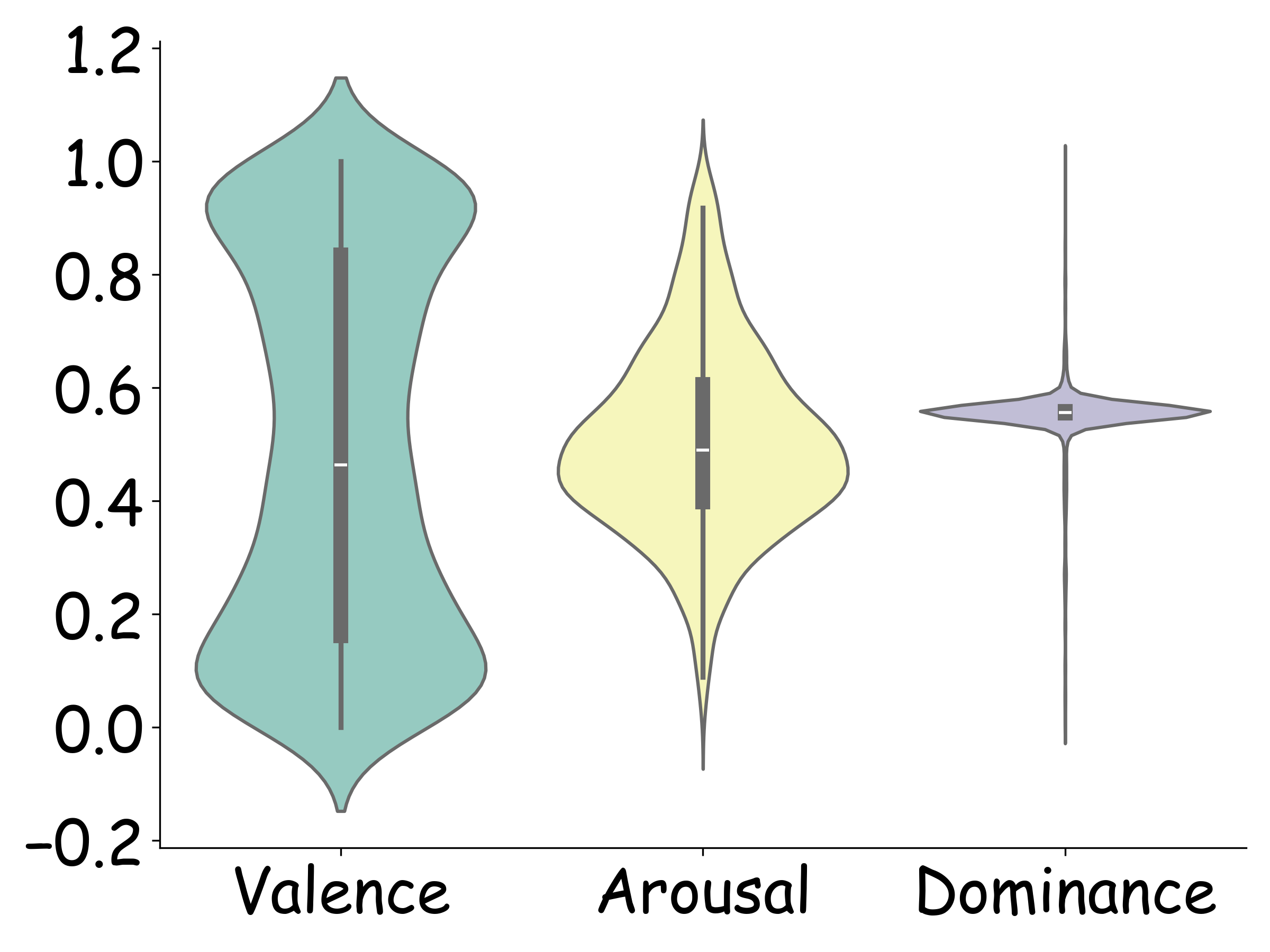}
        \caption{mPLUG-Owl3-7B}
    \end{subfigure}
    \hfill
    \begin{subfigure}[b]{0.19\textwidth}
        \centering
        \includegraphics[width=\linewidth]{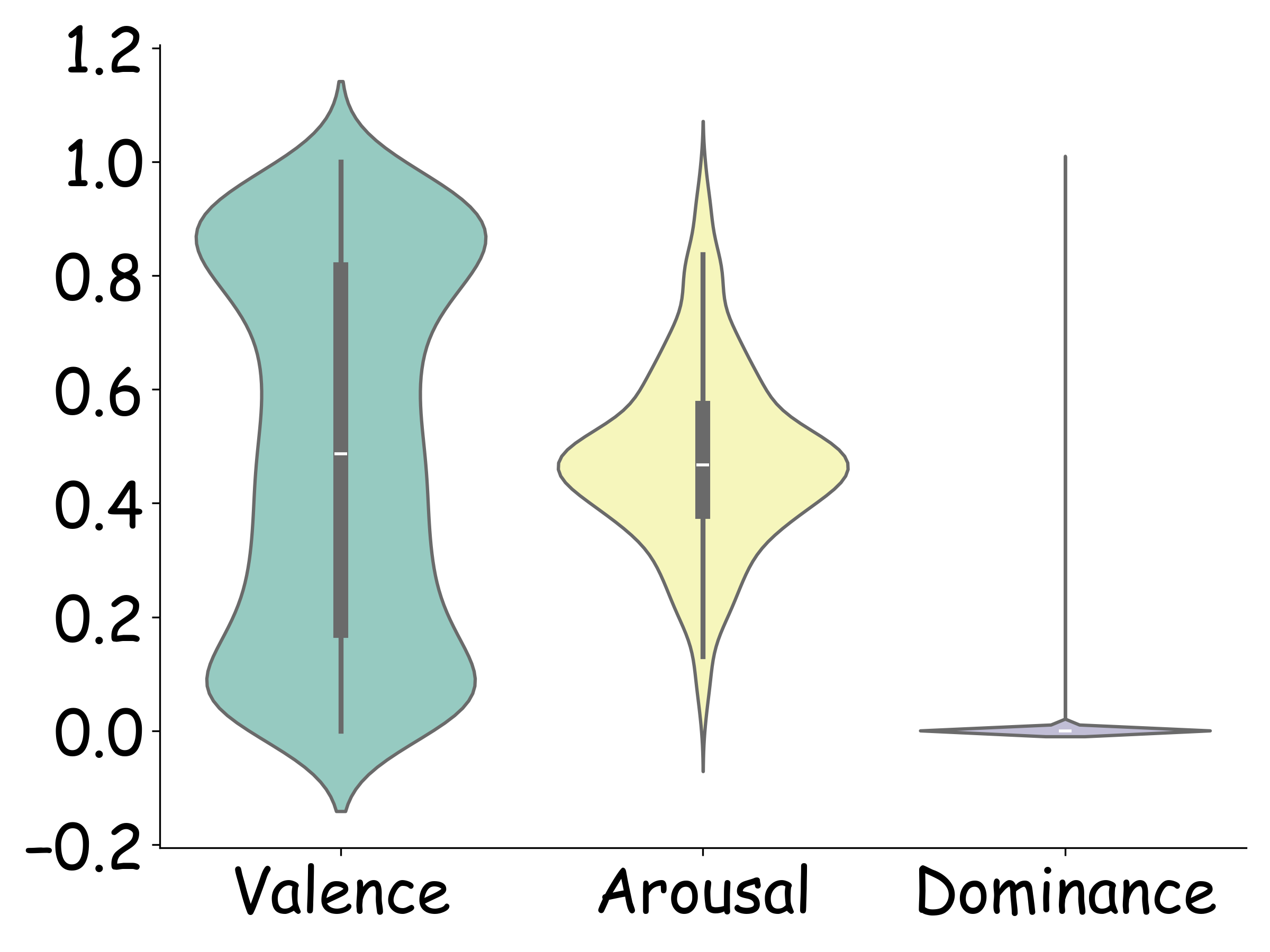}
        \caption{Qwen2-VL-7B}
    \end{subfigure}
    \hfill
    \begin{subfigure}[b]{0.19\textwidth}
        \centering
        \includegraphics[width=\linewidth]{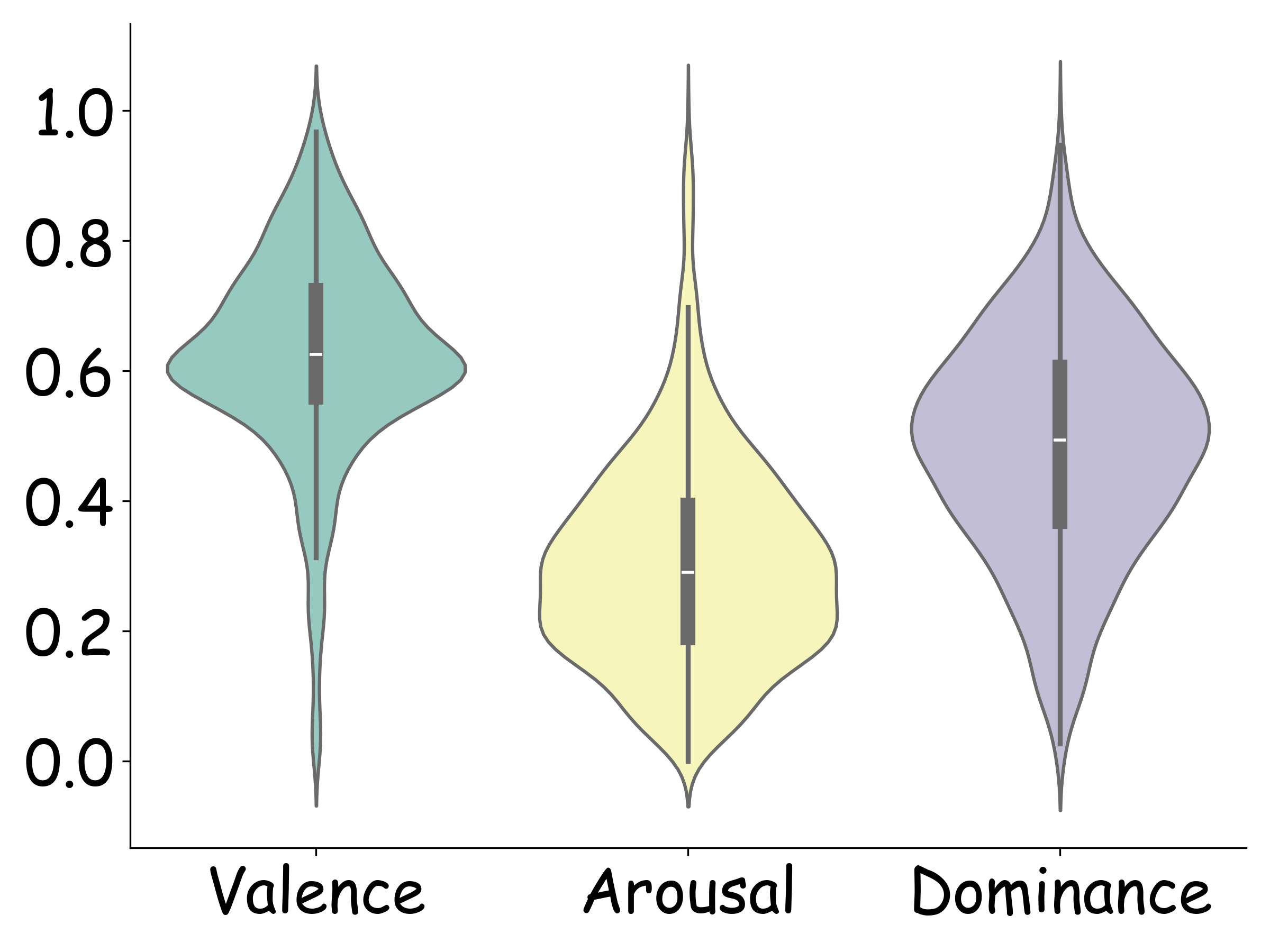}
        \caption{Qwen2.5-VL-7B}
    \end{subfigure}
    \hfill
    \begin{subfigure}[b]{0.19\textwidth}
        \centering
        \includegraphics[width=\linewidth]{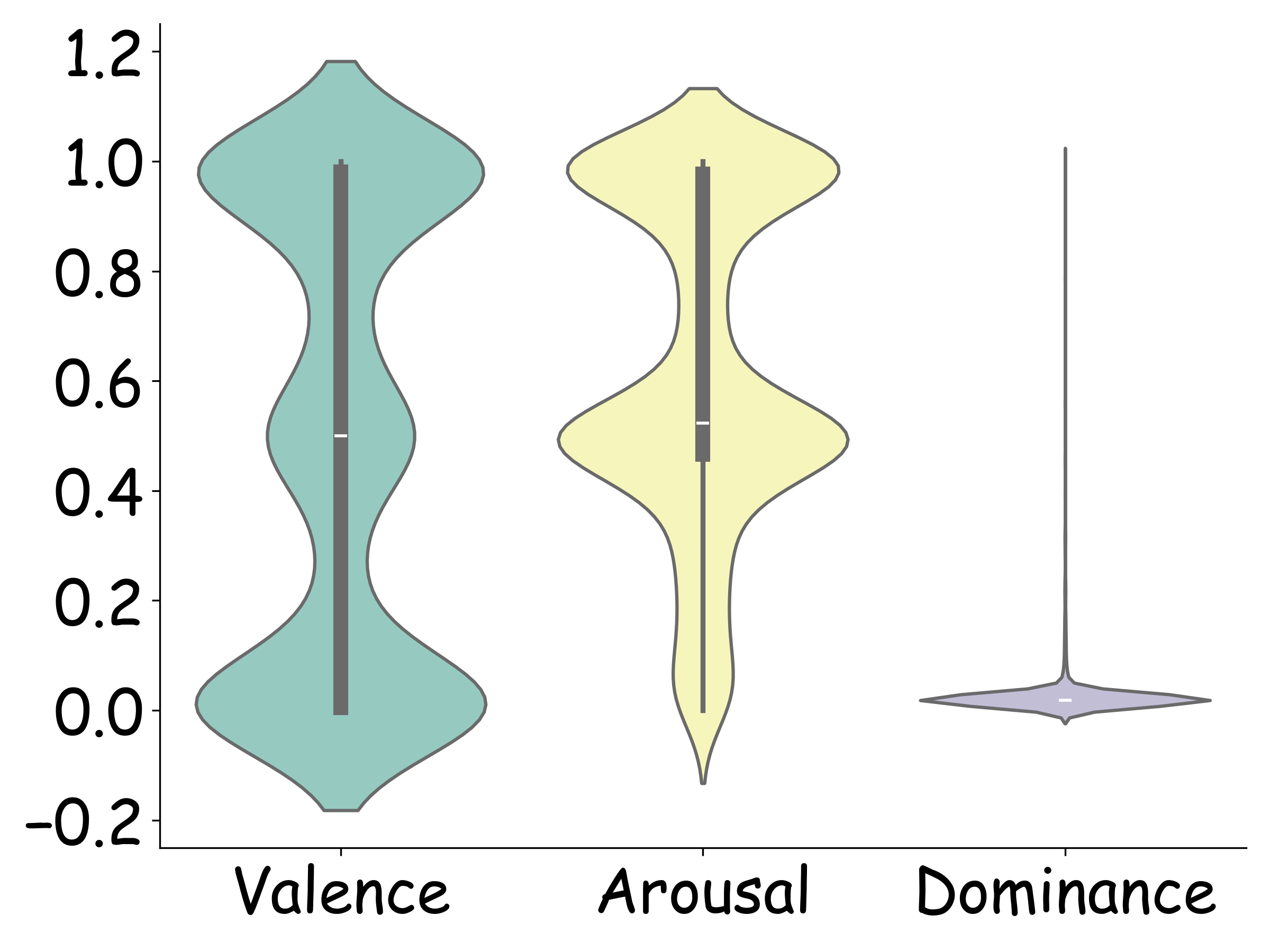}
        \caption{Qwen3-VL-8B}
    \end{subfigure}
    \hfill
    \begin{subfigure}[b]{0.19\textwidth}
        \centering
        \includegraphics[width=\linewidth]{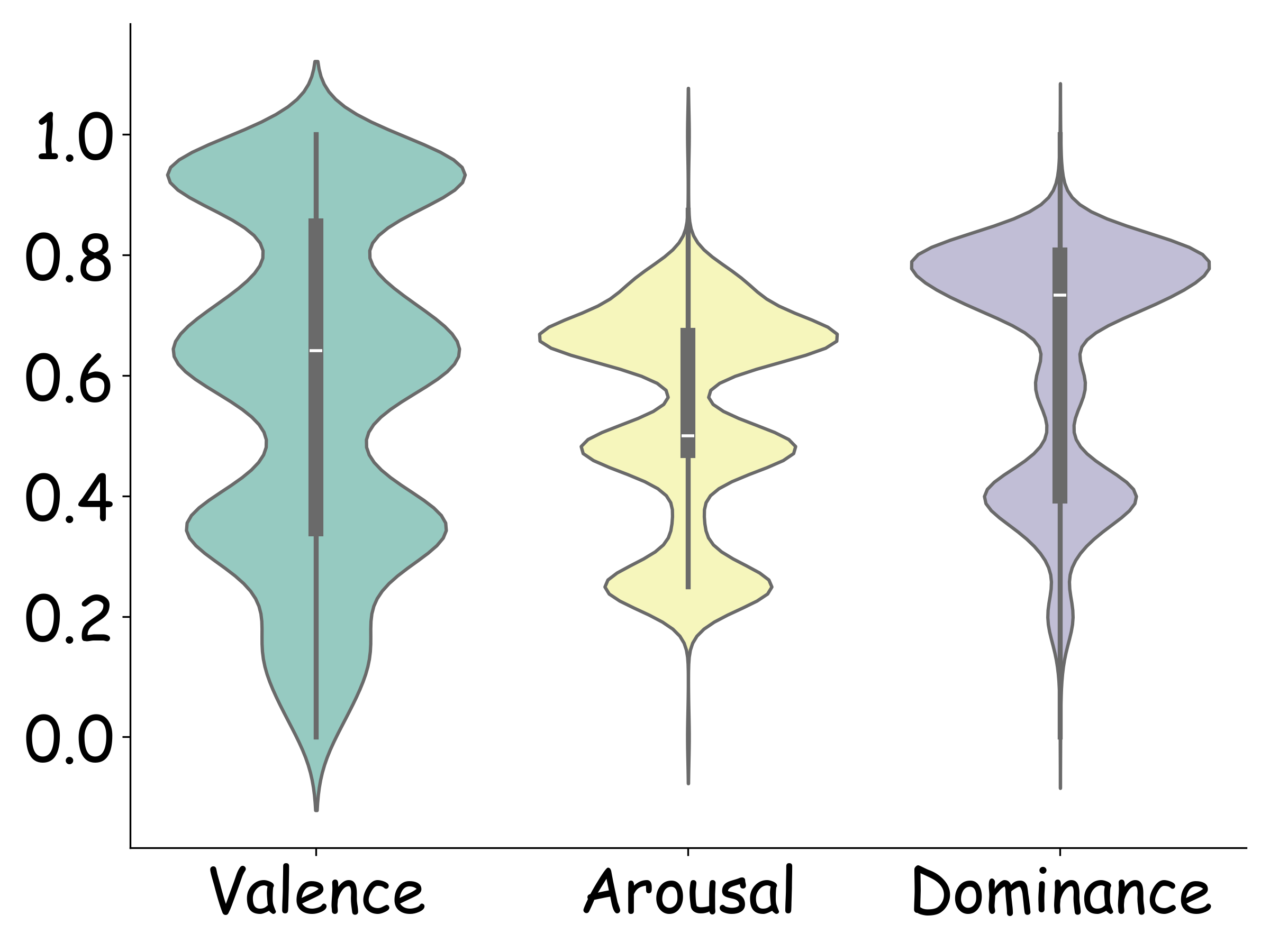}
        \caption{EEmo-Logic}
    \end{subfigure}
    
    \caption{Comparative visualization of VAD score distributions for EEmo-Logic and baseline models against ground-truth annotations.}
    \label{fig: vad distribution}
\end{figure}

\begin{table*}[]
\centering
\scriptsize
\renewcommand{\arraystretch}{1.2}   
\setlength{\tabcolsep}{5pt}        
\caption{Performance comparison after replacing original inference prompts with semantic equivalents.}
\begin{tabular}{l|c|ccc|cc|cc}
\hline
\multicolumn{1}{c|}{\textbf{Method}}     & \textbf{Ranking}       & \textbf{Valence}   & \textbf{Arousal}   & \textbf{Dominance} & \multicolumn{2}{c|}{\textbf{Artphoto}} & \multicolumn{2}{c}{\textbf{ArtEmis}} \\ \hline
\multicolumn{1}{c|}{Metrics}    & \textit{Emotion Score↑} & \textit{SRCC↑/PLCC↑} & \textit{SRCC↑/PLCC↑} & \textit{SRCC↑/PLCC↑} & \textit{F1↑}            & \textit{ACC↑}           & \textit{F1↑}           & \textit{ACC↑}          \\ \hline
EEmo-Logic (w/ original prompt) & 67.97\%       & 0.84/0.83 & 0.57/0.56 & 0.78/0.74 & 38.68\%       & 42.43\%       & 31.58\%      & 35.10\%      \\
EEmo-Logic (w/ modified prompt) & 67.99\%       & 0.84/0.83 & 0.56/0.56 & 0.78/0.75 & 40.54\%       & 43.92\%       & 27.40\%      & 32.03\%      \\ \hline
\end{tabular}
\label{tab: modify prompt}
\end{table*}

\subsection{Detail Analysis of Score Distribution on Assessment Task} \label{supp: distribution}

Fig.~\ref{fig: vad distribution} compares the VAD score distributions of the open-source models listed in Tab. \ref{tab: assessment}. Unlike EEmo-Logic, which directly predicts numerical scores, the baseline models utilize the probability-based conversion described in Appendix \ref{supp: softmax}. Consequently, these methods largely fail to satisfy distributional balance requirements, particularly for \textbf{dominance} where excessive concentration leads to degraded performance. Although LLaVA-NEXT-8B \cite{llava-next} and Qwen2.5-VL-7B \cite{Qwen2.5} produce relatively balanced distributions as shown in Figs.~\ref{fig: vad distribution}(e) and (f), their SRCC and PLCC metrics remain suboptimal for high-precision assessment. Moreover, their spindle-like distributions deviate significantly from the theoretical profile in Fig.~\ref{fig: vad distribution}(a). In contrast, EEmo-Logic (Fig.~\ref{fig: vad distribution}(j)) achieves superior overall performance. Its distributions closely align with the theoretical form and demonstrate dense coverage across multiple levels, reflecting a fine-grained understanding of VAD scales. Additionally, by generating explicit reasoning for its predictions, EEmo-Logic enhances reliability and versatility in application.

\begin{table*}[]
\centering
\footnotesize
\renewcommand{\arraystretch}{1.2}   
\setlength{\tabcolsep}{7pt}        
\caption{Performance comparison on the in-domain tasks of EEmo-Bench between EEmo-Logic and the strongest emotion-oriented baseline (Emotion-Qwen) fine-tuned on EEmoDB.}
\begin{tabular}{c|ccc|ccc|c|l}
\hline
\multirow{2}{*}{\textbf{In-Domain}}                        & \multicolumn{3}{c|}{\textbf{Perception}} & \multicolumn{3}{c|}{\textbf{Description}} & \multirow{2}{*}{\textbf{Ranking}} & \multirow{2}{*}{\textbf{Overall}} \\ \cline{2-7}
                                                  & Single    & Pair     & Overall  & Single    & Pair      & Overall  &                          &                          \\ \hline
\multicolumn{1}{l|}{Emotion-Qwen (SFT on EEmoDB)} & 65.90\%   & 62.90\%  & 64.88\%  & 53.39\%   & 55.07\%   & 54.02\%  & 60.49\%                  & 59.80\%                  \\
EEmo-Logic (Ours)                                       & 70.05\%   & 65.61\%  & 68.54\%  & 59.02\%   & 67.40\%   & 62.16\%  & 67.97\%                  & 66.22\%                  \\ \hline
\end{tabular}
\label{tab: sft on Emotionqwen}
\end{table*}

\begin{table*}[]
\centering
\footnotesize
\renewcommand{\arraystretch}{1.2}   
\setlength{\tabcolsep}{5pt}        
\caption{Ablation study on reward signal weights.}
\begin{tabular}{c|c|ccc|cc|cc}
\hline
$\lambda_0$ & \textbf{Ranking}                            & \textbf{Valence}                       & \textbf{Arousal}                       & \textbf{Dominance}                      & \multicolumn{2}{c|}{\textbf{Artphoto}} & \multicolumn{2}{c}{\textbf{ArtEmis}} \\ \hline
Metrics     & \multicolumn{1}{l|}{\textit{Emotion Score↑}} & \multicolumn{1}{l}{\textit{SRCC↑/PLCC↑}} & \multicolumn{1}{l}{\textit{SRCC↑/PLCC↑}} & \multicolumn{1}{l|}{\textit{SRCC↑/PLCC↑}} & \textit{F1↑}            & \textit{ACC↑}           & \textit{F1↑}           & \textit{ACC↑}          \\ \hline
1           & 56.88\%                            & 0.84/0.83                     & 0.53/0.52                     & 0.8/0.78                       & 39.55\%       & 42.06\%       & 26.47\%      & 32.29\%      \\
0.2 (Ours)  & 67.97\%                            & 0.84/0.83                     & 0.57/0.56                     & 0.78/0.74                      & 38.68\%       & 42.43\%       & 31.58\%      & 35.10\%      \\ \hline
\end{tabular}
\label{tab: lambda}
\end{table*}

\subsection{Detailed Analysis of the Improvement in Emotional Reasoning Ability} 

First, the substantial improvement in emotion reasoning stems from our model's comprehensive perception and reasoning capability, such as VAD levels and emotion intensity. While previous methods have explored only partial features, our model constructs a holistic understanding, which directly drives deeper and broader reasoning.

Second, this capability can be reflected from two perspectives: 1) performance on the description task, as shown in Tab. \ref{tab: 3 tasks}, and 2) the quality of the thinking process in the GRPO task. To rigorously evaluate the latter, we further conduct additional experiments. We assess the consistency between the model's thinking and final answers using both exact text matching and Qwen3-Embedding semantic matching. The results are reported in Tab.~\ref{tab: reasoning matching}.

The results indicate strong alignment between the model’s reasoning traces and its final answers. Notably, although we have not imposed any explicit constraints on the thinking process during GRPO, this consistency emerges naturally, which directly corroborates the findings of DeepSeek-R1. Since strong baselines also maintain high thinking-answer consistency, the fact that EEmo-Logic has additionally achieved state-of-the-art accuracy in fine-grained emotional attribute prediction confirms its superior reasoning capability over existing methods.

Third, to verify that the strong performance of our model does not come from prompt engineering, we have replaced the original inference prompts with semantically equivalent paraphrases and re-evaluated the model. The results, as illustrated in Tab.~\ref{tab: modify prompt}, show no significant performance degradation; in fact, accuracy on some tasks even improves slightly. These findings demonstrate that EEmo-Logic is robust to prompt variations and further confirm that its performance gains mainly come from our training strategy, rather than from carefully structured prompt design.

\subsection{Effectiveness of the Proposed Framework.} \label{supp: others training on the in-domain data}

In this paper, Tab.~\ref{tab: 3 tasks} compares models optimized on their respective corpora to demonstrate improvements over baselines, while Tab.~\ref{tab: OOD} evaluates generalization against other models. However, this setup may partially obscure the source of the performance gains. Because the baseline models are not trained on the proposed EEmoDB dataset for in-domain tasks, the superior performance of EEmo-Logic might not originate exclusively from the model architecture itself. Therefore, to isolate architectural effects from data effects, we conduct a strictly controlled in-domain experiment. Specifically, we fine-tune the strongest emotion-oriented baseline from Tab.~\ref{tab: 3 tasks}, i.e., Emotion-Qwen, on our dataset. As shown in Tab.~\ref{tab: sft on Emotionqwen}, although training on our dataset enhances the performance of Emotion-Qwen, it still underperforms EEmo-Logic. This confirms that our performance gains stem fundamentally from the proposed multi-stage framework, rather than merely the training corpus.

\subsection{More Ablation Experiments} \label{supp: more ablation experiments}

We provide a more systematic ablation study to justify the selection of the reward models and hyperparameters for different tasks on EEmoDB-Assess presented in the main text, as follows:

\textbf{Weights of the Reward Signals.} In Eq.~\ref{equ:total}, each training sample is optimized for only one task, so the trade-off at each step is only between the task-accuracy reward and the format reward for the current task. Since the accuracy reward is assigned only when the correct task prefix is generated, the format reward mainly acts as a gating mechanism rather than a dominant objective, suggesting that a relatively small $\lambda_0$ is more suitable. Consistent with this analysis, our ablation results shown in Tab.~\ref{tab: lambda} reveal that $\lambda_0 = 0.2$ performs better than the default choice of $\lambda_0 = 1$, especially on emotion ranking and arousal prediction.

\textbf{Reward Design for Emotion Ranking and VAD Prediction.} For emotion ranking, we test the contribution of the squared margin function. For VAD prediction, we compare a standard binary reward (threshold is set to $0.1$) with our Hybrid Regression Reward. Our results, presented in Tab.~\ref{tab: reward of emotion ranking}, show that removing the margin function makes the reward landscape less steep and negatively affects other tasks, especially arousal prediction. Moreover, the standard binary reward is less effective for both ranking and arousal tasks. 

\begin{table*}[]
\centering
\footnotesize
\renewcommand{\arraystretch}{1.2}   
\setlength{\tabcolsep}{7pt}        
\caption{Comparative experiments on reward designs for emotion ranking and VAD prediction.}
\begin{tabular}{llllll}
\hline
\textbf{Ranking Reward} & \textbf{VAD Reward}      & \textbf{Ranking} & \textbf{Valence} & \textbf{Arousal} & \textbf{Dominance} \\ \hline
w/o Squared Margin      & Hybrid Regression Reward & 67.45\%          & 0.84/0.83        & 0.47/0.47        & 0.81/0.78          \\
w/ Squared Margin       & Binary Reward            & 65.85\%          & 0.85/0.83        & 0.49/0.49        & 0.79/0.77          \\
w/ Squared Margin       & Hybrid Regression Reward & 67.97\%          & 0.84/0.83        & 0.57/0.56        & 0.78/0.74          \\ \hline
\end{tabular}
\label{tab: reward of emotion ranking}
\end{table*}

\begin{table*}[]
\centering
\footnotesize
\renewcommand{\arraystretch}{1.2}   
\setlength{\tabcolsep}{7pt}        
\caption{Comparative experiments on reward designs for dominant emotion recognition.}
\begin{tabular}{cc|lll}
\hline
\multicolumn{1}{c}{\multirow{2}{*}{\textbf{Reward Template}}} & \multicolumn{1}{c|}{\multirow{2}{*}{\textbf{Parameter Design}}} & \multicolumn{1}{c}{\textbf{Artphoto}} & \multicolumn{1}{c}{\textbf{ArtEmis}} & \multicolumn{1}{c}{\textbf{Overall}} \\ \cline{3-5} 
\multicolumn{1}{c}{}                                          & \multicolumn{1}{c|}{}                                           & \multicolumn{1}{c}{\textit{F1↑/ACC↑}}          & \multicolumn{1}{c}{\textit{F1↑/ACC↑}}         & \multicolumn{1}{c}{\textit{F1↑/ACC↑}}           \\ \hline
Binary Accuracy Reward                                        & m=2,p=3                                                         & 39.61\%/40.94\%                       & 30.17\%/33.80\%                      & 34.89\%/37.37\%                      \\
Emotion Similarity Reward                                     & m=1,p=2                                                         & 40.07\%/41.69\%                       & 28.31\%/31.56\%                      & 34.19\%/36.63\%                      \\
Emotion Similarity Reward                                     & m=2,p=3                                                         & 38.68\%/42.43\%                       & 31.58\%/35.10\%                      & 35.13\%/38.77\%                      \\ \hline
\end{tabular}
\label{tab: DEC reward}
\end{table*}

\textbf{Reward Design for Dominant Emotion Recognition.} To evaluate the necessity of the emotion similarity reward and the choice of its parameters, we compare our method against a binary exact-match reward and modified versions of Eq.~\ref{equ:sim} using smaller $m$ and $p$, which assign higher rewards to semantically similar emotions. The results in Tab.~\ref{tab: DEC reward} consistently show that the design used in EEmo-Logic achieves the best performance, confirming the effectiveness of our proposed setting.

\subsection{More Qualitative Results} \label{supp: qualitative results}

\begin{figure}
    \centering
    \includegraphics[width=0.9\linewidth]{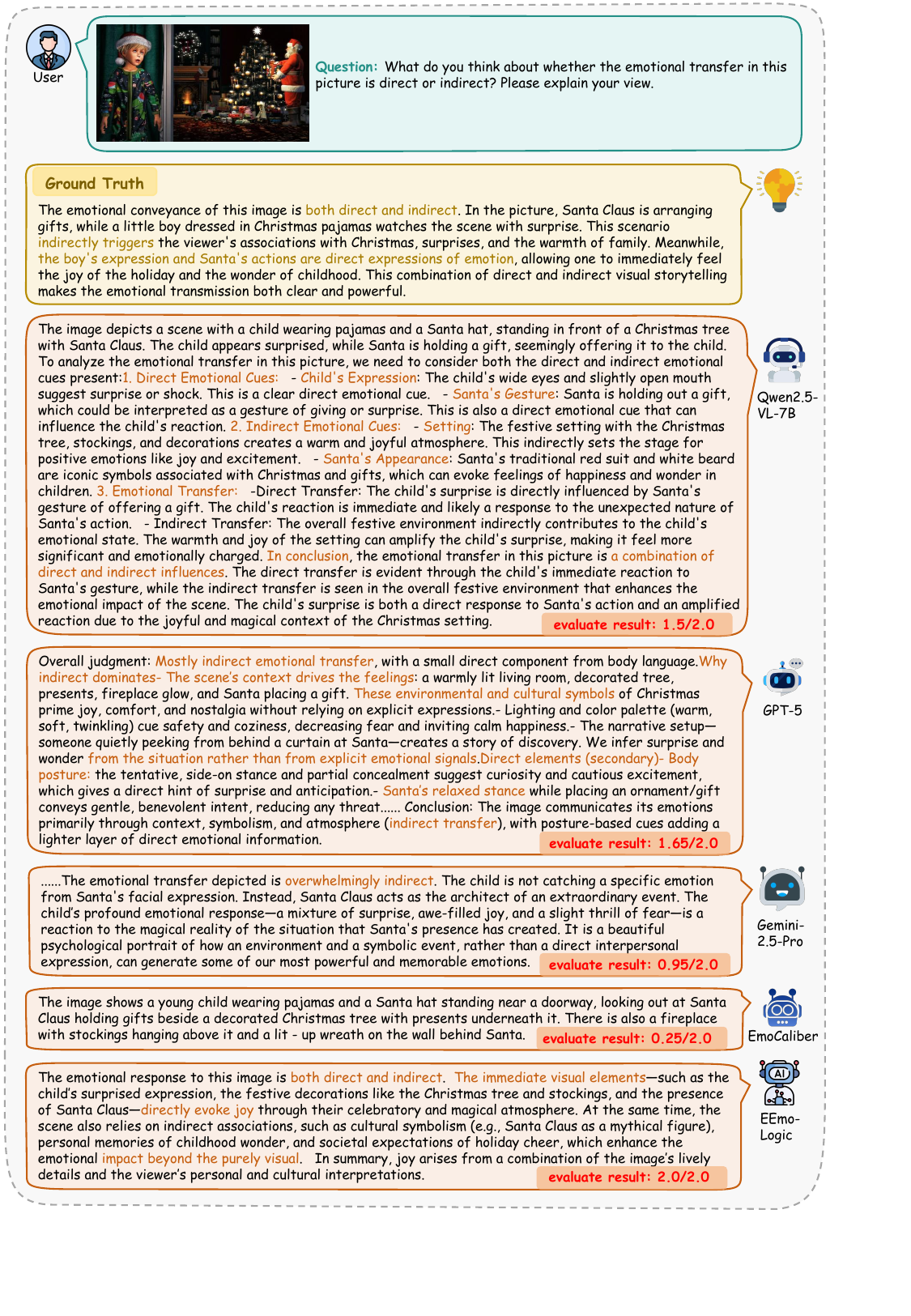}
    \vspace{-0.2cm}
    \caption{Qualitative comparison between EEmo-Logic and other MLLMs on the EEmo-Bench description tasks.}
    \label{fig:description}
\end{figure}

\begin{figure}
    \centering
    \includegraphics[width=0.8\linewidth]{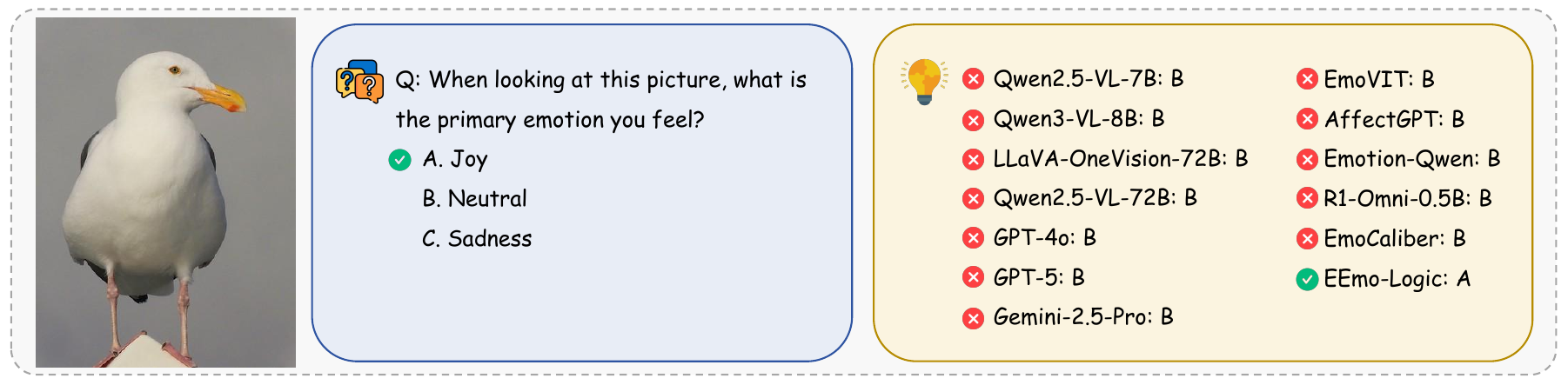}
    \vspace{-0.2cm}
    \caption{Qualitative comparison between EEmo-Logic and other MLLMs on the EEmo-Bench perception tasks.}
    \label{fig:perception}
\end{figure}

\begin{figure}
    \centering
    \includegraphics[width=0.8\linewidth]{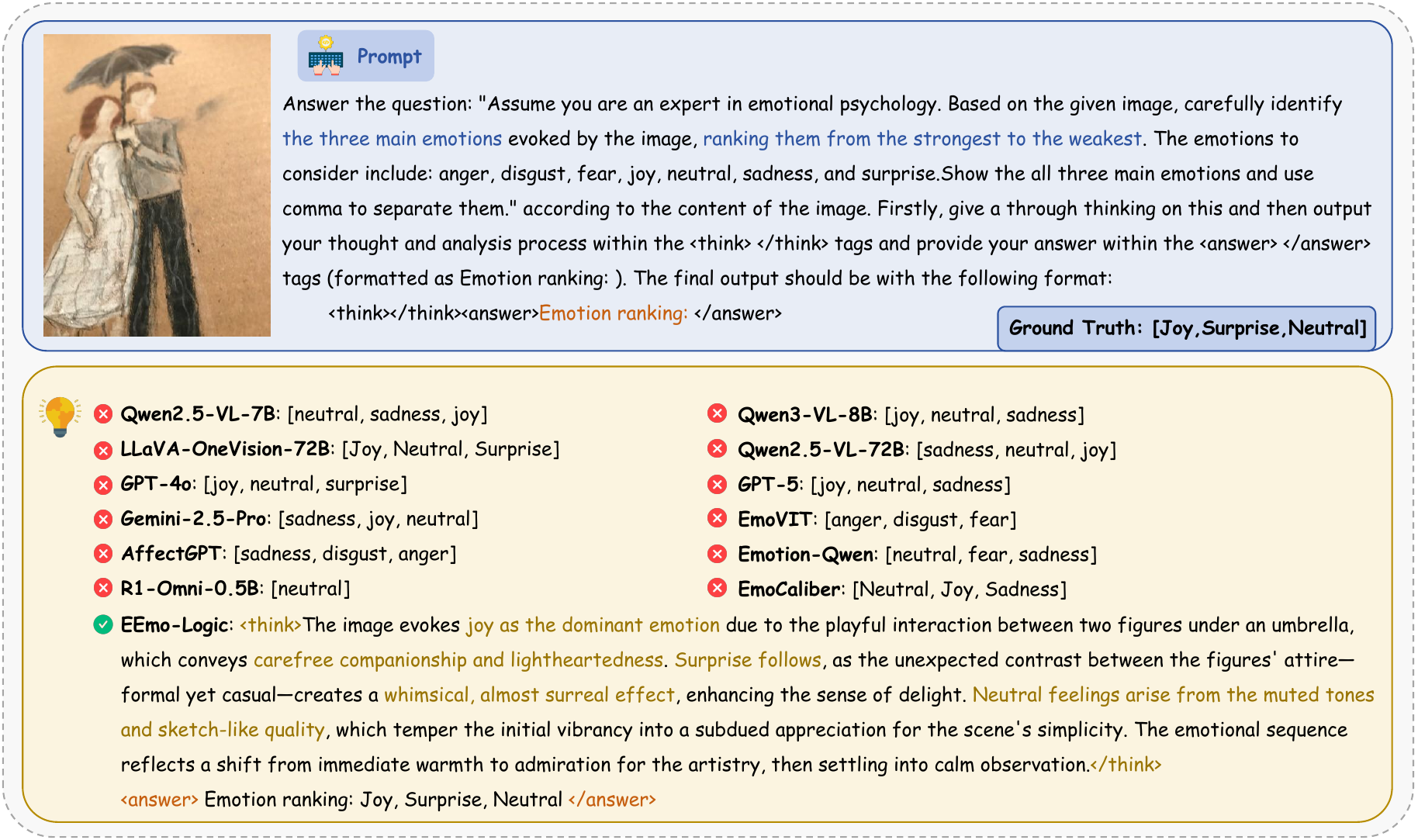}
    \vspace{-0.2cm}
    \caption{Qualitative comparison between EEmo-Logic and other MLLMs on the EEmo-Bench ranking tasks.}
    \label{fig:ranking}
\end{figure}

\begin{figure}
    \centering
    \includegraphics[width=0.8\linewidth]{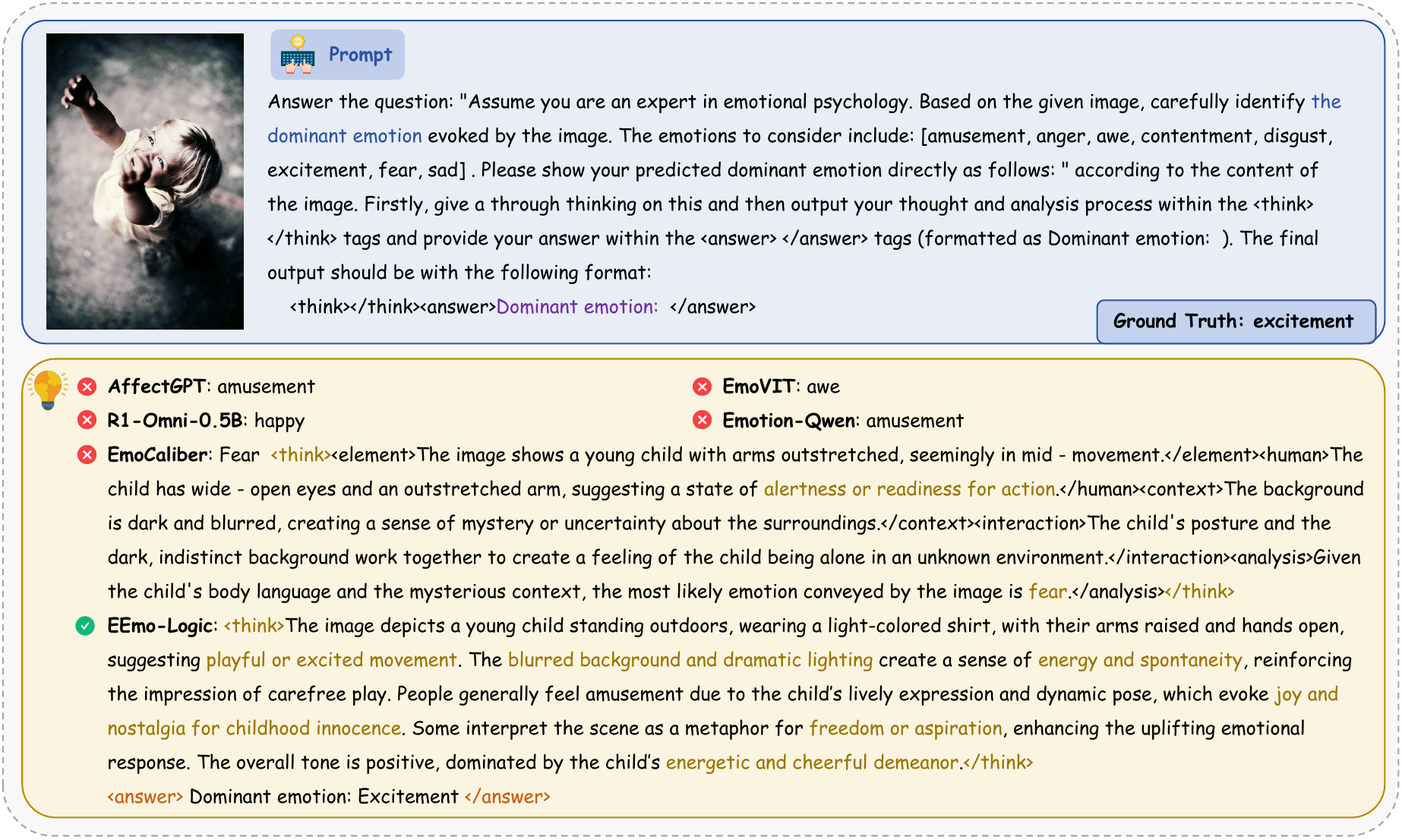}
    \vspace{-0.2cm}
    \caption{Qualitative comparison between EEmo-Logic and other MLLMs on the Artphoto dominant emotion classification.}
    \label{fig:DEC}
\end{figure}

\begin{figure}[!t]
    \centering
    \includegraphics[width=0.85\linewidth]{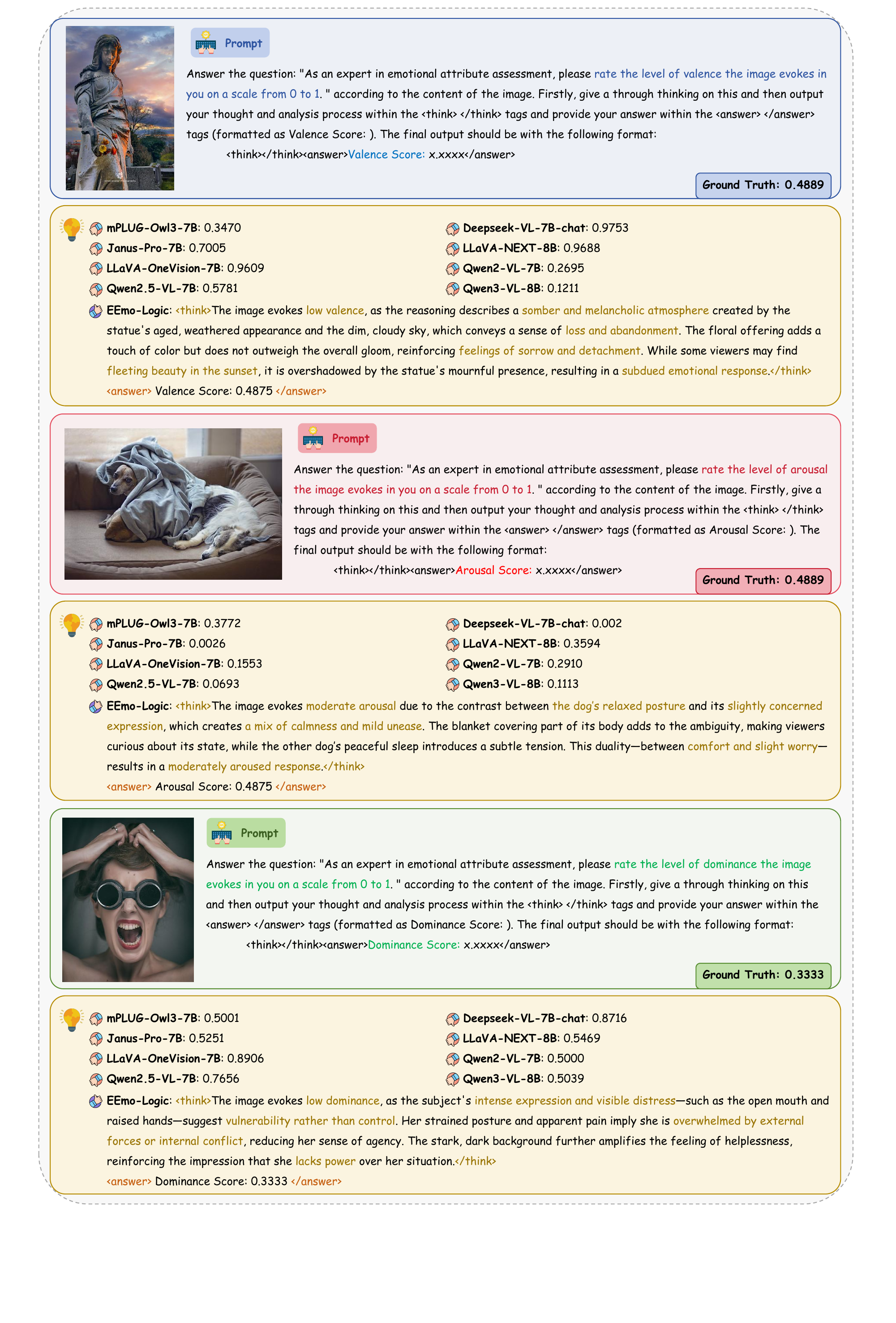}
    \vspace{-0.2cm}
    \caption{Qualitative comparison between EEmo-Logic and other MLLMs on the EEmo-Bench VAD assessment tasks.}
    \label{fig:VAD assessment}
\end{figure}

In addition to Fig.~\ref{fig: case study}, we present further qualitative results. Figs.~\ref{fig:description}, \ref{fig:perception} demonstrate EEmo-Logic's exceptional comprehension in challenging QA scenarios from EEmo-Bench, validating the efficacy of our Stage $1$ QA training. Figs.~\ref{fig:ranking}$-$\ref{fig:VAD assessment} showcase EEmo-Logic's fine-grained perception across $5$ emotion dimensions in a zero-shot setting, alongside its human-like causal reasoning. These results further confirm the effectiveness of our Stage $2$ training.

\end{document}